%% file: acl_latex.tex
\documentclass[11pt]{article}

\usepackage[preprint]{acl}

\usepackage{times}
\usepackage{latexsym}

\usepackage[T1]{fontenc}

\usepackage[utf8]{inputenc}

\usepackage{microtype}

\usepackage{inconsolata}

\usepackage{graphicx}

\usepackage{subcaption}
\usepackage{tikz}
\usetikzlibrary{shapes.geometric, arrows.meta, positioning, calc, chains}

\usepackage[inline]{enumitem}

\usepackage{multirow}
\usepackage{booktabs}
\usepackage{siunitx}
\usepackage{tabularx}
\usepackage{makecell}
\usepackage[table]{xcolor}
\usepackage{pifont}
\usepackage{amssymb}
\usepackage{amsmath}
\usepackage{cancel}
\usepackage{array}

\newcolumntype{C}{>{\centering\arraybackslash}X}
\usepackage{colortbl}

\usepackage{listings}

\newcommand{\red}[1]{{\color{red}#1}}
\newcommand{\best}[1]{\colorbox{purple!30}{#1}}
\newcommand{\secondbest}[1]{\colorbox{orange!25}{#1}}
\newcommand{\thirdbest}[1]{\colorbox{gray!30}{#1}}

\definecolor{darkgreen}{rgb}{0.0, 0.6, 0.0}

\newcommand{\gup}[1]{(\textcolor{darkgreen}{$\uparrow$ #1})}
\newcommand{\rd}[1]{(\red{$\downarrow$ #1})}

\definecolor{softpurple}{HTML}{E6CEE3}
\definecolor{softblue}{HTML}{E1E1F9}
\definecolor{softorange}{HTML}{FCD5B5}

\newcommand{\hlfor}[1]{\setlength{\fboxsep}{1.5pt}\colorbox{softpurple}{#1}}
\newcommand{\hlcons}[1]{\setlength{\fboxsep}{1.5pt}\colorbox{softblue}{#1}}
\newcommand{\hllogic}[1]{\setlength{\fboxsep}{1.5pt}\colorbox{softorange}{#1}}

\title{PersonaPath: Towards Knowledge-Centric Personalized\\ Learning Path Planning}

\author{
  \textbf{Yu Liu\textsuperscript{1}},
  \textbf{Zeming Liu\textsuperscript{1\textdagger}},
  \textbf{Tianle Zhang\textsuperscript{1}},
  \textbf{Zihao Cheng\textsuperscript{1}},
  \textbf{Yuhang Guo\textsuperscript{2}}, \\
  \textbf{Kehai Chen\textsuperscript{3}},
  \textbf{Min Zhang\textsuperscript{3}},
  \textbf{Yunhong Wang\textsuperscript{1}},
  \textbf{Haifeng Wang\textsuperscript{4}} \\[0.5ex]
  \textsuperscript{1}School of Computer Science and Engineering, Beihang University, Beijing, China \\
  \textsuperscript{2}Beijing Institute of Technology \quad
  \textsuperscript{3}Harbin Institute of Technology (Shenzhen) \quad
  \textsuperscript{4}Baidu Inc. \\
  \textsuperscript{\textdagger}Corresponding author \quad Email: \texttt{liuyuu@buaa.edu.cn, zmliu@buaa.edu.cn}
}

\usepackage{tcolorbox}
\tcbuselibrary{breakable}
\lstdefinestyle{promptstyle}{
    basicstyle=\ttfamily\small,
    breaklines=true,
    breakindent=0pt,
    showstringspaces=false,
    frame=none,
    numbers=none,
    keepspaces=true,
    columns=fullflexible,
    tabsize=2,
}

\tcbset {
  base/.style={
    arc=0mm,
    bottomtitle=-0.25mm,
    boxrule=0mm,
    colbacktitle=black!10!white,
    coltitle=black,
    fonttitle=\bfseries,
    left=2.5mm,
    leftrule=1mm,
    right=3.5mm,
    title={#1},
    toptitle=0.25mm,
    breakable,
  }
}

\definecolor{brandblue}{rgb}{0.34, 0.7, 1}
\newtcolorbox{mybox}[1]{
  colframe=brandblue,
  base={#1},
  unbreakable
}

\begin{document}
\maketitle

\input{sec/000abstract}

\input{sec/010introduction}

\input{sec/020relatedwork}

\input{sec/030benchmark}

\input{sec/040experiments}

\input{sec/050analysis}

\input{sec/060conclusion}

\section*{Acknowledgments}
Thanks for the insightful comments and feedback from the reviewers. This work was supported by the National Natural Science Foundation of China (No.~62406015).

\bibliography{custom}

\input{sec/070appendix}

\end{document}

%% file: sec/000abstract.tex
\begin{abstract}
    Adaptive learning systems commonly formulate learning path planning as Exercise-Centric (EC) recommendation, where the next step is inferred from item-level interaction logs.
    Evaluating goal-oriented guidance additionally requires explicit learner goals and curriculum-scale prerequisites: learners with similar exercise records may need different paths toward their targets.
    We therefore study Knowledge-Centric (KC) personalized learning path planning, where a planner must reason over learner profiles, mastery states, and prerequisite knowledge structures to decide which textbook, unit, and concept should be studied next.
    To support this setting, we introduce \textbf{PersonaPath}, a benchmark that pairs 2,000 fine-grained learner personas with a hierarchical knowledge graph of 347 textbooks, 1,751 units, and 4,092 concepts across 77 subjects.
    We evaluate representative LLMs on PersonaPath.
    Results show that even the strongest LLM reaches only a 29.5\% final pass rate in Basic Education, and that the main bottleneck lies in adaptivity, where no model exceeds 44.7\% in tailoring paths to individual learners\footnote{The data and the code are available at \url{https://github.com/BUAA-IRIP-LLM/PersonaPath}.}.
\end{abstract}

%% file: sec/010introduction.tex
\begin{figure*}[t!]
  \centering
  \includegraphics[width=\textwidth]{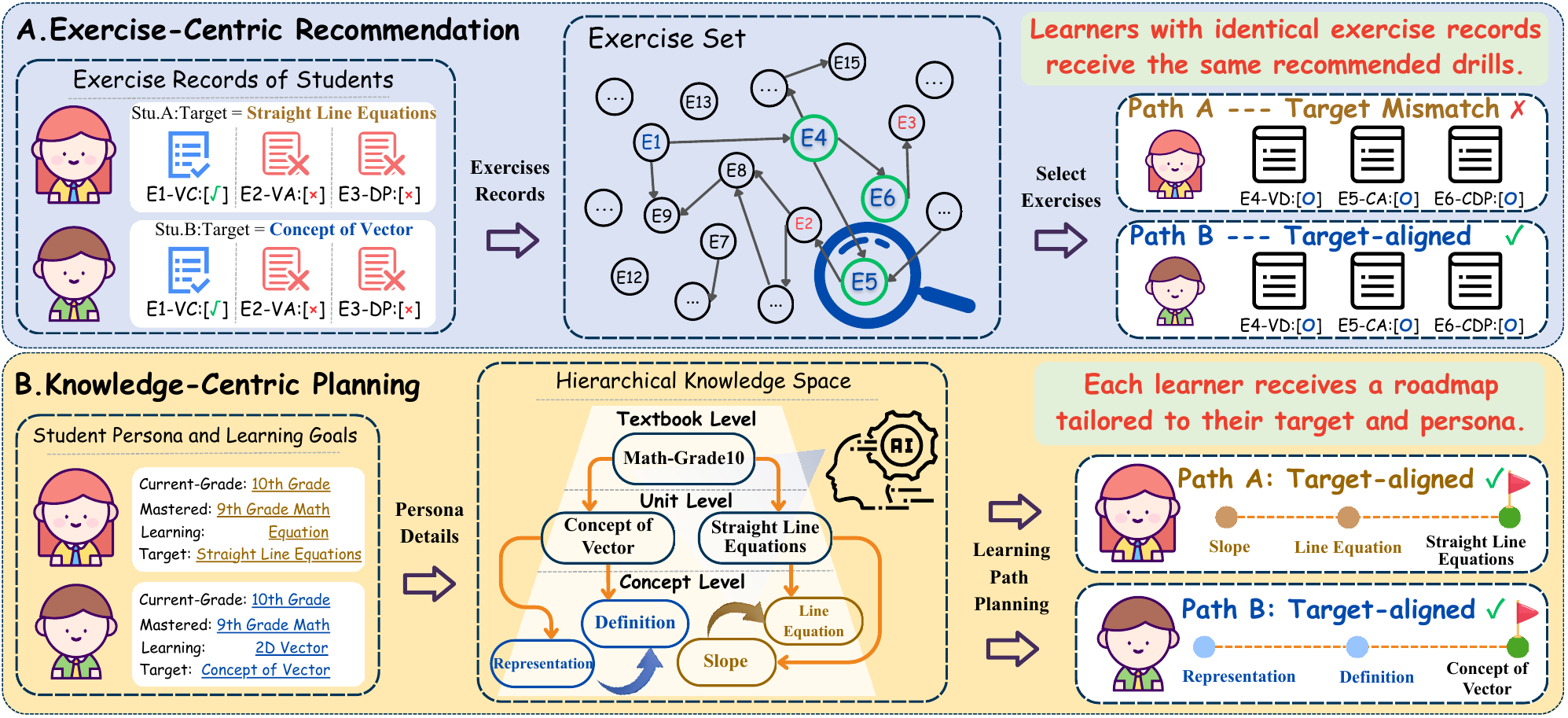}
  \caption{\textbf{Illustration of Exercise-Centric (EC) recommendation and Knowledge-Centric (KC) planning.} Panel A illustrates recommendations conditioned on correctness records alone; Panel B additionally uses explicit learner goals and mastery states. Each node (E1, E2, ...) represents an exercise item, and edges denote prerequisite relations among the knowledge points. Abbreviations: VC = Vector Coordinates, VA = Vector Addition, DP = Dot Product, VD = Vector Decomposition, CA = Component-wise Addition, and CDP = Component-wise Dot Product.}
  \label{fig:data_sample}
\end{figure*}

\section{Introduction}
\label{intro}
Learning path planning decides what a learner should study next and in what order~\citep{li2024bringinggenerativeaiadaptive, tu2025empowering, hedi2025artificial}. By adapting this sequence to each learner's current knowledge and goals, it makes study more efficient and better personalized~\citep{li2023graph}.
Previous methods typically follow an Exercise-Centric (EC) recommendation paradigm~\citep{liu2019exploiting, zhang2024item}, modeling each learner from item-level logs, such as which exercises were attempted and answered correctly, and recommending the next exercise to practice.

Interaction content supports learner-state estimation and sequential recommendation, while explicit goals specify which knowledge a path should ultimately reach. For example, in Figure~\ref{fig:data_sample}A, two students attempt the same exercises and produce identical correctness records. Both solve the vector-coordinate item (E1) but miss vector addition (E2) and dot product (E3). A recommendation based on this record alone can assign them the same sequence of vector decomposition (E4), component-wise addition (E5), and component-wise dot product (E6). Yet their goals diverge, with Student A targeting straight-line equations and Student B targeting vector concepts. The illustrated sequence follows Student B's goal, whereas Student A needs a path connecting her current mastery to straight-line equations. Evaluating this distinction requires the learner's target and the prerequisite structure connecting it to the current knowledge state.

To bring each learner's individual needs into focus, we propose a Knowledge-Centric (KC) paradigm, which decides what knowledge a learner should study and in what order. As shown in Figure~\ref{fig:data_sample}B, for two different learners, a KC planner reads each one's progress and goal, selects the knowledge they need from a structured knowledge space, and orders it into a learning path that respects the prerequisite relations among concepts. Planning thus shifts from which exercise to practice next to which knowledge to study next, and why.
Existing benchmarks primarily support exercise recommendation and student performance prediction. They rarely jointly provide explicit learner personas, long-term goals, and curriculum-scale prerequisite structures. Evaluating KC planning requires these components together: the goal determines the destination, the mastery state identifies the learner's starting point, and the prerequisite graph constrains the routes between them.

To evaluate Knowledge-Centric personalized learning path planning, we propose \textbf{PersonaPath}. It contains 2,000 fine-grained learner personas from primary school to higher education. The personas are grounded in a hierarchical knowledge space of 347 textbooks, 1,751 units, and 4,092 concepts across 77 subjects, connected by prerequisite dependencies. Given a learner profile and a target unit, a model must generate a path that respects prerequisites, adapts to prior knowledge, and efficiently reaches the goal. We benchmark representative large language models on PersonaPath and find that they still struggle with KC planning, especially in adapting paths to individual learners. PersonaPath thus advances personalized learning path planning toward a knowledge-centric paradigm, helping educational systems better address each learner's individual needs.

This work makes three contributions:
\begin{itemize}
  \item We identify an evaluation gap in personalized learning: existing benchmarks rarely combine explicit learner personas, long-term goals, and curriculum-scale prerequisites for assessing goal-directed knowledge paths.

  \item To address this, we propose KC path planning and construct \textbf{PersonaPath}, a benchmark that integrates learner personas, target goals, and prerequisite curriculum structures to evaluate LLMs on KC tasks.

  \item We evaluate representative baselines on PersonaPath and show that LLMs still struggle with KC planning, especially in personalized adaptation. Ablations examine how mastery information, contextual noise, and step-by-step feedback affect path quality.
\end{itemize}

%% file: sec/020relatedwork.tex
\input{tables/bench_compare}
\section{Related Work}
\label{sec:related_work}
\subsection{Learning Path Recommendation}
Most Learning Path Recommendation (LPR) studies \citep{zhang2021recommender} operationalize personalized learning path planning through the Exercise-Centric (EC) paradigm, where paths are inferred from item-level interaction content, such as exercise attempts, correctness records, hints, and response sequences \citep{liu2019exploiting, zhang2024item}. Existing approaches generally fall into two categories: similarity-based heuristics and Deep Reinforcement Learning (DRL) frameworks. Traditional methods often treat LPR as a combinatorial optimization problem, using techniques such as collaborative filtering \citep{yu2018multiple} or metaheuristic search, including Ant Colony Optimization \citep{niknam2020lpr}, immune algorithms \citep{cun2019adaptive}, and genetic algorithms \citep{elshani2021constructing}, to discover paths based on similar learner cohorts. More recently, DRL-based frameworks have become a prominent line of work \citep{li2023graph}, viewing learning as a sequential decision-making process. These models use architectures such as RNNs or graph-based agents to maximize a cumulative reward defined by the learner's performance on subsequent interactions \citep{zhou2018personalized, zhang2024item}. Such approaches already incorporate learner states and sequential decisions, including goal-oriented recommendation over graph structures~\citep{li2023graph,liu2022graph}.

Related benchmarks outside education evaluate dependency-aware API planning, preference-sensitive tool selection, and real-world travel planning~\citep{wang2024appbench,cheng2025toolspectrum,deng2025retail}. PersonaPath brings these concerns into curriculum planning, where actions must respect prerequisite knowledge and adapt to learner states.

Knowledge Tracing (KT) estimates and updates mastery from response sequences~\citep{corbett1995knowledge, piech2015deep}, while Cognitive Diagnosis (CD) infers concept-level proficiency from response data~\citep{cheng2019enhancing}. These methods provide learner-state estimates that can inform a downstream planner. PersonaPath supplies the mastery state as an input and evaluates the resulting prerequisite-aware path toward an explicit target. Its evaluation focus is the selection and ordering of knowledge given that state. Appendix~\ref{sec:appendix-knowledge-level} discusses the connection in more detail.
\subsection{Evaluation Benchmarks and Datasets}
Standard benchmarks in this domain primarily consist of large-scale interaction logs recorded from online learning platforms, including the \textbf{ASSISTments} family \citep{feng2009addressing, wang2015towards}, \textbf{Junyi Academy} \citep{chang2015modeling}, and \textbf{OLI Engineering Statics} \citep{Fall-2011-OLI-Data}, together with synthetic response data generated for Deep Knowledge Tracing \citep{piech2015deep}. These resources support student performance prediction and learner-state modeling from item-level records. We describe each dataset in detail in Appendix~\ref{ref:dataset_comparison}.
As summarized in Table~\ref{tab:comparison-between-datasets}, the gap for KC evaluation is the joint availability of explicit learner personas, long-term goals, and curriculum-scale prerequisites. Skill labels and interaction sequences provide useful evidence about mastery; evaluating a goal-directed path additionally requires a structured curriculum connecting the current state to the target. PersonaPath pairs learner personas and goals with a textbook--unit--concept hierarchy and prerequisite relations to support this evaluation.

%% file: tables/bench_compare.tex
\definecolor{Color1}{HTML}{FA7F6F}
\definecolor{Color2}{HTML}{8ECFC9}
\definecolor{Color3}{HTML}{FFBE7A}
\definecolor{Color4}{HTML}{82B0D2}
\definecolor{Color5}{HTML}{BEB8DC}
\definecolor{Color6}{HTML}{E7DAD4}

\begin{table*}[!t]
\centering
\resizebox{\textwidth}{!}{%
\begin{tabular}{@{}lccccccc@{}}
\toprule
Benchmark & Attribute & Task & TPG & PSN & BED & HKS & RS \\ \midrule

Junyi Academy~\cite{chang2015modeling}
    & Exercise-Centric
    & Question Recommendation
    & {\color[HTML]{FF0000} \ding{55}} & {\color[HTML]{FF0000} \ding{55}} & \textcolor{orange}{\scalebox{1}{\bcancel{\checkmark}}} & {\color[HTML]{FF0000} \ding{55}} & {\color[HTML]{FF0000} \ding{55}} \\

ASSISTments2009~\cite{feng2009addressing}
    & Exercise-Centric
    & Question Recommendation
    & {\color[HTML]{FF0000} \ding{55}} & {\color[HTML]{FF0000} \ding{55}} & \textcolor{orange}{\scalebox{1}{\bcancel{\checkmark}}} & {\color[HTML]{FF0000} \ding{55}} & {\color[HTML]{FF0000} \ding{55}} \\

ASSISTments2012~\cite{wang2015towards}
    & Exercise-Centric
    & Question Recommendation
    & {\color[HTML]{FF0000} \ding{55}} & \textcolor{orange}{\scalebox{1}{\bcancel{\checkmark}}} & \textcolor{orange}{\scalebox{1}{\bcancel{\checkmark}}} & {\color[HTML]{FF0000} \ding{55}} & {\color[HTML]{FF0000} \ding{55}} \\

OLI Engineering Statics~\cite{Fall-2011-OLI-Data}
    & Exercise-Centric
    & Question Recommendation
    & {\color[HTML]{FF0000} \ding{55}} & {\color[HTML]{FF0000} \ding{55}} & \textcolor{orange}{\scalebox{1}{\bcancel{\checkmark}}} & {\color[HTML]{FF0000} \ding{55}} & {\color[HTML]{FF0000} \ding{55}} \\

Synthetic Data from DKT~\cite{piech2015deep}
    & Exercise-Centric
    & Question Recommendation
    & {\color[HTML]{FF0000} \ding{55}} & {\color[HTML]{FF0000} \ding{55}} & \textcolor{orange}{\scalebox{1}{\bcancel{\checkmark}}} & {\color[HTML]{FF0000} \ding{55}} & {\color[HTML]{FF0000} \ding{55}} \\

\midrule

\textbf{PersonaPath (Ours)}
    & \textbf{Knowledge-Centric}
    & \textbf{Learning Path Planning}
    & {\color[HTML]{00B050} \ding{51}} & {\color[HTML]{00B050} \ding{51}} & {\color[HTML]{00B050} \ding{51}} & {\color[HTML]{00B050} \ding{51}} & {\color[HTML]{00B050} \ding{51}} \\ \bottomrule
\end{tabular}%
}

\caption{Comparison between PersonaPath and other benchmarks. ``TPG'', ``PSN'', ``BED'', ``HKS'', and ``RS'' represent Textbook-level Prerequisite Graph, Persona, Breadth of Educational Disciplines, Hierarchical Knowledge Structure, and Rich Semantics, respectively. ``{\color[HTML]{FF0000} \ding{55}}'' indicates the absence of a dimension, while ``\textcolor{orange}{\scalebox{1}{\bcancel{\checkmark}}}'' indicates that only part of it is included.}
\label{tab:comparison-between-datasets}
\end{table*}

%% file: sec/030benchmark.tex
\section{PersonaPath}
\begin{figure*}[t!]
    \centering
    \includegraphics[width=\textwidth]{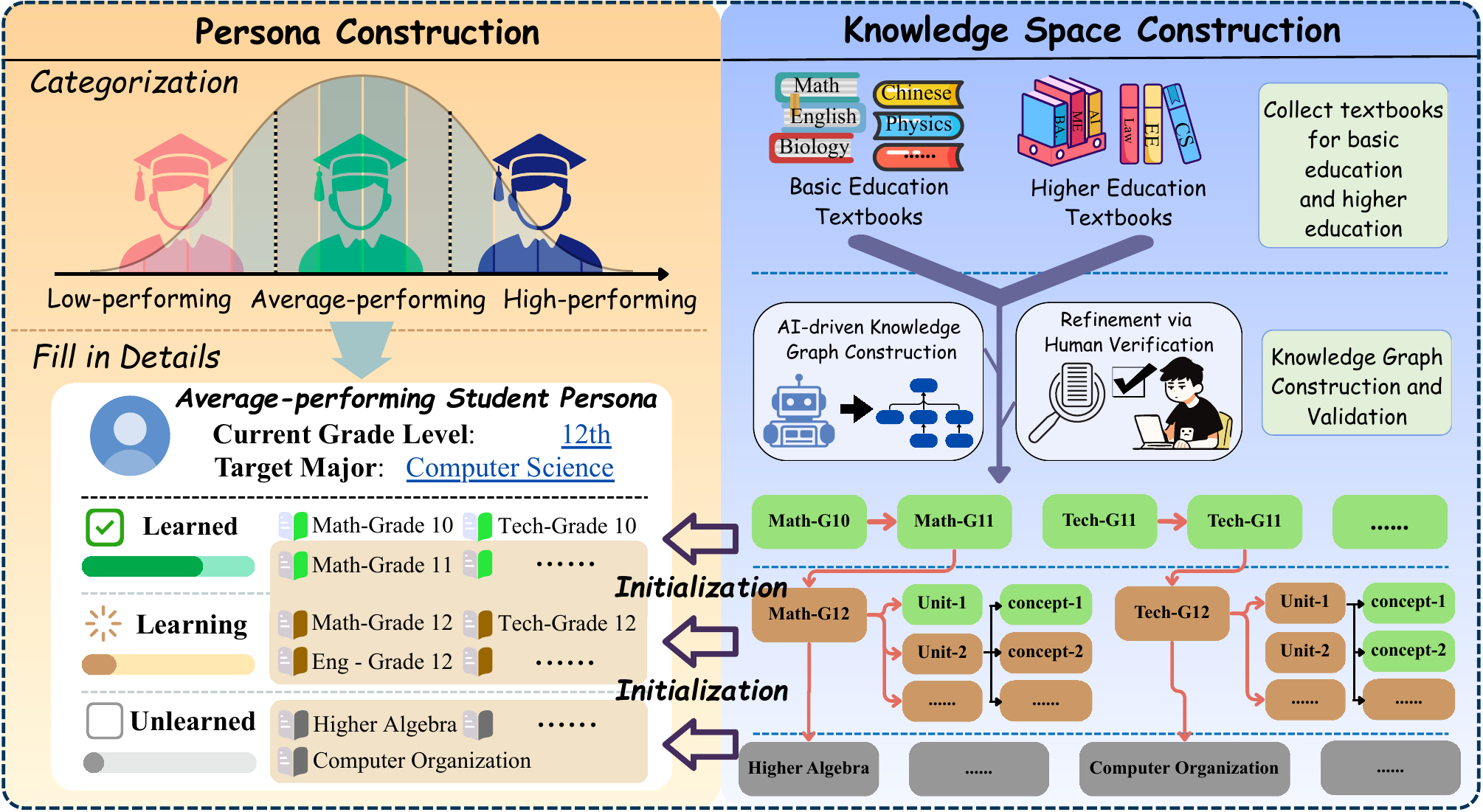}
    \caption{The construction pipeline of \textbf{PersonaPath}. The benchmark first builds explicit learner personas and a hierarchical knowledge space, then initializes each learner's mastery state so that KC planners can be evaluated on goal-oriented, prerequisite-aware, and persona-adaptive learning paths.}
    \label{fig:data-gen}
\end{figure*}

\subsection{Task Definition}
PersonaPath evaluates Knowledge-Centric personalized learning path planning. Given a learner profile, a mastery state, a target unit, and a hierarchical knowledge space, the model must generate a sequence of knowledge concepts that helps the learner reach the target while respecting curriculum prerequisites. The task evaluates how a planner uses the supplied knowledge structure and learner state to select its next action.

\vspace{1ex}
\noindent \textbf{Knowledge Space Definition.}
We define the knowledge space as a Hierarchical Knowledge Graph, denoted by $\mathcal{G} = (\mathcal{V}, \mathcal{E}, \mathcal{H})$. The vertex set $\mathcal{V}$ is hierarchically partitioned into three levels: Textbooks $\mathcal{B} = \{b_1, ..., b_K\}$, Units $\mathcal{U} = \{u_1, ..., u_M\}$, and Atomic Concepts $\mathcal{C} = \{c_1, ..., c_N\}$. The edge set $\mathcal{E} \subseteq \mathcal{B} \times \mathcal{B}$ defines the directed prerequisite order between textbooks. The set $\mathcal{H}$ encodes the vertical inclusion structure, where each atomic concept $c$ belongs to a unique unit $u$, and each unit $u$ belongs to a unique textbook $b$. Prerequisite relations at the unit and concept levels are induced from $\mathcal{E}$ via $\mathcal{H}$.

\vspace{1ex}
\noindent \textbf{Cognitive State and Knowledge Acquisition.}
We distinguish between the static attributes of knowledge and the dynamic cognitive state of the learner. Let $D: \mathcal{C} \to [0, 1]$ denote the inherent difficulty function associated with atomic concepts. Conversely, learner proficiency is tracked via a dynamic mastery vector $\mathbf{M}_t \in [0, 1]^{|\mathcal{U}|}$, defined at the unit level at time step $t$.
The planning process operates as follows: at each step $t$, given the learner's state $\mathcal{S}_t = (\mathbf{P}, \mathbf{M}_t)$ (where $\mathbf{P}$ represents the static persona profile), the agent selects a knowledge concept $a_t \in \mathcal{C}$. The environment then simulates the cognitive update of the corresponding parent unit $u$ via a transition function (see Appendix~\ref{sec:appendix_sim} for details of the calculation): $\mathbf{M}_{t+1}[u] \leftarrow \text{Sim}(\mathbf{M}_t[u], D(a_t))$. A detailed explanation of the hierarchical transition mechanism is provided in Appendix~\ref{sec:hierarchical-transition}.
Thus, the action and state have different granularities: concepts are the selectable skills, while units are the coordinates of the mastery vector. Concepts within a unit share its mastery value but retain their individual difficulty values. The updated unit mastery conditions the next concept selection. This representation tracks 1,751 unit states over a planning space of 4,092 concepts.

\vspace{1ex}
\noindent \textbf{Planning Objective.}
The goal is to generate an optimal learning path $\mathcal{P} = [a_1, a_2, ..., a_T]$ for a specified target unit $g_{target} \in \mathcal{U}$. The process terminates when the learner's mastery of the target unit meets the educational standard: $\mathbf{M}_T[g_{target}] \ge \tau$, where $\tau$ is a predefined proficiency threshold.

\subsection{Data Construction}
The data construction pipeline is designed around the three inputs required by KC planning: explicit learner goals, prerequisite knowledge structure, and learner-specific mastery states. As illustrated in Figure~\ref{fig:data-gen}, the pipeline consists of \textit{Persona Construction}, \textit{Knowledge Space Construction}, and \textit{Persona Initialization}. These components support evaluating whether a model selects appropriate content for a learner at each step toward the target.

\vspace{1ex}
\noindent \textbf{Persona Construction.}
\label{sec:data_construction}
Following the synthetic student modeling approach of \citet{piech2015deep}, we define a learner persona by a specific academic goal (e.g., a target major at university) and a structured knowledge mastery state. This design makes the planning target explicit: the model must decide how to bridge the gap between the learner's current state and the target knowledge. To simulate varied learning abilities, we categorize learners into \textit{Low-performing}, \textit{Average-performing}, and \textit{High-performing} archetypes in an approximate 1:3:1 ratio, which corresponds to tertile bins of the standard normal ability distribution commonly used in Item Response Theory~\citep{lord2012applications}. These archetypes govern the learner's simulated learning efficiency and difficulty tolerance, enabling the benchmark to evaluate planning performance across different learner profiles.
Furthermore, we position each learner persona at a specific \textit{Current Grade Level}, which splits the curriculum into three regions: \textit{Learned} (already mastered), \textit{Learning} (under active study), and \textit{Unlearned} (next objectives). This grade anchor fixes both the planning horizon and the learner's initial knowledge state.

\vspace{1ex}
\noindent \textbf{Knowledge Space Construction.} To provide the prerequisite structure required by KC planning, we built a knowledge base from authoritative textbooks: Basic Education textbooks from \textit{People's Education Press}\footnote{https://www.pep.com.cn/} and Higher Education resources from the \textit{Smart Education of China} platform\footnote{https://smartedu.tbook.com.cn/}. Textbooks were ordered by subject following the standard teaching sequence for Basic Education and the internal learning path of each major for Higher Education, with Basic Education subjects serving as prerequisites for related Higher Education majors (e.g., Chemistry and Biology before Clinical Medicine).
To transform these textbooks into a structured Knowledge Graph, we adopted a hybrid human-AI workflow. We used DeepSeek-V3 as the extraction engine because of its performance on entity and relation extraction \citep{zhao2025quantifying, xu2024large} and its proficiency in Chinese educational text. Following a hierarchical extraction strategy \citep{chen2025largelanguagemodelseffective, lu2019concept}, the model first parsed the tables of contents to build the structural hierarchy and then identified atomic concepts within each unit to form a fine-grained Knowledge Graph. Domain experts manually reviewed the output to correct structural errors and extraction inaccuracies.

\vspace{1ex}
\noindent \textbf{Persona Initialization.} We initialize mastery values so that each persona becomes an actionable planning state rather than a static user description. For \textit{Learned} content, mastery scores are assigned according to the persona's proficiency archetype (e.g., \textit{Low-performing} or \textit{High-performing}), with distribution details in Appendix~\ref{sec:appendix-Benchmark}. For \textit{Learning} content, mastery levels are randomly initialized within the interval $(0, 1)$. To simulate the cumulative nature of learning, we implement a dynamic knowledge transfer mechanism within the test environment. When a learner moves to a new textbook, the initial mastery for upcoming concepts is not zero but is derived from the proficiency levels attained in the corresponding prerequisite subjects.

\subsection{Quality Control}
\label{sec:quality}
Quality control focuses on the elements that determine whether KC paths can be evaluated reliably: prerequisite validity, entity correctness, and persona consistency. For prerequisite relations, we combined automated and manual checks: graph algorithms inspected the Directed Acyclic Graph (DAG) structure for circular dependencies, following the precedence axioms in Knowledge Space Theory \citep{doignon2012knowledge}. A panel of two domain experts also reviewed the pedagogical logic of the links and verified the semantic accuracy of the LLM-generated entities. To assess consistency, we calculated Inter-Annotator Agreement (IAA) on a validation subset reviewed by both experts, using Cohen's Kappa \citep{cohen1960coefficient}.
To improve the coherence of initialized personas, we implemented sanity checks to avoid logical inconsistencies. These constraints verify that (1) \textit{High-performing} personas have correspondingly high mastery scores in previously \textit{Learned} content and (2) a persona positioned at a higher \textit{Current Grade Level} has completed all curriculum requirements from preceding educational stages.

\noindent \textbf{Expert Evaluation of Personas.}
Two experienced educators independently evaluated 100 randomly sampled personas using a 5-point Likert scale. They assessed four dimensions: curriculum consistency, progression plausibility, mastery plausibility, and persona coherence. The study evaluates whether the generated learner profiles and initialized mastery states form pedagogically plausible, internally coherent representations. We report the mean rating for each dimension and inter-rater reliability using ICC(2,2), the two-way random-effects, absolute-agreement coefficient for the average of the two raters.

\subsection{Quality and Data Statistics}
\input{tables/bench_statistics}
We quantitatively assessed the structural integrity and logical validity of the \textbf{PersonaPath} dataset.
Regarding prerequisite relations, structural analysis verified strict DAG compliance across \textbf{347} textbook nodes and \textbf{411} prerequisite edges, confirming zero circular dependencies. Manual verification on the validation subset yielded a Cohen's Kappa of \textbf{0.93}, which falls within the ``almost perfect'' agreement range \citep{landis1977measurement}, indicating strong expert consensus. After expert correction, the verified prerequisite relations reached a precision of \textbf{99.5}\%. For content correctness, the evaluation confirmed the semantic accuracy of the LLM-generated entities, ensuring a reliable knowledge environment.
Initialized personas are filtered by two logical sanity rules: (1) \textit{High-performing} personas must hold mastery scores above \textbf{0.85} in learned content; (2) personas placed at a higher \textit{Current Grade Level} must satisfy chronological consistency (no missing prerequisites). The released personas satisfy these rules by construction.

\input{tables/persona_expert}
Table~\ref{tab:persona-expert} reports the expert evaluation of the sampled personas. Mean ratings range from 4.11 to 4.68 out of 5, with ICC(2,2) values from 0.74 to 0.85. Curriculum consistency receives the highest rating (4.68), while mastery plausibility has the lowest mean (4.11) and inter-rater reliability (0.74). These judgments support the pedagogical plausibility and internal coherence of the synthetic profiles.

Table~\ref{tab:benchmark_stats} presents the statistical breakdown of the benchmark's two core components. Part A depicts a vertically structured knowledge graph where the number of concepts grows with educational progression, expanding from \textbf{930} concepts at the Primary level to \textbf{1,601} at the University level.
Part B summarizes a cohort of \textbf{2,000} learners, evenly balanced between Basic and Higher Education. These learners are stratified into \textit{Low-performing}, \textit{Average-performing}, and \textit{High-performing} archetypes in an approximate \textbf{1:3:1} ratio, supporting the evaluation of personalization across different learner proficiency levels.

%% file: tables/bench_statistics.tex
\begin{table}[htbp]
\centering

\footnotesize
\setlength{\tabcolsep}{2pt}

\begin{tabularx}{\columnwidth}{lCCCC}

\toprule
\multicolumn{5}{l}{\textbf{A: Textbook Statistics}} \\
\midrule
Stage & Subject & Books & Units & Concepts \\
\midrule
Primary    & 8  & 71  & 416 & 930 \\
Middle     & 12 & 58  & 286 & 680 \\
High       & 12 & 78  & 344 & 881 \\
University & 45 & 140 & 705 & 1,601 \\
\midrule
\textbf{Total} & \textbf{77} & \textbf{347} & \textbf{1,751} & \textbf{4,092} \\
\bottomrule
\addlinespace[2ex]

\toprule
\multicolumn{5}{l}{\textbf{B: Persona Statistics}} \\
\midrule
Category & \multicolumn{2}{c}{Basic Education} & \multicolumn{2}{c}{Higher Education} \\
\midrule
Low-performing     & \multicolumn{2}{c}{193} & \multicolumn{2}{c}{200} \\
Average-performing & \multicolumn{2}{c}{620} & \multicolumn{2}{c}{609} \\
High-performing    & \multicolumn{2}{c}{187} & \multicolumn{2}{c}{191} \\
\midrule
\textbf{Total} & \multicolumn{2}{c}{\textbf{1,000}} & \multicolumn{2}{c}{\textbf{1,000}} \\
\bottomrule
\end{tabularx}
\caption{Overview of the benchmark statistics.}
\label{tab:benchmark_stats}
\end{table}

%% file: tables/persona_expert.tex
\begin{table}[t]
\centering
\small
\setlength{\tabcolsep}{5pt}
\begin{tabular*}{0.95\columnwidth}{@{\extracolsep{\fill}}lcc}
\toprule
Dimension & Mean / 5 & ICC(2,2) \\
\midrule
Curriculum consistency & 4.68 & 0.85 \\
Progression plausibility & 4.37 & 0.82 \\
Mastery plausibility & 4.11 & 0.74 \\
Persona coherence & 4.33 & 0.79 \\
\bottomrule
\end{tabular*}
\caption{Expert evaluation of 100 randomly sampled personas. Two educators independently rated each persona on a 5-point Likert scale. ICC(2,2) measures agreement for their average rating.}
\label{tab:persona-expert}
\end{table}

%% file: sec/040experiments.tex
\section{Experiments}
\subsection{Experimental Setup}

\noindent \textbf{Baselines.}
We evaluate ten open-source LLMs covering different architectures and parameter scales (1B to 30B+). The evaluation set includes: (1) the \textbf{Pangu series} \cite{chen2025panguembeddedefficientdualsystem} (1B, 7B); (2) the \textbf{Llama series} \cite{grattafiori2024llama} (3.2-1B, 3.1-8B); (3) the \textbf{Qwen series} \cite{yang2025qwen3} (Qwen3-4B-2507, Qwen3-30B-A3B); (4) the \textbf{DeepSeek series} \cite{guo2025deepseek}, comprising the reasoning-optimized R1-Distill models (7B, 14B) and DeepSeek-V3.1; and (5) \textbf{InnoSpark} \cite{song2025cultivatinghelpfulpersonalizedcreative}, specifically InnoSpark-7B, an education-focused model whose training includes supervised fine-tuning and multi-stage reinforcement learning. To examine whether reasoning chains help, we evaluate these models under both \textit{Zero-shot} and \textit{Chain-of-Thought} (CoT) settings \citep{wei2022chain}.

\noindent \textbf{Interactive Planning and Retrieval.}
The main evaluation uses a closed-loop protocol: each concept selection is followed by a simulated mastery update, and the updated state is supplied for the next decision (Figure~\ref{fig:workflow}). We also evaluate a retrieval-augmented Qwen3-4B-2507$_{Zero-shot}$ on Basic Education. It retrieves the top-3 knowledge chunks using \texttt{BAAI/bge-small-zh-v1.5} embeddings and FAISS, then prepends them to the planning prompt. Table~\ref{tab:rag-baseline} reports the results; Appendix~\ref{sec:appendix-rag} gives the retrieval configuration.

\noindent \textbf{Implementation Details.}
Experiments were conducted on a high-performance computing cluster with Huawei Ascend 910B1 NPUs (64~GB HBM) and Kunpeng-920 CPUs (192 cores, AArch64 architecture), running Huawei Cloud EulerOS 2.0. We used the \texttt{vLLM} library (v0.9.2) for inference, with all models loaded in bfloat16 precision. We set the sampling temperature to 0.1 and the maximum generation length to 32,768 tokens, while retaining the default vLLM settings for other parameters.

\subsection{Evaluation Metrics}
\label{sec:evaluation}
We evaluate each path along three dimensions.
\textit{Validity} contains \textit{Prerequisite Violation} (no concept appears before its antecedents, per Knowledge Space Theory~\citep{doignon2012knowledge}) and \textit{Hallucination} (no entity falls outside the knowledge space).

\textit{Adaptivity} contains \textit{Persona Alignment} (consistency with the learner's archetype and prior experience, motivated by the Zone of Proximal Development~\citep{vygotsky1978mind}) and \textit{Difficulty Adaptability}. We measure the latter using Cog-Gap~\citep{zhang2024item}, the average absolute difference between mastery and concept difficulty along the path:
\begin{equation}
\label{eq:cog-gap}
\text{Cog-Gap} = \frac{1}{n} \sum_{t=1}^{n} \left| \mathbf{M}_t[u] - D(a_t) \right|.
\end{equation}
Here, $n$ is the number of evaluated concept selections, and $u$ denotes the parent unit of the selected concept $a_t$ at each step. The unit mastery $\mathbf{M}_t[u]$ and concept difficulty $D(a_t)$ both lie in $[0,1]$, so Cog-Gap also lies in $[0,1]$. A smaller value indicates closer alignment between the learner's current mastery and the selected difficulty. The absolute difference captures both under-challenging and over-challenging choices. Because mastery is updated during a path, the same concept difficulty can yield different gaps at different steps. This metric complements prerequisite validity by assessing whether the selected content fits the learner's current state.

\textit{Efficiency} contains \textit{Goal Completion} (reaching target mastery within a step budget $T_{\max}$) and \textit{Progress Continuity} (the average proficiency increment over a five-step sliding window must exceed a threshold).
A path receives a \textit{Final Pass} only when all three hold, $Pass_{final} = \mathbb{I}(\text{Validity}) \cdot \mathbb{I}(\text{Adaptivity}) \cdot \mathbb{I}(\text{Efficiency})$. Full definitions and thresholds are in Appendix~\ref{app:constraints}.

\input{tables/main_res}
\subsection{Results}
Table~\ref{tab:main-result} presents the performance of various LLMs across different learning stages and prompt types on PersonaPath. Three findings emerge.

\textbf{\textit{Performance drops consistently from Basic to Higher Education across almost every metric.}} Notably, the Final Pass Rate of the top-performing DeepSeek-V3.1 drops by approximately half, falling from 29.5\% to just 14.6\%. The drop appears across all model scales: the runner-up Qwen3-30B-A3B (CoT) declines from 27.7\% to 10.8\%, while the lightweight Qwen3-4B-2507 (CoT) suffers an even steeper drop from 20.8\% to 5.5\%. This gap suggests that these models may have been exposed to less university-level educational data during training and highlights the difficulty of handling the escalated structural complexity in higher education planning.

\textbf{\textit{Performance varies significantly across model series.}} Llama-3.2-1B and Llama-3.1-8B achieve Final Pass Rates of at most 1.3\% across Basic and Higher Education. These scores describe performance on the Chinese textbooks and curriculum structures used by PersonaPath. Section~\ref{sec:analysis} examines changes within each model under the same language and curriculum conditions.
In addition, InnoSpark-7B, a model specialized for education, shows a distinct ``specialist'' profile: it outperforms similar-sized baselines in \textit{Validity} with a 41.3\% pass rate, yet lags behind the same counterparts in \textit{Adaptivity}. This pattern suggests that domain specialization may skew the model toward curriculum-aligned outputs at the cost of the learner-conditioned reasoning required for Adaptivity.

\textbf{\textit{Models show a sharp imbalance across constraints: they excel in Validity but struggle with Adaptivity.}}
Specifically, models perform relatively well in \textit{Validity}, exemplified by DeepSeek-V3.1's 90.9\% pass rate, which suggests that current LLMs can often structure logically valid teaching schedules under the provided constraints.
Conversely, \textit{Adaptivity} remains the primary bottleneck: even the top-performing Qwen3-30B-A3B (CoT) achieves only 44.7\%, while DeepSeek-V3.1 reaches 44.3\%. This reveals a gap in personalization: models handle general pedagogical rules well but struggle to tailor decisions to specific learner profiles.

\noindent \textbf{Effects of CoT Prompting.}
CoT changes performance differently across models and educational stages (Table~\ref{tab:main-result}). In Basic Education, Qwen3-30B-A3B's Final Pass rises from 10.0\% under Zero-shot prompting to 27.7\% with CoT, and Qwen3-4B-2507's rises from 3.2\% to 20.8\%. Their Adaptivity scores increase from 21.9\% to 44.7\% and from 12.5\% to 43.9\%, respectively. Thus, the gains for these Qwen models include better alignment with learner states as well as higher joint pass rates.

The improvements are smaller in Higher Education: Final Pass rises from 8.6\% to 10.8\% for Qwen3-30B-A3B and from 2.4\% to 5.5\% for Qwen3-4B-2507. Llama-3.1-8B shows a different response in Basic Education, with Validity declining from 19.2\% to 14.6\% and Final Pass from 1.3\% to 0.8\%. These within-model comparisons show that the effect of CoT depends on the model and curriculum stage. Reporting both prompting settings captures this variation across Validity, Adaptivity, and the joint planning objective.

\input{tables/rag_baseline}
\noindent \textbf{Effects of Retrieval Augmentation.}
Table~\ref{tab:rag-baseline} compares Qwen3-4B-2507$_{Zero-shot}$ with its RAG variant on the same Basic Education cohort. RAG raises Validity from 39.7\% to 73.8\%, a gain of 34.1 percentage points, and Adaptivity from 12.5\% to 19.3\%. Efficiency decreases from 25.8\% to 18.4\%, while Final Pass changes from 3.2\% to 2.9\%. Retrieved context therefore improves the rate of structurally valid paths in this setting, while adaptation and efficient goal completion remain more restrictive. Since Final Pass requires all three dimensions to hold for the same path, the increase in Validity alone does not yield a higher joint pass rate. This comparison separates the benefit of supplying relevant knowledge from the overall quality of the resulting learning sequence.

The two input strategies emphasize different dimensions for the same base model. In Basic Education, Qwen3-4B-2507 with RAG obtains higher Validity than its CoT setting (73.8\% versus 55.8\%). CoT achieves higher Adaptivity (43.9\% versus 19.3\%) and Efficiency (60.2\% versus 18.4\%), yielding a Final Pass of 20.8\% compared with 2.9\% for RAG. The highest Validity and highest joint pass rate therefore occur under different settings. Evaluating both prompting and retrieval across all three dimensions captures the learner adaptation and progress requirements that a curriculum-consistency score alone would miss.

%% file: tables/main_res.tex
\setlength\tabcolsep{3pt}

\begin{table*}
\resizebox{1.0\linewidth}{!}{
\begin{tabular}{lcccccccc}
\toprule
                             & \multicolumn{4}{c}{\textbf{Basic Education} (Test\#\num{1000})}
                             & \multicolumn{4}{c}{\textbf{Higher Education} (Test\#\num{1000})}
                             \\
\cmidrule(l){2-5} \cmidrule(l){6-9}

                             & \begin{tabular}[c]{@{}c@{}}Validity\\Pass Rate\end{tabular}
                             & \begin{tabular}[c]{@{}c@{}}Adaptivity\\Pass Rate\end{tabular}
                             & \begin{tabular}[c]{@{}c@{}}Efficiency\\Pass Rate\end{tabular}
                             & \begin{tabular}[c]{@{}c@{}}Final\\Pass Rate\end{tabular}
                             & \begin{tabular}[c]{@{}c@{}}Validity\\Pass Rate\end{tabular}
                             & \begin{tabular}[c]{@{}c@{}}Adaptivity\\Pass Rate\end{tabular}
                             & \begin{tabular}[c]{@{}c@{}}Efficiency\\Pass Rate\end{tabular}
                             & \begin{tabular}[c]{@{}c@{}}Final\\Pass Rate\end{tabular}
                             \\

\midrule

\multicolumn{1}{l}{\cellcolor{white}Pangu-1B$_{CoT}$}
    & \num{10.1} & \num{11.4} & \num{20.4} & \num{1.1}
    & \num{6.4} & \num{13.6} & \num{10.8} & \num{1.0}
    \\

\multicolumn{1}{l}{\cellcolor{white}Llama-3.2-1B$_{Zero-shot}$}
    & \num{3.1} & \num{13.6} & \num{11.5} & \num{0.3}
    & \num{4.4} & \num{11.8} & \num{6.0} & \num{0.9}
    \\

\multicolumn{1}{l}{\cellcolor{white}Llama-3.2-1B$_{CoT}$}
    & \num{3.4} & \num{9.0} & \num{9.0} & \num{0.7}
    & \num{0.6} & \num{2.7} & \num{1.6} & \num{0.0}
    \\

\midrule
\multicolumn{1}{l}{\cellcolor{white}Qwen3-4B-2507$_{Zero-shot}$}
    & \num{39.7} & \num{12.5} & \num{25.8} & \num{3.2}
    & \num{8.6} & \num{13.5} & \num{6.5} & \num{2.4}
    \\

\multicolumn{1}{l}{\cellcolor{white}Qwen3-4B-2507$_{CoT}$}
    & \num{55.8} & \thirdbest{43.9} & \secondbest{60.2} & \thirdbest{20.8}
    & \num{27.6} & \num{24.2} & \num{31.1} & \num{5.5}
    \\

\midrule

\multicolumn{1}{l}{\cellcolor{white}Pangu-7B$_{CoT}$}
    & \num{29.9} & \num{28.5} & \num{46.4} & \num{15.0}
    & \num{24.6} & \num{31.9} & \num{41.3} & \num{5.4}
    \\

\multicolumn{1}{l}{\cellcolor{white}InnoSpark-7B$_{Zero-shot}$}       & \num{41.1} & \num{11.8} & \num{23.8} & \num{2.4}
    & \num{19.8} & \num{16.0} & \num{16.7} & \num{2.2}
    \\

\multicolumn{1}{l}{\cellcolor{white}InnoSpark-7B$_{CoT}$}       & \num{41.3} & \num{10.7} & \num{24.2} & \num{2.9}
    & \num{13.4} & \num{17.4} & \num{5.0} & \num{2.0}
    \\

\multicolumn{1}{l}{\cellcolor{white}DeepSeek-R1-Distill-7B$_{Zero-shot}$}
    & \num{9.8} & \num{20.2} & \num{26.6} & \num{2.5}
    & \num{3.8} & \num{8.6} & \num{6.1} & \num{1.1}
    \\

\multicolumn{1}{l}{\cellcolor{white}DeepSeek-R1-Distill-7B$_{CoT}$}
    & \num{9.9} & \num{28.2} & \num{30.6} & \num{2.5}
    & \num{8.5} & \num{16.2} & \num{12.0} & \num{1.9}
    \\
\midrule

\multicolumn{1}{l}{\cellcolor{white}Llama-3.1-8B$_{Zero-shot}$}
    & \num{19.2} & \num{11.0} & \num{14.1} & \num{0.9}
    & \num{6.1} & \num{15.4} & \num{8.0} & \num{0.9}
    \\

\multicolumn{1}{l}{\cellcolor{white}Llama-3.1-8B$_{CoT}$}
    & \num{14.6} & \num{6.6} & \num{12.6} & \num{0.8}
    & \num{2.3} & \num{12.0} & \num{2.9} & \num{0.4}
    \\

\midrule

\multicolumn{1}{l}{\cellcolor{white}DeepSeek-R1-Distill-14B$_{Zero-shot}$}
    & \num{42.3} & \num{34.0} & \num{53.9} & \num{10.8}
    & \num{35.9} & \num{27.6} & \num{40.0} & \num{4.3}
    \\

\multicolumn{1}{l}{\cellcolor{white}DeepSeek-R1-Distill-14B$_{CoT}$}
    & \num{49.9} & \num{43.4} & \num{57.1} & \num{20.4}
    & \num{36.3} & \secondbest{34.1} & \thirdbest{41.9} & \num{7.3}
    \\

\midrule

\multicolumn{1}{l}{\cellcolor{white}Qwen3-30B-A3B$_{Zero-shot}$}
    & \thirdbest{63.5} & \num{21.9} & \num{48.9} & \num{10.0}
    & \thirdbest{38.3} & \thirdbest{34.0} & \num{41.5} & \thirdbest{8.6}
    \\

\multicolumn{1}{l}{\cellcolor{white}Qwen3-30B-A3B$_{CoT}$}
    & \secondbest{70.8} & \best{44.7} & \thirdbest{59.4} & \secondbest{27.7}
    & \secondbest{44.1} & \num{33.5} & \secondbest{46.0} & \secondbest{10.8}
    \\

\midrule

\multicolumn{1}{l}{\cellcolor{white}DeepSeek-V3.1$_{Zero-shot}$}
    & \best{90.9} & \secondbest{44.3} & \best{68.1} & \best{29.5}
    & \best{57.9} & \best{42.5} & \best{51.3} & \best{14.6}
    \\

\bottomrule

\end{tabular}
}
\caption{
        \textbf{LLM performance on PersonaPath across Basic and Higher Education settings.}
        Results are reported as pass rates in percentage (\%).
        \textbf{Basic Education} and \textbf{Higher Education} denote the two evaluation splits.
        The best, second-best, and third-best results are marked \colorbox{purple!30}{purple}, \colorbox{orange!25}{orange}, and \colorbox{gray!30}{gray}, respectively.
    }
\vspace{-0.5cm}
\label{tab:main-result}
\end{table*}

%% file: tables/rag_baseline.tex
\begin{table}[htbp]
\centering
\small
\setlength\tabcolsep{3pt}
\begin{tabular*}{\columnwidth}{@{\extracolsep{\fill}}lrrrr}
\toprule
Setting & Validity & Adaptivity & Efficiency & Final
 \\
\midrule

Zero-shot
    & \num{39.7} & \num{12.5} & \num{25.8} & \num{3.2}
    \\

+ RAG
    & \num{73.8} & \num{19.3} & \num{18.4} & \num{2.9}
    \\

\midrule

$\Delta$
    & $+34.1$ & $+6.8$ & $-7.4$ & $-0.3$
    \\

\bottomrule
\end{tabular*}
\caption{Retrieval augmentation for Qwen3-4B-2507 under zero-shot prompting on 1,000 Basic Education personas. Pass rates are percentages; $\Delta$ is the change in percentage points from Zero-shot to Zero-shot + RAG.}
\label{tab:rag-baseline}
\end{table}

%% file: sec/050analysis.tex
\section{Analysis}
\label{sec:analysis}
We investigate three research questions through within-model comparisons under the same language and curriculum conditions. \textbf{RQ1} (\hyperref[sec:no-mastery]{Sec.~5.1}) tests how explicit mastery information affects personalization. \textbf{RQ2} (\hyperref[sec:book-noise]{Sec.~5.2}) evaluates whether models can focus on the relevant curriculum structure when noisy resources are present. \textbf{RQ3} (\hyperref[sec:one-shot-plan]{Sec.~5.3}) compares step-by-step interaction with generating a static curriculum in one pass.

\subsection{Necessity of Explicit Knowledge State Modeling}
\label{sec:no-mastery}

We constructed a comparative setting by removing the \textit{Mastery} field from the learner persona while keeping all other configurations constant (prompt template in Figure~\ref{fig:rq1}). This ablation turns the input into a coarse learner description. The model still sees the learning goal and curriculum context, but no longer knows which units the learner has actually mastered. It therefore isolates whether explicit knowledge-state modeling contributes to KC planning beyond generic curriculum sequencing.

\input{tables/support1}

Table~\ref{tab:support1} presents the results for Basic Education; Higher Education results are in Appendix~\ref{sec:appendix-Experiment}. Removing mastery information reduces Adaptivity by 26.1 percentage points for Qwen3-30B-A3B (CoT), while Validity changes much less. With the goal and curriculum held fixed, the larger change in Adaptivity shows the role of the supplied mastery state in learner-specific content selection. The model needs this information to decide what a learner should skip, review, or study next.

\subsection{Robustness Against Contextual Noise}
\label{sec:book-noise}

In the main experiments, the candidate pool contains only domain-relevant textbooks. However, real-world retrieval systems can introduce noisy, out-of-domain items. We therefore designed a controlled noise-injection experiment in which irrelevant textbook titles were added at varying proportions. Table~\ref{tab:support2} presents performance under this noisy setting in Basic Education; Higher Education results are in Appendix~\ref{sec:appendix-Experiment}.

\input{tables/support2}

Contextual noise degrades performance broadly, most notably in Validity and Efficiency. Qwen3-30B-A3B (CoT) loses 54.9 percentage points in Validity and 35.7 points in Efficiency. Since learners are profiled for the target domain, selecting out-of-domain textbooks can introduce prerequisite violations and divert steps from the target. The planner must select the relevant part of the knowledge structure for the learner's goal.

\subsection{Effects of Dynamic Interaction}
\label{sec:one-shot-plan}
We contrast the step-by-step paradigm with one-shot generation, where the model plans the entire curriculum in a single turn. Table~\ref{tab:support3} presents results for Basic Education; Higher Education results are in Appendix~\ref{sec:appendix-Experiment}.

\input{tables/support3}

Table~\ref{tab:support3} shows a divergence between structural validity and adaptation. Static generation increases Validity by 30.8 percentage points for Qwen3-4B-2507 (CoT), while Adaptivity decreases by 28.8 points for Qwen3-30B-A3B (CoT) and Efficiency by 33.8 points for DeepSeek-R1-Distill-14B. In the interactive setting, each decision can use the mastery changes produced by previous concepts. The one-shot planner fixes its sequence before receiving these updates, reducing its ability to adjust content selection as the learner progresses.

%% file: tables/support1.tex
\setlength\tabcolsep{3pt}

\begin{table}[htbp]
\centering

\resizebox{1.0\linewidth}{!}{
\begin{tabular}{lcccc}
\toprule

\textbf{Model}               & \begin{tabular}[c]{@{}c@{}}Validity\\Pass Rate\end{tabular}
                             & \begin{tabular}[c]{@{}c@{}}Adaptivity\\Pass Rate\end{tabular}
                             & \begin{tabular}[c]{@{}c@{}}Efficiency\\Pass Rate\end{tabular}
                             & \begin{tabular}[c]{@{}c@{}}Final\\Pass Rate\end{tabular}
                             \\
\midrule

\multicolumn{1}{l}{\cellcolor{white}Qwen3-4B-2507$_{CoT}$}
    & \num{56.0}\gup{0.2} & \num{21.1}(\red{$\downarrow$ 22.8}) & \num{50.9}\rd{9.3} & \num{7.4}(\red{$\downarrow$ 13.4})
    \\

\multicolumn{1}{l}{\cellcolor{white}Pangu-7B$_{CoT}$}
    & \num{22.7}\rd{7.2} & \num{16.8}(\red{$\downarrow$ 11.7}) & \num{45.1}\rd{1.3} & \num{1.4}(\red{$\downarrow$ 13.6})
    \\

\multicolumn{1}{l}{\cellcolor{white}DeepSeek-R1-Distill-14B$_{CoT}$}
    & \num{49.8}\rd{0.1} & \num{26.3}(\red{$\downarrow$ 17.1}) & \num{56.2}\rd{0.9} & \num{10.0}(\red{$\downarrow$ 10.4})
    \\

\multicolumn{1}{l}{\cellcolor{white}Qwen3-30B-A3B$_{CoT}$}
    & \num{67.8}\rd{3.0} & \num{18.6}(\red{$\downarrow$ 26.1}) & \num{53.0}\rd{6.4} & \num{8.5}(\red{$\downarrow$ 19.2})
    \\

\bottomrule
\end{tabular}
}
\caption{Performance of LLMs on PersonaPath (Basic Education) with coarse-grained profiles that omit knowledge mastery.}
\label{tab:support1}
\end{table}

%% file: tables/support2.tex
\setlength\tabcolsep{3pt}

\begin{table}[h]
\centering
\resizebox{1.0\linewidth}{!}{
\begin{tabular}{lcccc}
\toprule
    \textbf{Model}           & \begin{tabular}[c]{@{}c@{}}Validity\\Pass Rate\end{tabular}
                             & \begin{tabular}[c]{@{}c@{}}Adaptivity\\Pass Rate\end{tabular}
                             & \begin{tabular}[c]{@{}c@{}}Efficiency\\Pass Rate\end{tabular}
                             & \begin{tabular}[c]{@{}c@{}}Final\\Pass Rate\end{tabular}
                             \\
\midrule

\multicolumn{1}{l}{\cellcolor{white}Qwen3-4B-2507$_{CoT}$}
    & \num{12.6}(\red{$\downarrow$ 43.2}) & \num{34.3}(\red{$\downarrow$ 9.6}) & \num{37.4}(\red{$\downarrow$ 22.8}) & \num{3.5}(\red{$\downarrow$ 17.3})
    \\

\multicolumn{1}{l}{\cellcolor{white}Pangu-7B$_{CoT}$}
    & \num{2.7}(\red{$\downarrow$ 27.2}) & \num{27.7}(\red{$\downarrow$ 0.8}) & \num{22.5}(\red{$\downarrow$ 23.9}) & \num{1.8}(\red{$\downarrow$ 13.2})
    \\

\multicolumn{1}{l}{\cellcolor{white}DeepSeek-R1-Distill-14B$_{CoT}$}
    & \num{12.5}(\red{$\downarrow$ 37.4}) & \num{25.6}(\red{$\downarrow$ 17.8}) & \num{29.1}(\red{$\downarrow$ 28.0}) & \num{4.5}(\red{$\downarrow$ 15.9})
    \\

\multicolumn{1}{l}{\cellcolor{white}Qwen3-30B-A3B$_{CoT}$}
    & \num{15.9}(\red{$\downarrow$ 54.9}) & \num{25.4}(\red{$\downarrow$ 19.3}) & \num{23.7}(\red{$\downarrow$ 35.7}) & \num{6.5}(\red{$\downarrow$ 21.2})
    \\

\bottomrule
\end{tabular}
}
\caption{Performance of LLMs on PersonaPath (Basic Education) in the presence of contextual noise.}
\label{tab:support2}
\end{table}

%% file: tables/support3.tex
\setlength\tabcolsep{3pt}

\begin{table}[h]
\centering
\resizebox{1.0\linewidth}{!}{
\begin{tabular}{lcccc}
\toprule
    \textbf{Model}           & \begin{tabular}[c]{@{}c@{}}Validity\\Pass Rate\end{tabular}
                             & \begin{tabular}[c]{@{}c@{}}Adaptivity\\Pass Rate\end{tabular}
                             & \begin{tabular}[c]{@{}c@{}}Efficiency\\Pass Rate\end{tabular}
                             & \begin{tabular}[c]{@{}c@{}}Final\\Pass Rate\end{tabular}
                             \\
\midrule

\multicolumn{1}{l}{\cellcolor{white}Qwen3-4B-2507$_{CoT}$}
    & \num{86.6}\gup{30.8} & \num{20.6}(\red{$\downarrow$ 23.3}) & \num{38.1}(\red{$\downarrow$ 22.1}) & \num{5.7}(\red{$\downarrow$ 15.1})
    \\

\multicolumn{1}{l}{\cellcolor{white}Pangu-7B$_{CoT}$}
    & \num{48.0}\gup{18.1} & \num{24.0}(\red{$\downarrow$ 4.5}) & \num{16.0}(\red{$\downarrow$ 30.4}) & \num{5.4}(\red{$\downarrow$ 9.6})
    \\

\multicolumn{1}{l}{\cellcolor{white}DeepSeek-R1-Distill-14B$_{CoT}$}
    & \num{77.9}\gup{28.0} & \num{19.9}(\red{$\downarrow$ 23.5}) & \num{23.3}(\red{$\downarrow$ 33.8}) & \num{4.2}(\red{$\downarrow$ 16.2})
    \\

\multicolumn{1}{l}{\cellcolor{white}Qwen3-30B-A3B$_{CoT}$}
    & \num{84.4}\gup{13.6} & \num{15.9}(\red{$\downarrow$ 28.8}) & \num{30.2}(\red{$\downarrow$ 29.2}) & \num{3.6}(\red{$\downarrow$ 24.1})
    \\

\bottomrule
\end{tabular}
}
\caption{Performance of LLMs on PersonaPath (Basic Education) with one-shot path generation.}
\label{tab:support3}
\end{table}

%% file: sec/060conclusion.tex
\section{Conclusion}
We study Knowledge-Centric personalized learning path planning, where models use explicit learner goals, mastery states, and curriculum prerequisites to select and order knowledge. To evaluate this setting, we introduce \textbf{PersonaPath}, a benchmark that pairs 2,000 learner personas with a hierarchical knowledge graph built from 347 textbooks and organized into units and concepts. Expert ratings support the pedagogical plausibility and internal coherence of the sampled personas. Experiments on representative LLMs show that current models still struggle with KC planning, especially in adapting paths to learner states. The ablations identify mastery information and step-by-step feedback as important inputs for personalized planning.

\clearpage
\section*{Limitations}
PersonaPath uses synthetic learner states and mastery updates. Expert evaluation covers the pedagogical plausibility and internal coherence of 100 personas, not the relationship between benchmark scores and real learner outcomes. Longitudinal traces and classroom studies could measure observed progress and teacher interventions, while broader expert evaluation could assess generated paths.

The benchmark is grounded in Chinese textbooks and curricula. Cross-language and cross-curriculum transfer requires adapting the source materials and validating prerequisite structures and simulation settings. Controlled studies should distinguish language and curriculum effects from planning performance; Appendix~\ref{sec:appendix-transfer} describes the adaptation procedure.

\section*{Ethics Statement}
The learner personas used in PersonaPath are synthetic and generated from statistical archetypes (Low-performing, Average-performing, and High-performing) to simulate diverse cognitive states. No real-world student data or Personally Identifiable Information (PII) were collected, stored, or processed in the construction of this benchmark. This design reduces privacy risks associated with real learner data.

The knowledge substrate of our benchmark is derived from 347 textbooks sourced from authoritative platforms, including People's Education Press and the Smart Education of China platform. These materials were used strictly for the purpose of constructing an academic benchmark, extracting knowledge graphs and prerequisite relations. We do not distribute the full raw text of copyrighted books; instead, we release the structured knowledge graphs and derived metadata necessary for reproducing the experiments.

The observed hallucinations and prerequisite violations motivate teacher review of generated learning paths before classroom use. Educational applications should retain teacher oversight of content selection and learning goals.

%% file: sec/070appendix.tex
\appendix

\setcounter{table}{0}
\renewcommand\thetable{\Alph{section}.\arabic{table}}
\setcounter{figure}{0}
\renewcommand\thefigure{\Alph{section}.\arabic{figure}}

\section{Benchmark Details}
\label{sec:appendix-Benchmark}
\input{sec/071sim}

\input{sec/074dataset_comparison}

\subsection{Data Collection}
\input{sec/073data_collection}

\subsection{Human Verification and Compensation}
As detailed in Section~\ref{sec:quality}, we recruited domain experts to verify prerequisite relations and content correctness in the knowledge space. These experts were selected based on their academic background in the respective disciplines. All annotators were informed of the purpose of the data before the task and were compensated at an hourly rate above the local minimum wage.
\subsubsection{Instructions for Prerequisite Relation Validation}
Experts were presented with candidate pairs of concepts $(C_i, C_j)$ and were instructed to validate the directed link $C_i \to C_j$ based on the following criteria:

\begin{itemize}
    \item \textbf{Logical Precedence:} ``Does concept $C_i$ provide the foundational knowledge required to understand $C_j$?'' (Binary: Yes/No)
    \item \textbf{Pedagogical Alignment:} ``Is the proposed sequence consistent with the standard teaching order found in the source textbooks?'' (Binary: Yes/No)
    \item \textbf{Action:} Mark the link as \textit{Invalid} if it creates a logical cycle or violates pedagogical norms.
\end{itemize}

\subsubsection{Instructions for Content Correctness Audit}
For the entities and semantic descriptions generated by the LLM, experts were provided with the source textbooks as ground truth. The specific instructions were:

\begin{enumerate}
    \item \textbf{Fact Verification:} ``Verify that the definition and properties of the generated entity align strictly with the textbook content. Mark any factual deviations.''
    \item \textbf{Hallucination Check:} ``Flag any terms or relations that do not exist in the domain context or appear to be fabricated by the model.''
    \item \textbf{Terminology Standard:} ``Ensure that the technical terminology used is appropriate for the target educational stage.''
\end{enumerate}

\subsection{Hierarchical Transition Mechanism}
\label{sec:hierarchical-transition}
The agent operates through a three-level hierarchical transition mechanism:

\textbf{Textbook Transition.} Once all units within the current textbook achieve the mastery threshold $\tau$, the agent consults the prerequisite graph to determine the next textbook. This is a graph-level decision that considers the learner's global mastery state rather than a fixed linear sequence.

\textbf{Unit Transition.} Within a given textbook, the agent traverses units sequentially. It advances to the next unit only when the current unit's aggregated mastery reaches the target threshold $\tau$.

\textbf{Concept Transition.} Within an unmastered unit, the agent selects the next concept by matching the unit's current mastery level against each candidate concept's difficulty parameter. This selection mechanism is designed to target the learner's Zone of Proximal Development (ZPD) \cite{obukhova2009zone}, ensuring that the chosen concept is neither trivially easy nor prohibitively difficult.

\section{Experiment Details}
\label{sec:appendix-Experiment}

\subsection{Constraints}
\label{app:constraints}
\input{tables/constraint_description}

Table~\ref{tab:constraints} provides detailed definitions for the three categories of constraints (Validity, Adaptivity, and Efficiency) introduced in Section~\ref{sec:evaluation}. Together, these constraints assess whether an agent can produce a pedagogically appropriate learning sequence under the benchmark setting.

\noindent \textbf{Validity Constraints.}
Validity evaluates whether a generated path is structurally and factually sound within the provided curriculum. The \textit{Prerequisite Violation} constraint enforces the cumulative nature of learning. Following Knowledge Space Theory \citep{doignon2012knowledge}, meaningful learning depends on a structured surmise relation, where advanced concepts should not be introduced before their antecedent concepts have been sufficiently mastered. The \textit{Hallucination} constraint safeguards factual accuracy by prohibiting fictitious educational entities, such as nonexistent textbooks, units, or concepts outside the provided knowledge space. Such entities are a documented failure mode of LLMs \citep{ji2023survey}.

\noindent \textbf{Adaptivity Constraints.}
Adaptivity evaluates whether the path fits the specific learner rather than a generic curriculum order. The \textit{Persona Alignment} constraint is motivated by the Zone of Proximal Development (ZPD) \citep{vygotsky1978mind, obukhova2009zone}: a pedagogically appropriate path should remain consistent with the learner's cognitive archetype, prior learning experience, and behavioral profile. The \textit{Difficulty Adaptability} constraint uses Cog-Gap (Eq.~\ref{eq:cog-gap}) to compare current unit mastery with the difficulty of the selected concept. Concept difficulty follows the Item Response Theory parametrization~\citep{cheng2019enhancing}.

\noindent \textbf{Efficiency Constraints.}
Efficiency evaluates whether the path reaches the learning goal within practical bounds. The \textit{Goal Completion} constraint requires the agent to achieve the target mastery level within a maximum allowable number of steps, in line with step-bounded goal-achievement evaluation for LLM planning \citep{valmeekam2023planbench}. The maximum step limit $T_{\max}$ is set in proportion to the number of concepts in the target stage, so longer target stages receive a larger but still bounded planning horizon. The \textit{Progress Continuity} constraint prevents learning stagnation by requiring the average proficiency increment over a sliding window ($n=5$) to exceed a predefined threshold. This criterion is also pedagogically motivated: sustained mastery experiences help protect learner \textit{self-efficacy} \citep{bandura1977self}, whereas long stretches of negligible progress indicate inefficient or poorly sequenced instruction.

\noindent \textbf{Final Pass Rate.}
Following the multi-constraint evaluation protocol \citep{jiang2024followbench}, a learning path passes the evaluation only if it satisfies all three dimensions simultaneously. We define the final pass indicator as $Pass_{final} = \mathbb{I}(Validity) \cdot \mathbb{I}(Adaptivity) \cdot \mathbb{I}(Efficiency)$, where $\mathbb{I}(\cdot)$ outputs 1 when the corresponding dimension is fully satisfied and 0 otherwise. The Final Pass Rate is the proportion of generated paths whose $Pass_{final}$ equals 1.

\subsection{Overall Performance}
DeepSeek-V3.1 achieves the highest final pass rates in our experiments, yet reaches only 29.5\% and 14.6\% in Basic and Higher Education, respectively. These low absolute scores suggest that PersonaPath is challenging for current LLMs in KC learning path planning.
We observe that model performance generally improves with larger model sizes. For instance, Pangu-7B (CoT) achieves a final pass rate of 15.0\% in Basic Education, whereas the corresponding Pangu-1B (CoT) reaches only 1.1\%.
However, increasing model size does not guarantee improvements across all constraints, echoing inverse-scaling observations in which larger models can underperform on certain tasks \citep{mckenzie2023inverse}. Qwen3-4B-2507 (CoT) (60.2\%) slightly surpasses Qwen3-30B-A3B (CoT) (59.4\%) on the \textit{Efficiency} constraint, showing that larger scale alone does not guarantee higher Efficiency.

\subsection{Additional Experimental Results}
\label{sec:sup_expr}
This section provides additional experimental results for the analyses in Section~\ref{sec:analysis}. Specifically, Table~\ref{tab:sup1max} presents the results for Section~\ref{sec:no-mastery}, Table~\ref{tab:sup2max} presents the results for Section~\ref{sec:book-noise}, and Table~\ref{tab:sup3max} presents the results for Section~\ref{sec:one-shot-plan}.

Table~\ref{tab:sup1max} presents LLM performance under this coarse-grained setting in both Basic and Higher Education. Compared with the full-profile baseline in the main text, \textit{Validity} changes much less than Adaptivity. Qwen3-4B-2507 (CoT) shows a marginal improvement (+0.2\%) in Validity.

We attribute this pattern to the lower cognitive load. Complex mastery constraints typically require models to deviate from standard curricular paths to accommodate specific gaps. Removing these constraints allows the models, especially smaller ones, to revert to canonical teaching sequences, which are frequent in pretraining data and often logically valid. This suggests that fine-grained modeling is important for personalization but not necessarily for generic logical correctness.

\input{tables/support_1_max}
\input{tables/support_2_max}
\input{tables/support_3_max}

\subsection{Retrieval Configuration}
\label{sec:appendix-rag}

To investigate whether retrieval-augmented generation (RAG) can improve KC learning path planning, we augment the base model Qwen3-4B-2507$_{Zero-shot}$ with a RAG pipeline and evaluate it in the Basic Education setting. Specifically, we use \texttt{BAAI/bge-small-zh-v1.5} as the embedding model with L2-normalized vectors and FAISS with inner-product search (equivalent to cosine similarity) for retrieval. For each query, the top-3 most relevant knowledge chunks are retrieved and prepended to the prompt as reference context.

Table~\ref{tab:rag-baseline} in the main text reports all four evaluation dimensions for this comparison.

\section{Cross-Language and Cross-Curriculum Transferability}
\label{sec:appendix-transfer}
While our current instantiation uses Chinese educational textbooks, the structural framework of PersonaPath can in principle be adapted to other languages and curricula. The hierarchical knowledge graph construction pipeline, the IRT-based student simulator, and the three evaluation dimensions (Validity, Adaptivity, Efficiency) are not tied to a specific language, but they would require curriculum-specific calibration and validation. To adapt PersonaPath to an English-language curriculum such as the US K-12 Common Core, a researcher would (1) replace the source corpus with standard English textbooks, (2) employ an LLM to extract the Textbook-Unit-Concept hierarchy from the new corpus, and (3) validate the prerequisite dependencies with domain experts. The remaining pipeline, including student simulation, path planning, and evaluation, could then be reused after such adaptation.

\section{Knowledge-Level Approaches}
\label{sec:appendix-knowledge-level}
This section supplements Section~\ref{sec:related_work} with the relationship between learner-state estimation and knowledge-path planning. Knowledge Tracing and Cognitive Diagnosis characterize what a learner currently knows, providing state estimates that can inform a planning component.

Knowledge Tracing (KT) estimates and dynamically updates a learner's mastery over each knowledge component from interaction sequences. Early probabilistic models such as Bayesian Knowledge Tracing track each skill with a small set of latent parameters \citep{corbett1995knowledge}, and later deep models improve predictive accuracy with recurrent, memory, and attention architectures \citep{piech2015deep, zhang2017dynamic, pandey2019self}. Cognitive Diagnosis infers a learner's latent proficiency on fine-grained concepts from response data \citep{cheng2019enhancing, lord2012applications}. A system combining KT/CD with a curriculum planner can use these estimates to select subsequent content. PersonaPath evaluates the planning component with an explicitly supplied mastery vector: the output is an ordered knowledge path, assessed for prerequisite validity, learner adaptation, and goal-directed efficiency. Applying a mastery estimator and a planner together would additionally involve aligning their state representations and specifying the planning policy.

\input{sec/075error.tex}

\section{Prerequisite Graph Presentation}

We illustrate the prerequisite graphs generated for the collected textbooks. As formalized in Section~\ref{sec:data_construction}, these directed acyclic structures specify the order in which agents must navigate textbook-level resources.

\input{figures/workflow.tex}

\input{figures/sequence/ch_base}
\input{figures/sequence/english_base}
\input{figures/sequence/physics_base}
\input{figures/sequence/chem_base}
\input{figures/sequence/bio_base}
\input{figures/sequence/geo_base}
\input{figures/sequence/history_base}
\input{figures/sequence/mora_base}
\input{figures/sequence/music_base}
\input{figures/sequence/it_base}
\input{figures/sequence/art_base}
\input{figures/sequence/pe_base}

\input{figures/sequence/law_pro}
\input{figures/sequence/gong_pro}

\section{Prompt Templates}
\label{sec:appendix-prompt}
\input{figures/prompt/gen_kg}

\input{figures/prompt/l1_cot}
\input{figures/prompt/l1_0shot}

\input{figures/prompt/l3_cot}
\input{figures/prompt/l3_0shot}

\input{figures/prompt/ab_prompt}
\input{figures/prompt/fb_prompt}
\input{figures/prompt/Global_Planner_Agent}

%% file: sec/071sim.tex
\subsection{Cognitive State Transition Dynamics}
\label{sec:appendix_sim}

This section provides the formal definition of the environment transition function $\text{Sim}(\mathbf{M}_t[u], D(a_t))$, which simulates the dynamic update of a learner's proficiency based on their interaction with an atomic concept $a_t$. Let $u \in \mathcal{U}$ be the parent unit of concept $a_t$. The transition consists of two stochastic phases: \textit{Performance Simulation} and \textit{Mastery Gain Computation}.

\subsubsection{Performance Simulation (IRT Model)}
We model the probability of a learner successfully mastering concept $a_t$ using the logistic IRT model~\citep{lord2012applications}. Given the learner's current unit mastery $\mathbf{M}_t[u]$ and the concept's inherent difficulty $D(a_t)$, the probability of success $P(Y_t=1)$ is defined as:

\begin{equation}
    P(Y_t=1 | \mathbf{M}_t, a_t) = \frac{1}{1 + \exp\left( -k \cdot \Delta_{\text{gap}} \right)}
\end{equation}
\noindent where:
\begin{itemize}
    \item $\Delta_{\text{gap}} = \mathbf{M}_t[u] - D(a_t) + \delta_{\mathbf{P}}$ denotes the effective ability gap adjusted by the learner's persona.
    \item $Y_t \in \{0, 1\}$ is the binary outcome of the interaction (1 for success, 0 for failure).
    \item $k$ is the discrimination parameter, controlling the sensitivity of the outcome to the ability-difficulty gap.
    \item $\delta_{\mathbf{P}}$ is a bias term derived from the learner's static persona $\mathbf{P}$.
\end{itemize}

\subsubsection{Difficulty-Aware Gain Function}
The core of the transition dynamics is the \textit{Difficulty-Aware Gain Function}. Unlike constant-gain models, we posit that the magnitude of proficiency improvement $\Delta m_t$ depends on both the complexity of the task and the learner's relative performance.

\textbf{Case 1: Successful Interaction ($Y_t = 1$).}
A successful learning event yields a gain proportional to the concept's difficulty. We also incorporate a \textbf{Challenge Bonus} based on the Zone of Proximal Development (ZPD) theory~\citep{obukhova2009zone}: if a learner successfully masters a concept harder than their current proficiency ($D(a_t) > \mathbf{M}_t[u]$), the gain is amplified.

\begin{equation}
\begin{split}
    \Delta m_t^{+} = & \underbrace{(\lambda_1 D(a_t) + \lambda_0)}_{\text{\scriptsize Info. Utility}} \cdot \underbrace{\left( \frac{\eta_{\mathbf{P}}}{\eta_{\text{ref}}} \right)}_{\text{\scriptsize Persona Scaling}} \\
    & \cdot \underbrace{\left( 1 + \gamma \cdot \mathbb{I}(D(a_t) > \mathbf{M}_t[u]) \right)}_{\text{\scriptsize Breakthrough Bonus}}
\end{split}
\end{equation}

\noindent where $\lambda_1$ and $\lambda_0$ are difficulty utility coefficients, $\eta_{\mathbf{P}}$ is the individual learning rate from persona $\mathbf{P}$, and $\gamma$ is the bonus coefficient for cognitive breakthroughs.

\textbf{Case 2: Failed Interaction ($Y_t = 0$).}
Failure results in a marginal gain, simulating the ``familiarity effect'' derived from trial and error:
\begin{equation}
    \Delta m_t^{-} = \eta_{\mathbf{P}} \cdot \epsilon
\end{equation}

\subsubsection{State Update Rule}
Finally, the unit-level mastery vector is updated for the next time step $t+1$:
\begin{equation}
    \mathbf{M}_{t+1}[u] = \min\left( \mathbf{M}_t[u] + \Delta m_t, 1.0 \right)
\end{equation}
where $\Delta m_t$ is determined by the sampled outcome $Y_t$.
\input{tables/app_sim}

\subsubsection{Sensitivity Analysis of Simulation Hyperparameters}
\label{sec:sensitivity}

The cognitive state transition function (Equations 2--5) relies on a set of hyperparameters (Table~A.1) whose values were selected empirically. To evaluate the sensitivity of our conclusions to these values, we varied each parameter individually while holding the others at their default values and measured the effect on learning trajectories across all three proficiency archetypes (Low-performing, Average-performing, and High-performing). We performed 1,000 Monte Carlo runs per configuration.

\noindent \textbf{Parameter Selection Rationale.} The hyperparameters in our simulation were selected based on established principles from Item Response Theory and educational modeling:

\begin{itemize}
    \item \textbf{Discrimination parameter ($k$):} Following IRT conventions~\citep{lord2012applications}, $k$ is set within $[3.0, 5.0]$. A lower $k$ yields poor discrimination because success probabilities flatten near 50\% regardless of ability, while an excessively high $k$ creates an unrealistic step function, causing abrupt pass/fail outcomes based on marginal ability-difficulty differences.

    \item \textbf{Difficulty utility coefficients ($\lambda_1, \lambda_0$):} These govern the base mastery gain per successful interaction. Their values are calibrated so that an average learner requires approximately 3--5 interactions to master a unit, reflecting realistic learning trajectories observed in educational practice.

    \item \textbf{Challenge bonus ($\gamma$):} We set $\gamma$ within $[0.1, 0.2]$ to regulate the extra cognitive gain from succeeding at tasks above the learner's current proficiency. This range accelerates learning for high-ability students while preventing extreme, unstable mastery spikes caused by lucky guesses.

    \item \textbf{Failure compensation ($\epsilon$):} This small constant simulates the ``familiarity effect'' from trial and error. Its value ensures that failed interactions still contribute marginal progress without undermining the distinction between success and failure.
\end{itemize}

\begin{figure*}[t]
    \centering
    \includegraphics[width=\textwidth]{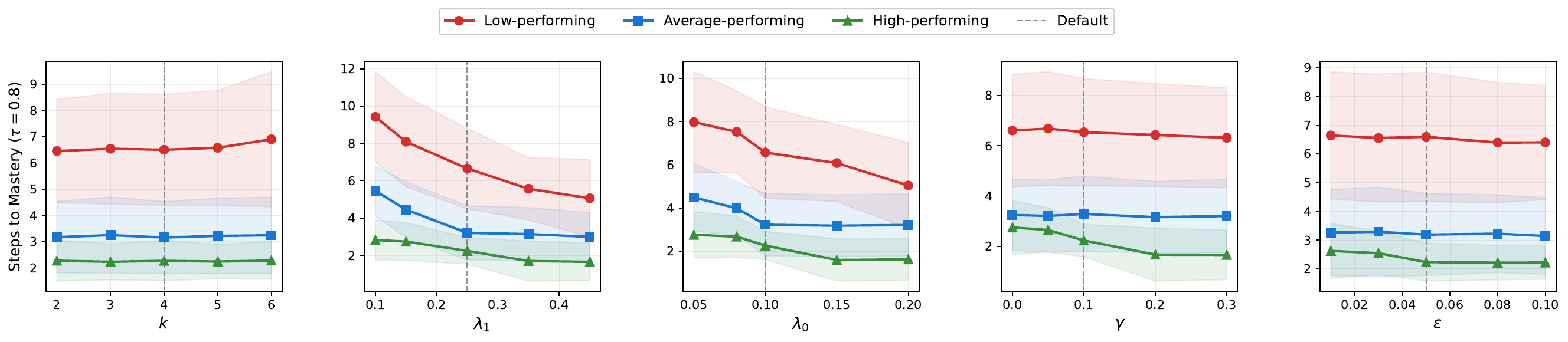}
    \caption{Mean steps to mastery ($\tau = 0.8$) as each hyperparameter is varied individually. Shaded regions denote $\pm 1$ standard deviation across 1,000 runs. Dashed vertical lines indicate default values. The relative ordering among proficiency archetypes is preserved across all configurations.}
    \label{fig:sa_steps}
\end{figure*}

\noindent \textbf{Impact on Learning Speed.} Figure~\ref{fig:sa_steps} presents the mean steps to mastery ($\tau = 0.8$) as each parameter is varied across a wide range. Two key observations emerge:

\textbf{(1) Parameters primarily affect learning speed, not relative ordering.} Across all 25 parameter configurations tested, the archetype ordering (Low-performing $>$ Average-performing $>$ High-performing in required steps) is preserved with 100\% consistency, as shown in Table~\ref{tab:sensitivity_ordering}. This indicates that parameter variations uniformly scale the difficulty of the environment without distorting the distinction between learner archetypes.

\textbf{(2) Parameter sensitivity varies substantially.} We quantify each parameter's influence using the Coefficient of Variation (CV) of mean steps across its tested range. As shown in Table~\ref{tab:sensitivity_cv}, the gain coefficients $\lambda_1$ (CV\,$=$\,0.234) and $\lambda_0$ (CV\,$=$\,0.177) exhibit the highest sensitivity, as they directly control the magnitude of mastery increments per interaction. In contrast, the discrimination parameter $k$ (CV\,$=$\,0.015) and the failure compensation $\epsilon$ (CV\,$=$\,0.036) have negligible impact on overall learning speed, since they modulate probabilities and marginal gains rather than the primary mastery update pathway.

\begin{table}[t]
\centering
\small
\begin{tabular}{lccccc}
\toprule
 & $k$ & $\lambda_1$ & $\lambda_0$ & $\gamma$ & $\epsilon$ \\
\midrule
CV & 0.015 & 0.234 & 0.177 & 0.081 & 0.036 \\
\bottomrule
\end{tabular}
\caption{Coefficient of Variation (CV) of mean steps to mastery across parameter ranges. Higher CV indicates greater sensitivity.}
\label{tab:sensitivity_cv}
\end{table}

\noindent \textbf{Preservation of Archetype Ordering.} Table~\ref{tab:sensitivity_ordering} provides detailed results for the three most representative parameters ($k$, $\lambda_1$, and $\gamma$). Across all 15 configurations, the step ratio between adjacent archetypes (Low/Average and Average/High) remains strictly above 1.0, confirming that no parameter setting causes the archetype ordering to invert. Even for the most sensitive parameter, $\lambda_1$, where the mean steps range from 5.1 to 9.4 for Low-performing learners, the ordering Low-performing $>$ Average-performing $>$ High-performing is preserved.

\begin{table}[t]
\centering
\small

\begin{tabular}{clccccc}
\toprule
\textbf{Param} & \textbf{Value} & \textbf{Low-perf.} & \textbf{Avg-perf.} & \textbf{High-perf.} & \textbf{Order} \\
\midrule
\multirow{5}{*}{$k$}
 & 2.0 & 6.5\,{\scriptsize$\pm$\,2.0} & 3.2\,{\scriptsize$\pm$\,1.4} & 2.3\,{\scriptsize$\pm$\,0.8} & \checkmark \\
 & 3.0 & 6.5\,{\scriptsize$\pm$\,2.1} & 3.3\,{\scriptsize$\pm$\,1.5} & 2.2\,{\scriptsize$\pm$\,0.7} & \checkmark \\
 & 4.0$^\ast$ & 6.5\,{\scriptsize$\pm$\,2.1} & 3.2\,{\scriptsize$\pm$\,1.4} & 2.3\,{\scriptsize$\pm$\,0.8} & \checkmark \\
 & 5.0 & 6.6\,{\scriptsize$\pm$\,2.2} & 3.2\,{\scriptsize$\pm$\,1.4} & 2.2\,{\scriptsize$\pm$\,0.7} & \checkmark \\
 & 6.0 & 6.9\,{\scriptsize$\pm$\,2.6} & 3.2\,{\scriptsize$\pm$\,1.5} & 2.3\,{\scriptsize$\pm$\,0.7} & \checkmark \\
\midrule
\multirow{5}{*}{$\lambda_1$}
 & 0.10 & 9.4\,{\scriptsize$\pm$\,2.4} & 5.5\,{\scriptsize$\pm$\,1.3} & 2.8\,{\scriptsize$\pm$\,1.1} & \checkmark \\
 & 0.15 & 8.1\,{\scriptsize$\pm$\,2.4} & 4.5\,{\scriptsize$\pm$\,1.5} & 2.7\,{\scriptsize$\pm$\,1.0} & \checkmark \\
 & 0.25$^\ast$ & 6.7\,{\scriptsize$\pm$\,2.2} & 3.2\,{\scriptsize$\pm$\,1.5} & 2.2\,{\scriptsize$\pm$\,0.7} & \checkmark \\
 & 0.35 & 5.6\,{\scriptsize$\pm$\,1.7} & 3.1\,{\scriptsize$\pm$\,1.4} & 1.7\,{\scriptsize$\pm$\,1.1} & \checkmark \\
 & 0.45 & 5.1\,{\scriptsize$\pm$\,2.1} & 3.0\,{\scriptsize$\pm$\,1.3} & 1.7\,{\scriptsize$\pm$\,1.0} & \checkmark \\
\midrule
\multirow{5}{*}{$\gamma$}
 & 0.00 & 6.6\,{\scriptsize$\pm$\,2.2} & 3.2\,{\scriptsize$\pm$\,1.4} & 2.8\,{\scriptsize$\pm$\,1.1} & \checkmark \\
 & 0.05 & 6.7\,{\scriptsize$\pm$\,2.3} & 3.2\,{\scriptsize$\pm$\,1.4} & 2.7\,{\scriptsize$\pm$\,0.9} & \checkmark \\
 & 0.10$^\ast$ & 6.5\,{\scriptsize$\pm$\,2.1} & 3.3\,{\scriptsize$\pm$\,1.5} & 2.2\,{\scriptsize$\pm$\,0.6} & \checkmark \\
 & 0.20 & 6.4\,{\scriptsize$\pm$\,2.0} & 3.2\,{\scriptsize$\pm$\,1.4} & 1.7\,{\scriptsize$\pm$\,1.1} & \checkmark \\
 & 0.30 & 6.3\,{\scriptsize$\pm$\,2.0} & 3.2\,{\scriptsize$\pm$\,1.5} & 1.7\,{\scriptsize$\pm$\,1.0} & \checkmark \\
\bottomrule
\end{tabular}
\caption{Mean steps to mastery under varied parameter settings. ``$\ast$'' marks the default configuration. The archetype ordering is preserved across all 25 tested configurations (\checkmark).}
\label{tab:sensitivity_ordering}
\end{table}

\subsubsection{Implications for Evaluation Validity}

The sensitivity analysis measures simulated learning speed and the ordering of proficiency archetypes. Across the tested settings, the archetype ordering remains stable while the number of steps to mastery varies. These results characterize the behavior of the learner simulator under parameter changes.

In closed-loop planning, simulation parameters affect the mastery states supplied to subsequent decisions. Although the constraint definitions remain fixed, changes in the state trajectory can alter concept selection, Cog-Gap, and goal completion. The main model comparisons therefore use a common simulator configuration; evaluating model rankings under alternative configurations would require rerunning the planners with those dynamics.

%% file: tables/app_sim.tex
\begin{table}[h]
\centering
\label{tab:sim_params}
\small
\begin{tabular}{l l l}
\hline
\textbf{Symbol} & \textbf{Value} & \textbf{Description} \\ \hline
$k$ & 4.0 & IRT discrimination factor \\
$\lambda_1, \lambda_0$ & 0.25, 0.1 & Coefficients for difficulty-based reward \\
$\eta_{\text{ref}}$ & 0.15 & Reference learning rate for normalization \\
$\gamma$ & 0.1 & Challenge bonus factor (10\%) \\
$\epsilon$ & 0.05 & Failure compensation factor \\
\hline
\end{tabular}
\caption{Hyperparameters used in the simulation function.}
\end{table}

%% file: sec/074dataset_comparison.tex
\subsection{Dataset Comparison}
\label{ref:dataset_comparison}

Existing benchmarks for evaluating educational planning, including Junyi Academy \cite{chang2015modeling}, ASSISTments 2009/2012 \cite{feng2009addressing, wang2015towards}, OLI Engineering Statics \cite{Fall-2011-OLI-Data}, and synthetic data from DKT \cite{piech2015deep}, primarily support \textbf{Exercise-Centric (EC)} evaluation.
These benchmarks are valuable for localized item recommendation and student performance prediction, but they provide limited support for evaluating whether a model can construct a goal-oriented curriculum path from explicit learner states and prerequisite knowledge structures.
\noindent\textbf{Existing Benchmarks in Detail.} The \textbf{ASSISTments} family \citep{feng2009addressing, wang2015towards} is widely used for student performance prediction. The 2009 version focuses on correctness logs, while the 2012 version additionally incorporates a predicted affective state of the student. The \textbf{Junyi Academy} dataset \citep{chang2015modeling} provides extensive records of student attempts, hints, and time spent across topics ranging from arithmetic to geometry. The \textbf{OLI Engineering Statics} dataset \citep{Fall-2011-OLI-Data} contains 189,297 trials from a college-level engineering statics course. Synthetic datasets, such as those generated for Deep Knowledge Tracing (DKT) \citep{piech2015deep}, simulate virtual student responses using Item Response Theory to support model testing. Despite their differences in scale and domain, all of these resources record item-level interactions rather than the explicit knowledge structures and learner goals required for KC path planning.

PersonaPath instantiates a \textbf{Knowledge-Centric (KC)} planning benchmark, described by the following five dimensions.

\input{tables/DBH}
\label{app:data_comp}

\subsubsection{Textbook-level Prerequisite Graph}
\textbf{Textbook-level Prerequisite Graph (TPG)} describes dependencies among textbooks and supports the evaluation of prerequisite-aware planning across a curriculum. Item-level skill annotations and exercise relations in existing datasets operate at different granularities of learner modeling. PersonaPath provides textbook-level dependencies together with the nested units and concepts needed to evaluate paths across these resources.

We provide visualizations of these TPG structures across various disciplines in PersonaPath: Figures~\ref{fig:ch_math} through \ref{fig:pe} illustrate the prerequisite graphs for 13 subjects in Basic Education, while Figures~\ref{fig:law} and \ref{fig:gong} depict professional learning paths for Higher Education. Taking the ``Mechanics'' major within the Engineering category as an example (the teal path in Fig.~\ref{fig:gong}), the TPG serves as a benchmark constraint: the LLM agent is expected to identify and follow a learning sequence that covers fundamental Mathematics and Physics before Theoretical Mechanics and Material Mechanics. By introducing TPG as a benchmark requirement, we provide a structured way to evaluate whether LLMs can follow curriculum-level planning constraints, a capability that is difficult to assess using prior, less-structured datasets.

\subsubsection{Persona}
\textbf{Persona (PSN)} describes the learner information available to a planner. PersonaPath constructs explicit profiles with \textbf{Proficiency Archetypes} (Low-performing, Average-performing, and High-performing) in an approximate 1:3:1 ratio and \textbf{Academic Ambitions} represented by target majors that define long-term goals. \textbf{Temporal Context} establishes a current grade level and partitions the knowledge base into already learned, currently learning, and yet-to-be-learned content. These attributes accompany the mastery state, giving the planner both a starting point and a target. Interaction datasets such as ASSISTments and Junyi instead support estimating learner states from recorded behavior.

\subsubsection{Breadth of Educational Disciplines}
\textbf{Breadth of Educational Disciplines (BED)} defines the evaluation boundaries of the benchmark, covering both general knowledge and higher-level professional expertise. As detailed in Table~\ref{tab:discipline_comparison}, PersonaPath organizes 77 subjects. In contrast, existing benchmarks are more limited in scope; for example, Junyi Academy and ASSISTments focus mainly on K-12 Mathematics, while OLI Engineering Statics is confined to a single specialized course.

\subsubsection{Hierarchical Knowledge Structure}
\textbf{Hierarchical Knowledge Structure (HKS)} establishes a structured framework beyond the flat skill labeling found in traditional datasets. While previous benchmarks often represent knowledge as an unstructured collection of independent skills, PersonaPath formalizes the knowledge space into a three-tiered nested architecture: \textbf{Textbooks}, \textbf{Units}, and \textbf{Atomic Concepts}. This hierarchy, encompassing 347 textbooks, 1,751 units, and 4,092 concepts, approximates the structural organization of institutional education.

\subsubsection{Rich Semantics}
\textbf{Rich Semantics (RS)} measures whether a benchmark provides informative semantic details rather than only textual labels. In PersonaPath, every textbook, unit, and concept is enriched with semantic content, including definitions, learning objectives, and pedagogical descriptions. Most existing datasets are collections of exercise records or problem sets in which knowledge nodes are often reduced to opaque IDs or brief strings. This lack of semantic depth makes it difficult to evaluate whether models can reason over educational content when planning a path.

%% file: tables/DBH.tex
\begin{table*}[htbp]
\centering
\renewcommand{\tabularxcolumn}[1]{m{#1}}
\renewcommand{\arraystretch}{2}
\small

\begin{tabularx}{\textwidth}{| >{\hsize=0.5\hsize\bfseries\centering\arraybackslash}X | >{\hsize=1.5\hsize\raggedright\arraybackslash}X |}
\hline
\rowcolor{gray!25} \textbf{Dataset Name} & \multicolumn{1}{c|}{\textbf{Disciplines Covered}} \\ \hline

\makecell{Junyi Academy \\ \cite{chang2015modeling}} & Mathematics, Biology, Computer Science (basic education). \\ \hline

\makecell{ASSISTments2009 \\ \cite{feng2009addressing}} & Mathematics (basic education). \\ \hline

\makecell{ASSISTments2012 \\ \cite{wang2015towards}} & Mathematics (basic education). \\ \hline

\makecell{OLI Engineering Statics \\ \cite{Fall-2011-OLI-Data}} & Engineering Statics (higher education). \\ \hline

\textbf{PersonaPath (Ours)} &
Chinese, Mathematics, English, Physics, Chemistry, Biology, History, Geography, Morality and the Rule of Law, Information Technology, Music, Art, P.E. (basic education). \par \vspace{0.5ex}
Law, Engineering, Management, Education, Economics, Science, History, Agronomy, Literature, Medicine, Arts, Philosophy (higher education). \\ \hline

\end{tabularx}
\caption{Comparison of discipline coverage across datasets.}
\label{tab:discipline_comparison}
\end{table*}

%% file: sec/073data_collection.tex
\subsubsection{Rationality of Data Source Selection}

\begin{figure*}[t]
    \centering
    \includegraphics[width=\linewidth]{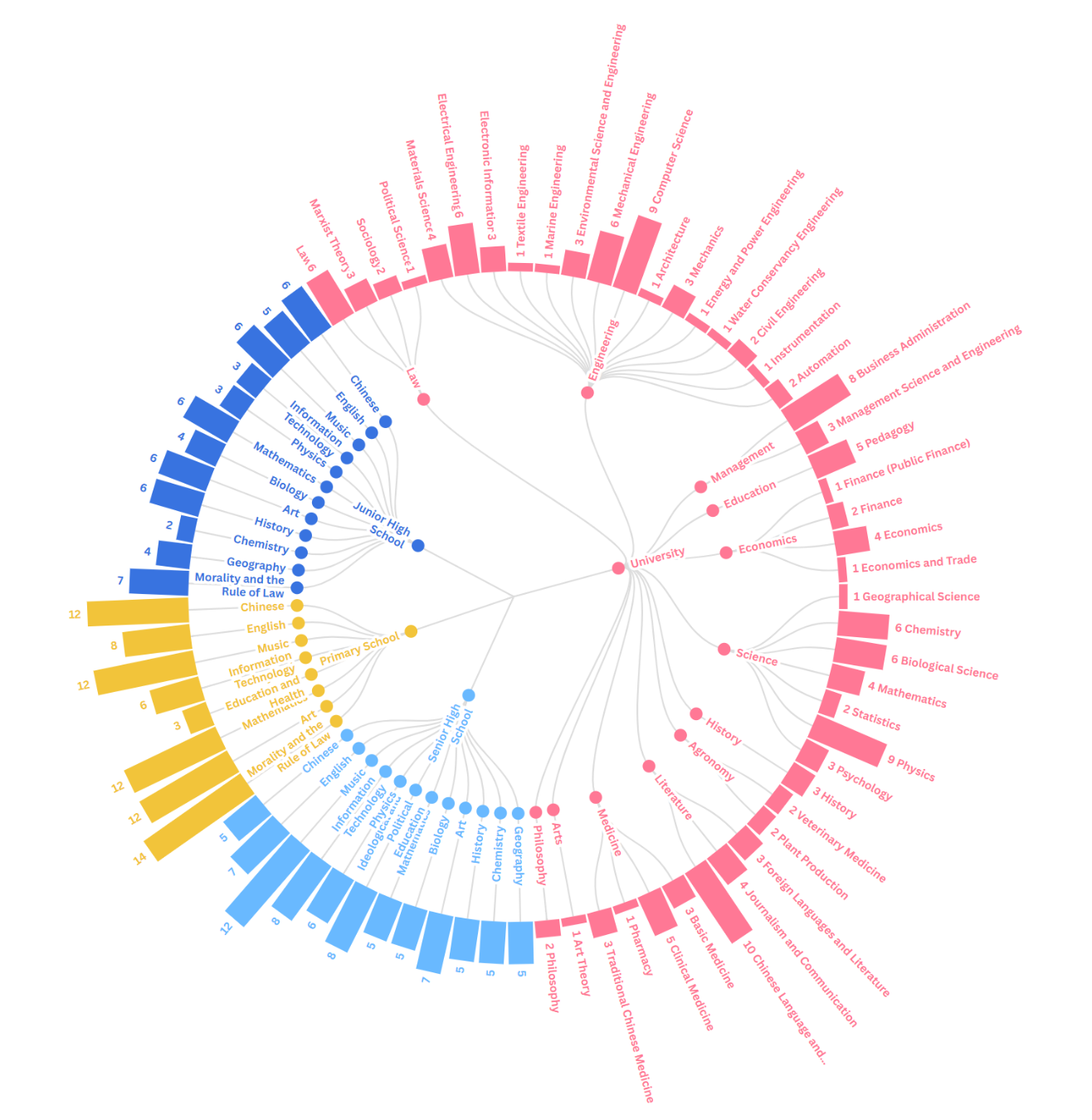}
    \caption{Overview of the disciplines in PersonaPath.}
    \label{fig:category_map}
\end{figure*}

To support the pedagogical validity and broad applicability of PersonaPath, our data collection strategy followed these principles:

\textbf{Authoritative Provenance.} We prioritized reliability in educational content. For Basic Education, materials were sourced from the People's Education Press, a widely used publisher for Chinese K-12 curriculum. For Higher Education, resources were curated from the Smart Education of China platform to align with national academic standards. This selection strategy reduces the noise often found in crowdsourced open-web educational data.

\textbf{Holistic Curriculum Continuity.} A key rationale for our dataset construction was to bridge the gap between disjointed educational stages. Unlike existing datasets that focus on isolated exercises, we constructed a dependency graph that explicitly links Basic Education subjects (e.g., Biology) as prerequisites for Higher Education majors (e.g., Medicine). This structural design allows for the evaluation of long-horizon, cross-stage curriculum planning capabilities.

\textbf{Representative Learner Modeling.} We use synthetic personas derived from statistical archetypes (Low-performing, Average-performing, and High-performing) to simulate a controlled yet diverse range of cognitive states. This approach provides a balanced distribution for difficulty-adaptability tests (an approximate 1:3:1 ratio), which is often difficult to obtain from skewed real-world interaction logs.

\subsubsection{Compliance and Ethical Standards}
Our data construction and release process follows legal and ethical guidelines regarding intellectual property, privacy, and labor rights:

\textbf{Intellectual Property and Copyright Compliance.} We strictly distinguish between the raw content of textbooks and the derived knowledge structures. \textit{Usage:} The textbooks served solely as the source for extracting knowledge graphs, prerequisite relations, and concept hierarchies.
\textit{Distribution:} To comply with copyright laws, we do not distribute, host, or reproduce the full raw text of the copyrighted books. The public release of PersonaPath is limited to the structured knowledge graphs (metadata), relationship triplets, and the synthetic persona profiles. This constitutes a transformative use of the data for academic research purposes.

\textbf{Privacy and Human Subjects Exemption.}
The learner profiles used in the benchmark are synthetic. No real-world student data or Personally Identifiable Information (PII) were collected, stored, or processed.
Since no real learner data are used, the benchmark is designed to avoid privacy risks associated with real student records.
For the human-in-the-loop verification process involving domain experts, we followed ethical labor standards. All annotators were informed of the purpose of the data beforehand and were compensated at an hourly rate above the local minimum wage.

%% file: tables/constraint_description.tex
\begin{table*}[t]
\centering
\small
\renewcommand{\arraystretch}{1.4}

\begin{tabularx}{\linewidth}{@{} l X @{}}
\toprule
\textbf{Constraint} & \textbf{Description} \\
\midrule

\rowcolor[gray]{0.92}
\multicolumn{2}{c}{\textit{\textbf{Validity Constraints}}} \\

Prerequisite Violation & Occurs when a learning item is recommended before its necessary foundational prerequisites have been mastered by the learner. \\
\addlinespace[0.5em]

Hallucination & The generation of spurious or non-existent educational entities, such as fictitious textbooks or concepts. \\

\midrule

\rowcolor[gray]{0.92}
\multicolumn{2}{c}{\textit{\textbf{Adaptivity Constraints}}} \\

Difficulty Adaptability & Quantifies the discrepancy between a learner's current mastery level and the intrinsic difficulty of the concept to learn. \\
\addlinespace[0.5em]

Persona Alignment & The degree of consistency between the planning strategy and the learner's explicit cognitive profile. \\

\midrule

\rowcolor[gray]{0.92}
\multicolumn{2}{c}{\textit{\textbf{Efficiency Constraints}}} \\

Goal Completion & Indicates whether the learning task is completed within the prescribed step limit. \\
\addlinespace[0.5em]

Progress Continuity & Mandates that the average proficiency increment calculated over a sliding window ($n=5$) must exceed a predefined threshold to ensure sustained learning momentum. \\

\bottomrule
\end{tabularx}
\caption{Description of constraints used in the evaluation.}
\label{tab:constraints}
\end{table*}

%% file: tables/support_1_max.tex
\begin{table*}[htbp]
\centering
\setlength\tabcolsep{3pt}

\resizebox{1.0\linewidth}{!}{
\begin{tabular}{lcccccccc}
\toprule
 & \multicolumn{4}{c}{\textbf{Basic Education} (\#\num{1000})}
 & \multicolumn{4}{c}{\textbf{Higher Education} (\#\num{1000})}
 \\
\cmidrule(l){2-5} \cmidrule(l){6-9}
 & \begin{tabular}[c]{@{}c@{}}Validity\\Pass Rate\end{tabular}
 & \begin{tabular}[c]{@{}c@{}}Adaptivity\\Pass Rate\end{tabular}
 & \begin{tabular}[c]{@{}c@{}}Efficiency\\Pass Rate\end{tabular}
 & \begin{tabular}[c]{@{}c@{}}Final\\Pass Rate\end{tabular}
 & \begin{tabular}[c]{@{}c@{}}Validity\\Pass Rate\end{tabular}
 & \begin{tabular}[c]{@{}c@{}}Adaptivity\\Pass Rate\end{tabular}
 & \begin{tabular}[c]{@{}c@{}}Efficiency\\Pass Rate\end{tabular}
 & \begin{tabular}[c]{@{}c@{}}Final\\Pass Rate\end{tabular}
 \\
\midrule

\multicolumn{1}{l}{\cellcolor{white}Qwen3-4B-2507$_{CoT}$}
    & \num{56.0} & \num{21.1} & \num{50.9} & \num{7.4}
    & \num{26.7} & \num{25.0} & \num{29.2} & \num{5.7}
    \\

\multicolumn{1}{l}{\cellcolor{white}Pangu-7B$_{CoT}$}
    & \num{22.7} & \num{16.8} & \num{45.1} & \num{1.4}
    & \num{20.7} & \num{35.6} & \num{40.7} & \num{3.6}
    \\

\multicolumn{1}{l}{\cellcolor{white}Llama-3.1-8B$_{CoT}$}
    & \num{9.6} & \num{6.8} & \num{7.7} & \num{0.8}
    & \num{3.2} & \num{12.0} & \num{3.1} & \num{0.4}
    \\

\multicolumn{1}{l}{\cellcolor{white}DeepSeek-R1-Distill-14B$_{CoT}$}
    & \num{49.8} & \num{26.3} & \num{56.2} & \num{10.0}
    & \num{37.0} & \num{32.9} & \num{43.0} & \num{7.7}
    \\

\multicolumn{1}{l}{\cellcolor{white}Qwen3-30B-A3B$_{CoT}$}
    & \num{67.8} & \num{18.6} & \num{53.0} & \num{8.5}
    & \num{44.36} & \num{32.3} & \num{45.3} & \num{9.5}
    \\
\bottomrule
\end{tabular}
}
\caption{Performance of LLMs on PersonaPath with coarse-grained profiles that omit knowledge mastery.}
\label{tab:sup1max}

\end{table*}

%% file: tables/support_2_max.tex
\begin{table*}[htbp]
\centering
\setlength\tabcolsep{3pt}
\resizebox{1.0\linewidth}{!}{
\begin{tabular}{lcccccccc}
\toprule
 & \multicolumn{4}{c}{\textbf{Basic Education} (\#\num{1000})}
 & \multicolumn{4}{c}{\textbf{Higher Education} (\#\num{1000})}
 \\
\cmidrule(l){2-5} \cmidrule(l){6-9}
 & \begin{tabular}[c]{@{}c@{}}Validity\\Pass Rate\end{tabular}
 & \begin{tabular}[c]{@{}c@{}}Adaptivity\\Pass Rate\end{tabular}
 & \begin{tabular}[c]{@{}c@{}}Efficiency\\Pass Rate\end{tabular}
 & \begin{tabular}[c]{@{}c@{}}Final\\Pass Rate\end{tabular}
 & \begin{tabular}[c]{@{}c@{}}Validity\\Pass Rate\end{tabular}
 & \begin{tabular}[c]{@{}c@{}}Adaptivity\\Pass Rate\end{tabular}
 & \begin{tabular}[c]{@{}c@{}}Efficiency\\Pass Rate\end{tabular}
 & \begin{tabular}[c]{@{}c@{}}Final\\Pass Rate\end{tabular}
 \\
\midrule

\multicolumn{1}{l}{\cellcolor{white}Qwen3-4B-2507$_{CoT}$}
    & \num{12.6} & \num{34.3} & \num{37.4} & \num{3.5}
    & \num{7.3} & \num{43.5} & \num{12.7} & \num{1.5}
    \\

\multicolumn{1}{l}{\cellcolor{white}Pangu-7B$_{CoT}$}
    & \num{2.7} & \num{27.7} & \num{22.5} & \num{1.8}
    & \num{2.7} & \num{37.4} & \num{7.1} & \num{0.1}
    \\

\multicolumn{1}{l}{\cellcolor{white}Llama-3.1-8B$_{CoT}$}
    & \num{9.6} & \num{6.8} & \num{7.7} & \num{0.9}
    & \num{2.5} & \num{30.0} & \num{1.5} & \num{0.0}
    \\

\multicolumn{1}{l}{\cellcolor{white}DeepSeek-R1-Distill-14B$_{CoT}$}
    & \num{12.5} & \num{25.6} & \num{29.1} & \num{4.5}
    & \num{3.9} & \num{34.4} & \num{12.5} & \num{0.8}
    \\

\multicolumn{1}{l}{\cellcolor{white}Qwen3-30B-A3B$_{CoT}$}
    & \num{15.9} & \num{25.4} & \num{23.7} & \num{6.5}
    & \num{3.6} & \num{36.3} & \num{3.5} & \num{0.2}
    \\
\bottomrule
\end{tabular}
}
\caption{Performance of LLMs on PersonaPath in the presence of contextual noise.}
\label{tab:sup2max}
\end{table*}

%% file: tables/support_3_max.tex
\begin{table*}[htbp]
\centering

\setlength\tabcolsep{3pt}
\resizebox{1.0\linewidth}{!}{
\begin{tabular}{lcccccccc}
\toprule
 & \multicolumn{4}{c}{\textbf{Basic Education} (\#\num{1000})}
 & \multicolumn{4}{c}{\textbf{Higher Education} (\#\num{1000})}
 \\
\cmidrule(l){2-5} \cmidrule(l){6-9}
 & \begin{tabular}[c]{@{}c@{}}Validity\\Pass Rate\end{tabular}
 & \begin{tabular}[c]{@{}c@{}}Adaptivity\\Pass Rate\end{tabular}
 & \begin{tabular}[c]{@{}c@{}}Efficiency\\Pass Rate\end{tabular}
 & \begin{tabular}[c]{@{}c@{}}Final\\Pass Rate\end{tabular}
 & \begin{tabular}[c]{@{}c@{}}Validity\\Pass Rate\end{tabular}
 & \begin{tabular}[c]{@{}c@{}}Adaptivity\\Pass Rate\end{tabular}
 & \begin{tabular}[c]{@{}c@{}}Efficiency\\Pass Rate\end{tabular}
 & \begin{tabular}[c]{@{}c@{}}Final\\Pass Rate\end{tabular}
 \\
\midrule

\multicolumn{1}{l}{\cellcolor{white}Qwen3-4B-2507$_{CoT}$}
    & \num{86.6} & \num{20.6} & \num{38.1} & \num{5.7}
    & \num{52.4} & \num{30.0} & \num{15.5} & \num{4.9}
    \\

\multicolumn{1}{l}{\cellcolor{white}Pangu-7B$_{CoT}$}
    & \num{48.0} & \num{24.0} & \num{16.0} & \num{5.4}
    & \num{32.4} & \num{21.5} & \num{9.7} & \num{3.6}
    \\

\multicolumn{1}{l}{\cellcolor{white}Llama-3.1-8B$_{CoT}$}
    & \num{39.9} & \num{4.9} & \num{6.2} & \num{0.5}
    & \num{28.1} & \num{9.3} & \num{7.1} & \num{0.5}
    \\

\multicolumn{1}{l}{\cellcolor{white}DeepSeek-R1-Distill-14B$_{CoT}$}
    & \num{77.9} & \num{19.9} & \num{23.3} & \num{4.2}
    & \num{57.4} & \num{24.9} & \num{20.2} & \num{4.6}
    \\

\multicolumn{1}{l}{\cellcolor{white}Qwen3-30B-A3B$_{CoT}$}
    & \num{84.4} & \num{15.9} & \num{30.2} & \num{3.6}
    & \num{53.3} & \num{23.3} & \num{12.7} & \num{3.6}
    \\
\bottomrule
\end{tabular}
}
\caption{Performance of LLMs on PersonaPath with one-shot path generation.}
\label{tab:sup3max}

\end{table*}

%% file: sec/075error.tex
\section{Error Case Analysis}
\label{sec:appendix-error-cases}

To provide an intuitive understanding of how learning paths can violate evaluation constraints, we construct three representative error cases based on the same persona and target: an \textit{Average-performing} learner targeting \textit{Math-Primary-Grade4B-Four Operations}. Keeping the learner profile and goal fixed allows the cases to isolate different failure modes in the generated path.

\textbf{Case 1: Validity failure.}
The first case is a prerequisite violation. During the first two rounds, the model correctly traverses prerequisite content in \textit{Math-Primary-Grade3B}. However, at Round~3, the learner still has three unmastered units in Grade-3B, and \textit{Math-Primary-Grade4A} has not been learned at all. Instead of continuing through these prerequisites, the model jumps directly to the target textbook \textit{Math-Primary-Grade4B}. This exposes the learner to ``Four Operations'' before the necessary foundations in decimal arithmetic, multi-digit multiplication, and Grade-4A content have been established.

\textbf{Case 2: Adaptivity failure.}
The second case shows a mismatch between concept difficulty and learner mastery. After Round~1, the learner's mastery of the current unit reaches $0.75$, so a suitable next concept should be closer to this level. Instead, the model selects \textit{Oral Division Basics} with difficulty $d=0.1$, yielding a large Cog-Gap of $0.65$. The same pattern appears again in later rounds: the model repeatedly chooses the easiest available concept even when a moderately harder concept would better match the learner's current mastery. For an \textit{Average-performing} learner, such choices provide little cognitive challenge and waste learning opportunities.

\textbf{Case 3: Efficiency failure.}
The third case illustrates inefficient planning caused by repeated over-challenging selections. The model consistently selects the hardest available concept within each unit. For example, at Round~1 the learner's mastery is $0.53$, but the model assigns a concept with difficulty $d=0.4$. This choice yields only a small mastery gain, whereas a better matched concept with $d=0.2$ would move the learner much faster toward the unit threshold. Because the model keeps repeating this pattern, progress continuity breaks down and the path consumes most of the step budget before entering later prerequisites. By Round~20, the learner remains in Grade-3B and never reaches the target textbook.

Figures~\ref{fig:case-validity}--\ref{fig:case-efficiency} show the corresponding model-generated outputs.

\input{tables/badcase_detail}

%% file: tables/badcase_detail.tex
\begin{figure*}[htbp]
\begin{tcolorbox}[
  base={Case 1: Validity Violation \textcolor{red}{\ding{55}} -- Model Output (Partial)},
  colback=white,
  breakable
]
\begin{lstlisting}[style=promptstyle, escapeinside={(*@}{@*)}]
// Round 1-2: Correct prerequisite traversal in Grade-3B
{ "round": 1, "book": "Math-Primary-Grade3B",
  "unit": "Division by One-Digit Divisors",
  "sub_node": "Pen Division",
  "mastery_old": 0.5255, "mastery_new": 0.7522, "difficulty": 0.2,
  "book_reason": "This textbook is the most fundamental
    prerequisite for the target textbook" }

{ "round": 2, "book": "Math-Primary-Grade3B",
  "unit": "Two-Digit by Two-Digit Multiplication",
  "sub_node": "Pen Multiplication Rules",
  "mastery_old": 0.4437, "mastery_new": 0.7270, "difficulty": 0.3,
  "book_reason": "This textbook is the most fundamental
    prerequisite for the target textbook" }

// Round 3: PREREQUISITE VIOLATION
// (*@\hlcons{State: Grade-3B still has 3 unmastered units; Grade-4A is entirely unmastered.}@*)
// (*@\hlfor{Expected: continue Grade-3B, then enter Grade-4A before Grade-4B.}@*)
{ "round": 3,
  "book": "(*@\hllogic{Math-Primary-Grade4B}@*)",
  "unit": "Four Operations",
  "sub_node": "Operation Rules Application",
  "mastery_old": 0.4951, "mastery_new": 0.7218, "difficulty": 0.2,
  "book_reason": "The student has basically mastered
    the prerequisite content and can start the target textbook",
  "diagnosis": "(*@\hllogic{Invalid jump: prerequisite textbooks are skipped.}@*)",
  "concept_reason": "Operation Rules Application with difficulty
    0.2 has the smallest gap to mastery 0.4951" }
\end{lstlisting}
\end{tcolorbox}
\caption{Case 1: Validity violation. The orange highlight marks the invalid jump to the target textbook; blue marks the unmet prerequisite state; purple marks the expected prerequisite traversal.}
\label{fig:case-validity}
\end{figure*}

\begin{figure*}[htbp]
\begin{tcolorbox}[
  base={Case 2: Adaptivity Violation \textcolor{red}{\ding{55}} -- Model Output (Partial)},
  colback=white,
  breakable
]
\begin{lstlisting}[style=promptstyle, escapeinside={(*@}{@*)}]
// Round 1: Reasonable selection (Cog-Gap = |0.53 - 0.2| = 0.33)
{ "round": 1, "book": "Math-Primary-Grade3B",
  "unit": "Division by One-Digit Divisors",
  "sub_node": "Pen Division",
  "mastery_old": 0.5255, "mastery_new": 0.7522, "difficulty": 0.2,
  "concept_reason": "Pen Division with difficulty 0.2 has
    the smallest gap to mastery 0.5255" }

// Round 2: ADAPTIVITY VIOLATION
// (*@\hlcons{Mastery = 0.7522, selected d = 0.1, Cog-Gap = 0.6522.}@*)
// (*@\hlfor{Better match: choose a harder available concept, e.g., d = 0.2.}@*)
{ "round": 2, "book": "Math-Primary-Grade3B",
  "unit": "Division by One-Digit Divisors",
  "sub_node": "(*@\hllogic{Oral Division Basics}@*)",
  "mastery_old": 0.7522, "mastery_new": 0.9788,
  "difficulty": (*@\hllogic{0.1}@*),
  "diagnosis": "(*@\hllogic{Too easy for the learner's current mastery.}@*)",
  "concept_reason": "Oral Division Basics with difficulty 0.1
    is the simplest concept, suitable for consolidation" }

// Round 3: ADAPTIVITY VIOLATION
// (*@\hlcons{Mastery = 0.6486, selected d = 0.1, Cog-Gap = 0.5486.}@*)
// (*@\hlfor{Better match: available d = 0.4 gives a smaller gap of 0.2486.}@*)
{ "round": 3, "book": "Math-Primary-Grade3B",
  "unit": "Year, Month and Day",
  "sub_node": "(*@\hllogic{Calendar Basic Knowledge}@*)",
  "mastery_old": 0.6486, "mastery_new": 0.6571,
  "difficulty": (*@\hllogic{0.1}@*),
  "diagnosis": "(*@\hllogic{The model again chooses the easiest concept.}@*)",
  "concept_reason": "Select the most basic concept
    to help the student build a solid foundation" }

// Round 5: ADAPTIVITY VIOLATION
// (*@\hlcons{Mastery = 0.4783, selected d = 0.1, Cog-Gap = 0.3783.}@*)
// (*@\hlfor{Better match: Decimal Arithmetic with d = 0.2.}@*)
{ "round": 5, "book": "Math-Primary-Grade3B",
  "unit": "Introduction to Decimals",
  "sub_node": "(*@\hllogic{Decimal Basic Knowledge}@*)",
  "mastery_old": 0.4783, "mastery_new": 0.4868,
  "difficulty": (*@\hllogic{0.1}@*),
  "diagnosis": "(*@\hllogic{Repeated under-challenge inflates Cog-Gap.}@*)",
  "concept_reason": "Decimal Basic Knowledge with difficulty 0.1
    is the simplest, suitable for introductory learning" }
\end{lstlisting}
\end{tcolorbox}
\caption{Case 2: Adaptivity violation. Orange highlights repeated under-challenging selections; blue shows the learner mastery and resulting Cog-Gap; purple indicates the better matched alternatives.}
\label{fig:case-adaptivity}
\end{figure*}

\begin{figure*}[htbp]
\begin{tcolorbox}[
  base={Case 3: Efficiency Violation \textcolor{red}{\ding{55}} -- Model Output (Partial)},
  colback=white,
  breakable
]
\begin{lstlisting}[style=promptstyle, escapeinside={(*@}{@*)}]
// Round 1: EFFICIENCY VIOLATION
// (*@\hlcons{Mastery = 0.5255, but the model selects the hardest concept d = 0.4.}@*)
// (*@\hlfor{Better match: Pen Division with d = 0.2 would yield much faster progress.}@*)
{ "round": 1, "book": "Math-Primary-Grade3B",
  "unit": "Division by One-Digit Divisors",
  "sub_node": "(*@\hllogic{Division Word Problems}@*)",
  "mastery_old": 0.5255, "mastery_new": (*@\hllogic{0.5340}@*),
  "difficulty": (*@\hllogic{0.4}@*),
  "diagnosis": "(*@\hllogic{Only +0.0085 mastery gain; progress is too slow.}@*)",
  "concept_reason": "Division Word Problems is the most
    challenging concept, helps the student aim higher" }

// Rounds 2-6: The model keeps choosing the same hard concept.
// (*@\hlcons{Each round produces roughly the same tiny gain; the unit stays far below 0.8.}@*)
{ "round": 2, ... "sub_node": "(*@\hllogic{Division Word Problems}@*)",
  "mastery_old": 0.5340, "mastery_new": (*@\hllogic{0.5425}@*),
  "difficulty": (*@\hllogic{0.4}@*) }
{ "round": 3, ... "mastery_old": 0.5425, "mastery_new": 0.5510 ... }
{ "round": 4, ... "mastery_old": 0.5510, "mastery_new": 0.5595 ... }
{ "round": 5, ... "mastery_old": 0.5595, "mastery_new": 0.5680 ... }
{ "round": 6, ... "mastery_old": 0.5680, "mastery_new": 0.5765 ... }
// (*@\hlfor{Reference: the correct path reaches 0.8 in 2 rounds with d = 0.2.}@*)

// Round 15: The first unit finally crosses 0.8.
// (*@\hlcons{Fifteen rounds are consumed by one unit; later prerequisites lose the step budget.}@*)
{ "round": 15, "book": "Math-Primary-Grade3B",
  "unit": "Division by One-Digit Divisors",
  "sub_node": "(*@\hllogic{Division Word Problems}@*)",
  "mastery_old": 0.7935, "mastery_new": 0.8020, "difficulty": 0.4,
  "concept_reason": "Continue practicing the hardest concept
    to fully master division" }

// Round 16: Same pattern in the next unit.
{ "round": 16, "book": "Math-Primary-Grade3B",
  "unit": "Two-Digit by Two-Digit Multiplication",
  "sub_node": "(*@\hllogic{Multiplication Word Problems}@*)",
  "mastery_old": 0.4437, "mastery_new": 0.4522,
  "difficulty": (*@\hllogic{0.4}@*),
  "diagnosis": "(*@\hllogic{Over-challenging selection repeats in a new unit.}@*)",
  "concept_reason": "Select the most challenging multiplication
    concept to push the student's limits" }

// Round 20: GOAL NOT REACHED
// (*@\hlcons{Still in Grade-3B; Grade-4A and the target textbook are never reached.}@*)
{ "round": 20, "book": "(*@\hllogic{Math-Primary-Grade3B}@*)",
  "unit": "Two-Digit by Two-Digit Multiplication",
  "sub_node": "(*@\hllogic{Multiplication Word Problems}@*)",
  "mastery_old": 0.4862, "mastery_new": 0.4947,
  "difficulty": (*@\hllogic{0.4}@*),
  "diagnosis": "(*@\hllogic{Step budget exhausted before reaching the target.}@*)" }
\end{lstlisting}
\end{tcolorbox}
\caption{Case 3: Efficiency violation. Orange marks over-challenging repeated choices and slow mastery updates; blue marks the accumulated progress problem; purple marks the faster reference choice.}
\label{fig:case-efficiency}
\end{figure*}

%% file: figures/workflow.tex
\begin{figure*}[t!]
\centering
\includegraphics[width=\textwidth]{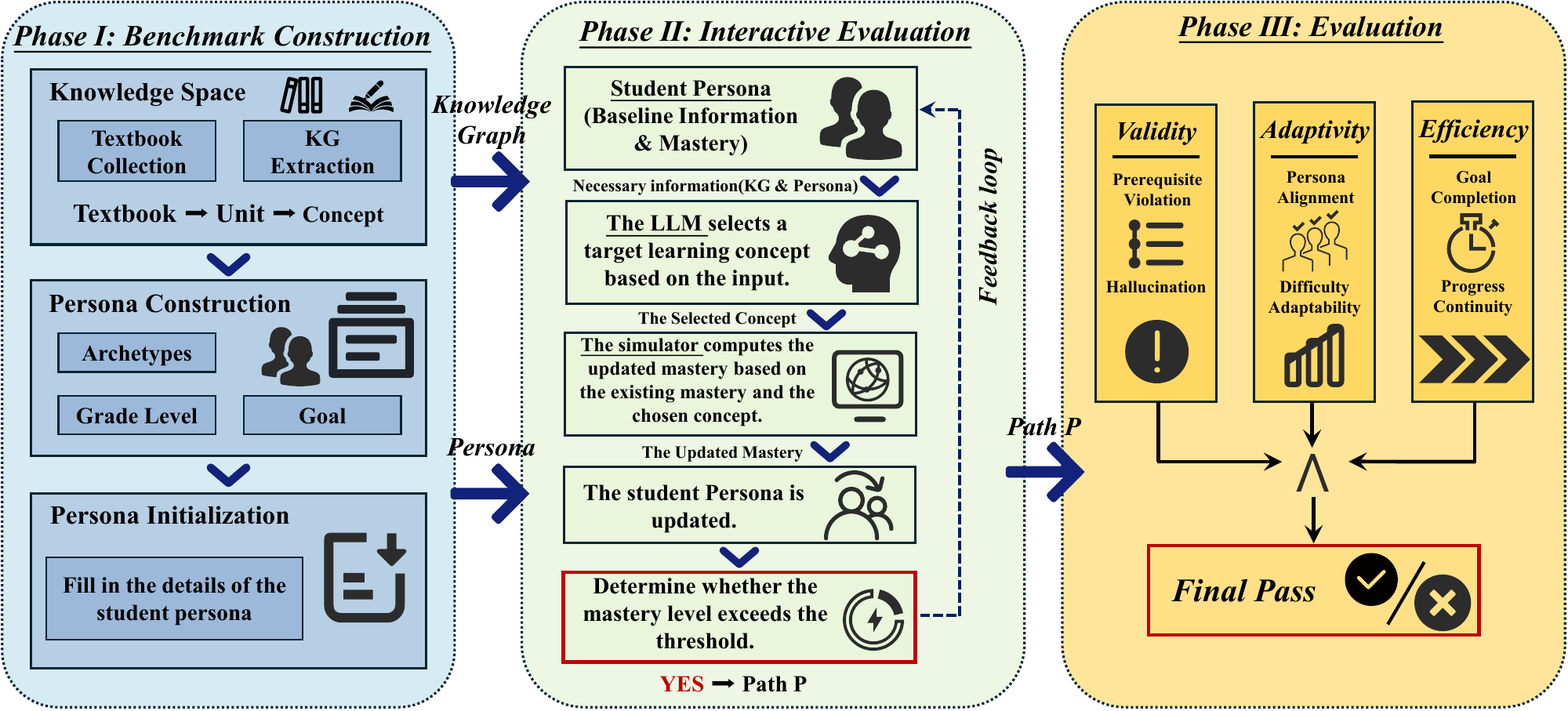}
\caption{Overview of the entire \textbf{PersonaPath} workflow.
\textbf{Phase~I}: A hierarchical knowledge graph and diverse learner personas are constructed from authoritative textbooks.
\textbf{Phase~II}: The LLM agent interacts with the environment in a step-by-step loop, selecting concepts and receiving mastery updates until the target proficiency is reached.
\textbf{Phase~III}: The generated learning path is evaluated across three constraint dimensions (Validity, Adaptivity, and Efficiency), whose conjunction determines the Final Pass Rate.}
\label{fig:workflow}
\end{figure*}

%% file: figures/sequence/ch_base.tex
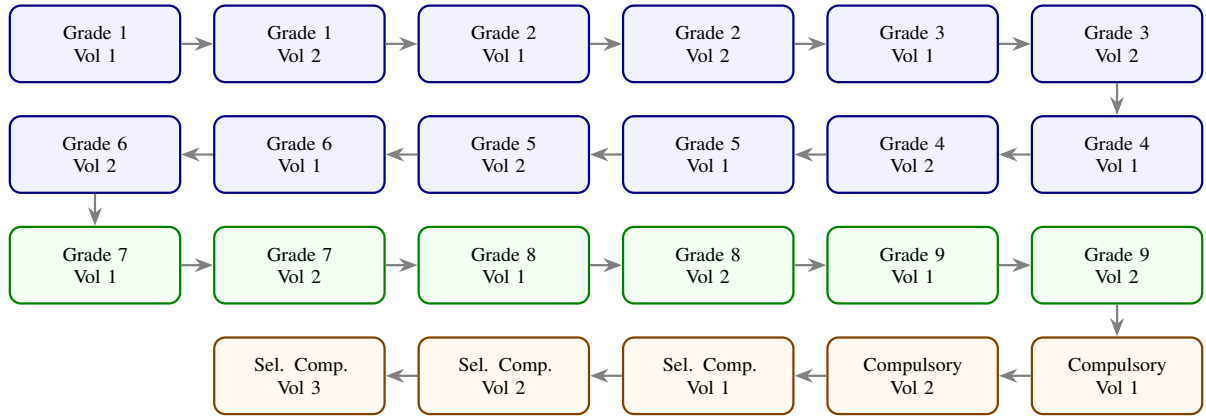
\begin{figure*}[htbp]
    \centering
    \resizebox{\linewidth}{!}{
    \begin{tikzpicture}[
        node distance=0.4cm and 0.4cm,
        box/.style={
            rectangle,
            draw,
            thick,
            rounded corners,
            minimum height=2.5em,
            text width=2.0cm,
            font=\scriptsize,
            align=center,
            inner sep=2pt
        },
        primary/.style={box, draw=blue!50!black, fill=blue!5},
        junior/.style={box, draw=green!50!black, fill=green!5},
        senior/.style={box, draw=orange!50!black, fill=orange!5},
        arrow/.style={-{Stealth[scale=1.0]}, thick, gray}
    ]

    \node[primary] (n1) {Grade 1 \\ Vol 1};
    \node[primary, right=of n1] (n2) {Grade 1 \\ Vol 2};
    \node[primary, right=of n2] (n3) {Grade 2 \\ Vol 1};
    \node[primary, right=of n3] (n4) {Grade 2 \\ Vol 2};
    \node[primary, right=of n4] (n5) {Grade 3 \\ Vol 1};
    \node[primary, right=of n5] (n6) {Grade 3 \\ Vol 2};

    \node[primary, below=of n6] (n7) {Grade 4 \\ Vol 1};
    \node[primary, left=of n7] (n8) {Grade 4 \\ Vol 2};
    \node[primary, left=of n8] (n9) {Grade 5 \\ Vol 1};
    \node[primary, left=of n9] (n10) {Grade 5 \\ Vol 2};
    \node[primary, left=of n10] (n11) {Grade 6 \\ Vol 1};
    \node[primary, left=of n11] (n12) {Grade 6 \\ Vol 2};

    \node[junior, below=of n12] (n13) {Grade 7 \\ Vol 1};
    \node[junior, right=of n13] (n14) {Grade 7 \\ Vol 2};
    \node[junior, right=of n14] (n15) {Grade 8 \\ Vol 1};
    \node[junior, right=of n15] (n16) {Grade 8 \\ Vol 2};
    \node[junior, right=of n16] (n17) {Grade 9 \\ Vol 1};
    \node[junior, right=of n17] (n18) {Grade 9 \\ Vol 2};

    \node[senior, below=of n18] (n19) {Compulsory \\ Vol 1};
    \node[senior, left=of n19] (n20) {Compulsory \\ Vol 2};
    \node[senior, left=of n20] (n21) {Sel. Comp. \\ Vol 1};
    \node[senior, left=of n21] (n22) {Sel. Comp. \\ Vol 2};
    \node[senior, left=of n22] (n23) {Sel. Comp. \\ Vol 3};

    \draw[arrow] (n1) -- (n2); \draw[arrow] (n2) -- (n3); \draw[arrow] (n3) -- (n4);
    \draw[arrow] (n4) -- (n5); \draw[arrow] (n5) -- (n6);
    \draw[arrow] (n6.south) -- (n7.north);
    \draw[arrow] (n7) -- (n8); \draw[arrow] (n8) -- (n9); \draw[arrow] (n9) -- (n10);
    \draw[arrow] (n10) -- (n11); \draw[arrow] (n11) -- (n12);
    \draw[arrow] (n12.south) -- (n13.north);
    \draw[arrow] (n13) -- (n14); \draw[arrow] (n14) -- (n15); \draw[arrow] (n15) -- (n16);
    \draw[arrow] (n16) -- (n17); \draw[arrow] (n17) -- (n18);
    \draw[arrow] (n18.south) -- (n19.north);
    \draw[arrow] (n19) -- (n20); \draw[arrow] (n20) -- (n21); \draw[arrow] (n21) -- (n22);
    \draw[arrow] (n22) -- (n23);

    \end{tikzpicture}
    }
    \caption{The learning sequences for Chinese and Mathematics in basic education.}
    \label{fig:ch_math}
\end{figure*}

%% file: figures/sequence/english_base.tex
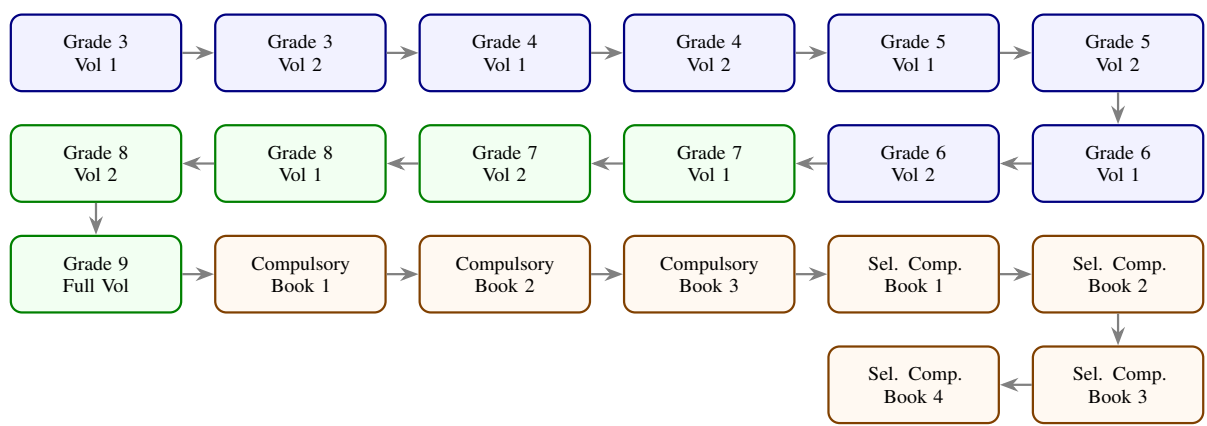
\begin{figure*}[htbp]
    \centering
    \resizebox{\linewidth}{!}{
    \begin{tikzpicture}[
        node distance=0.4cm and 0.4cm,
        box/.style={
            rectangle,
            draw,
            thick,
            rounded corners,
            minimum height=2.5em,
            text width=2.0cm,
            font=\scriptsize,
            align=center,
            inner sep=2pt
        },
        primary/.style={box, draw=blue!50!black, fill=blue!5},
        junior/.style={box, draw=green!50!black, fill=green!5},
        senior/.style={box, draw=orange!50!black, fill=orange!5},
        arrow/.style={-{Stealth[scale=1.0]}, thick, gray}
    ]

    \node[primary] (n1) {Grade 3 \\ Vol 1};
    \node[primary, right=of n1] (n2) {Grade 3 \\ Vol 2};
    \node[primary, right=of n2] (n3) {Grade 4 \\ Vol 1};
    \node[primary, right=of n3] (n4) {Grade 4 \\ Vol 2};
    \node[primary, right=of n4] (n5) {Grade 5 \\ Vol 1};
    \node[primary, right=of n5] (n6) {Grade 5 \\ Vol 2};

    \node[primary, below=of n6] (n7) {Grade 6 \\ Vol 1};
    \node[primary, left=of n7] (n8) {Grade 6 \\ Vol 2};
    \node[junior, left=of n8] (n9) {Grade 7 \\ Vol 1};
    \node[junior, left=of n9] (n10) {Grade 7 \\ Vol 2};
    \node[junior, left=of n10] (n11) {Grade 8 \\ Vol 1};
    \node[junior, left=of n11] (n12) {Grade 8 \\ Vol 2};

    \node[junior, below=of n12] (n13) {Grade 9 \\ Full Vol};
    \node[senior, right=of n13] (n14) {Compulsory \\ Book 1};
    \node[senior, right=of n14] (n15) {Compulsory \\ Book 2};
    \node[senior, right=of n15] (n16) {Compulsory \\ Book 3};
    \node[senior, right=of n16] (n17) {Sel. Comp. \\ Book 1};
    \node[senior, right=of n17] (n18) {Sel. Comp. \\ Book 2};

    \node[senior, below=of n18] (n19) {Sel. Comp. \\ Book 3};
    \node[senior, left=of n19] (n20) {Sel. Comp. \\ Book 4};

    \draw[arrow] (n1) -- (n2); \draw[arrow] (n2) -- (n3); \draw[arrow] (n3) -- (n4);
    \draw[arrow] (n4) -- (n5); \draw[arrow] (n5) -- (n6);
    \draw[arrow] (n6.south) -- (n7.north);

    \draw[arrow] (n7) -- (n8); \draw[arrow] (n8) -- (n9); \draw[arrow] (n9) -- (n10);
    \draw[arrow] (n10) -- (n11); \draw[arrow] (n11) -- (n12);
    \draw[arrow] (n12.south) -- (n13.north);

    \draw[arrow] (n13) -- (n14); \draw[arrow] (n14) -- (n15); \draw[arrow] (n15) -- (n16);
    \draw[arrow] (n16) -- (n17); \draw[arrow] (n17) -- (n18);
    \draw[arrow] (n18.south) -- (n19.north);

    \draw[arrow] (n19) -- (n20);

    \end{tikzpicture}
    }
    \caption{The learning sequence for English in basic education.}
\end{figure*}

%% file: figures/sequence/physics_base.tex
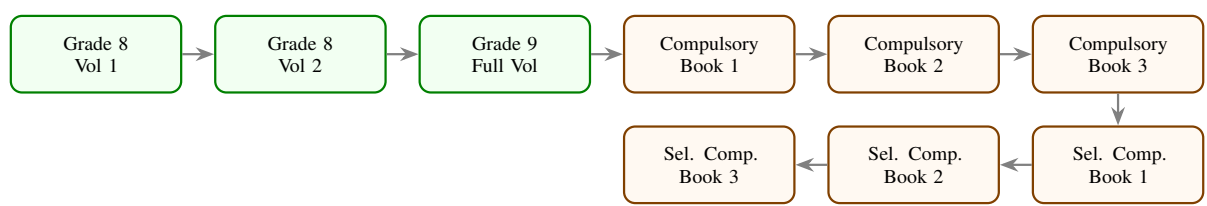
\begin{figure*}[htbp]
    \centering
    \resizebox{\linewidth}{!}{
    \begin{tikzpicture}[
        node distance=0.4cm and 0.4cm,
        box/.style={
            rectangle,
            draw,
            thick,
            rounded corners,
            minimum height=2.5em,
            text width=2.0cm,
            font=\scriptsize,
            align=center,
            inner sep=2pt
        },
        junior/.style={box, draw=green!50!black, fill=green!5},
        senior/.style={box, draw=orange!50!black, fill=orange!5},
        arrow/.style={-{Stealth[scale=1.0]}, thick, gray}
    ]

    \node[junior] (n1) {Grade 8 \\ Vol 1};
    \node[junior, right=of n1] (n2) {Grade 8 \\ Vol 2};
    \node[junior, right=of n2] (n3) {Grade 9 \\ Full Vol};
    \node[senior, right=of n3] (n4) {Compulsory \\ Book 1};
    \node[senior, right=of n4] (n5) {Compulsory \\ Book 2};
    \node[senior, right=of n5] (n6) {Compulsory \\ Book 3};

    \node[senior, below=of n6] (n7) {Sel. Comp. \\ Book 1};
    \node[senior, left=of n7] (n8) {Sel. Comp. \\ Book 2};
    \node[senior, left=of n8] (n9) {Sel. Comp. \\ Book 3};

    \draw[arrow] (n1) -- (n2); \draw[arrow] (n2) -- (n3); \draw[arrow] (n3) -- (n4);
    \draw[arrow] (n4) -- (n5); \draw[arrow] (n5) -- (n6);
    \draw[arrow] (n6.south) -- (n7.north);

    \draw[arrow] (n7) -- (n8); \draw[arrow] (n8) -- (n9);

    \end{tikzpicture}
    }
    \caption{The learning sequence for Physics in basic education.}
\end{figure*}

%% file: figures/sequence/chem_base.tex
\begin{figure*}[htbp]
    \centering
    \resizebox{\linewidth}{!}{
    \begin{tikzpicture}[
        node distance=0.4cm and 0.4cm,
        box/.style={rectangle, draw, thick, rounded corners, minimum height=2.5em, text width=2.2cm, font=\scriptsize, align=center, inner sep=2pt},
        junior/.style={box, draw=green!50!black, fill=green!5},
        senior/.style={box, draw=orange!50!black, fill=orange!5},
        arrow/.style={-{Stealth[scale=1.0]}, thick, gray}
    ]
    \node[junior] (n1) {Grade 9 \\ Vol 1};
    \node[junior, right=of n1] (n2) {Grade 9 \\ Vol 2};
    \node[senior, right=of n2] (n3) {Compulsory \\ Book 1};
    \node[senior, right=of n3] (n4) {Compulsory \\ Book 2};
    \node[senior, right=of n4] (n5) {Reaction \\ Principles};
    \node[senior, right=of n5] (n6) {Structure \& \\ Properties};
    \node[senior, below=of n6] (n7) {Organic \\ Chemistry};
    \draw[arrow] (n1) -- (n2); \draw[arrow] (n2) -- (n3); \draw[arrow] (n3) -- (n4);
    \draw[arrow] (n4) -- (n5); \draw[arrow] (n5) -- (n6); \draw[arrow] (n6.south) -- (n7.north);
    \end{tikzpicture}
    }
    \caption{The learning sequence for Chemistry in basic education.}
\end{figure*}
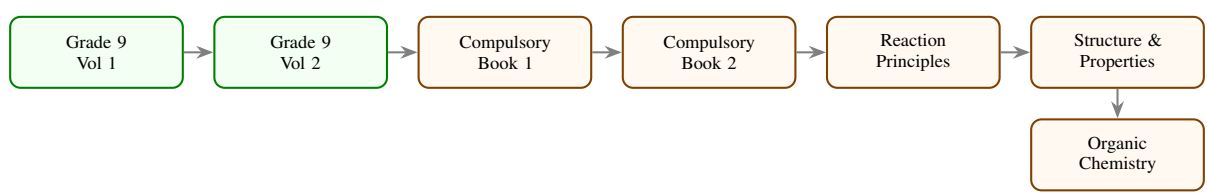

%% file: figures/sequence/bio_base.tex
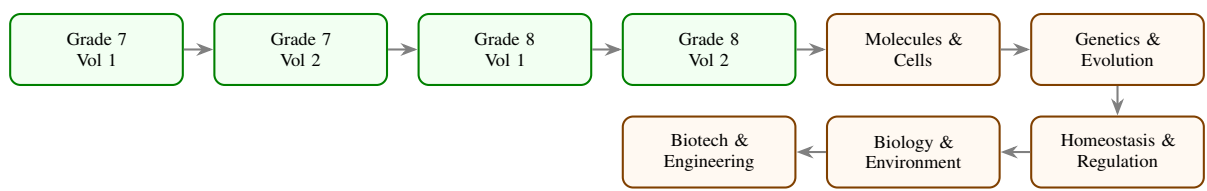
\begin{figure*}[htbp]
    \centering
    \resizebox{\linewidth}{!}{
    \begin{tikzpicture}[
        node distance=0.4cm and 0.4cm,
        box/.style={rectangle, draw, thick, rounded corners, minimum height=2.5em, text width=2.2cm, font=\scriptsize, align=center, inner sep=2pt},
        junior/.style={box, draw=green!50!black, fill=green!5},
        senior/.style={box, draw=orange!50!black, fill=orange!5},
        arrow/.style={-{Stealth[scale=1.0]}, thick, gray}
    ]
    \node[junior] (n1) {Grade 7 \\ Vol 1};
    \node[junior, right=of n1] (n2) {Grade 7 \\ Vol 2};
    \node[junior, right=of n2] (n3) {Grade 8 \\ Vol 1};
    \node[junior, right=of n3] (n4) {Grade 8 \\ Vol 2};
    \node[senior, right=of n4] (n5) {Molecules \& \\ Cells};
    \node[senior, right=of n5] (n6) {Genetics \& \\ Evolution};
    \node[senior, below=of n6] (n7) {Homeostasis \& \\ Regulation};
    \node[senior, left=of n7] (n8) {Biology \& \\ Environment};
    \node[senior, left=of n8] (n9) {Biotech \& \\ Engineering};
    \draw[arrow] (n1) -- (n2); \draw[arrow] (n2) -- (n3); \draw[arrow] (n3) -- (n4);
    \draw[arrow] (n4) -- (n5); \draw[arrow] (n5) -- (n6); \draw[arrow] (n6.south) -- (n7.north);
    \draw[arrow] (n7) -- (n8); \draw[arrow] (n8) -- (n9);
    \end{tikzpicture}
    }
    \caption{The learning sequence for Biology in basic education.}
\end{figure*}

%% file: figures/sequence/geo_base.tex
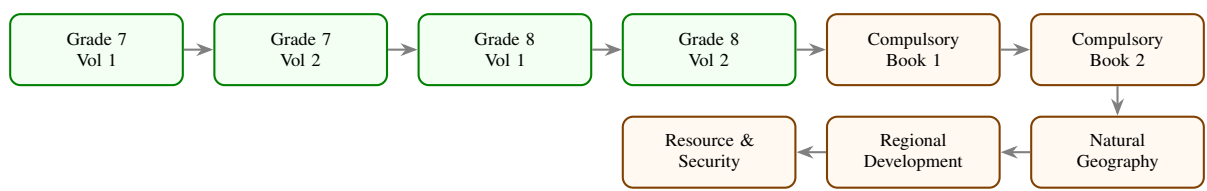
\begin{figure*}[htbp]
    \centering
    \resizebox{\linewidth}{!}{
    \begin{tikzpicture}[
        node distance=0.4cm and 0.4cm,
        box/.style={rectangle, draw, thick, rounded corners, minimum height=2.5em, text width=2.2cm, font=\scriptsize, align=center, inner sep=2pt},
        junior/.style={box, draw=green!50!black, fill=green!5},
        senior/.style={box, draw=orange!50!black, fill=orange!5},
        arrow/.style={-{Stealth[scale=1.0]}, thick, gray}
    ]
    \node[junior] (n1) {Grade 7 \\ Vol 1};
    \node[junior, right=of n1] (n2) {Grade 7 \\ Vol 2};
    \node[junior, right=of n2] (n3) {Grade 8 \\ Vol 1};
    \node[junior, right=of n3] (n4) {Grade 8 \\ Vol 2};
    \node[senior, right=of n4] (n5) {Compulsory \\ Book 1};
    \node[senior, right=of n5] (n6) {Compulsory \\ Book 2};
    \node[senior, below=of n6] (n7) {Natural \\ Geography};
    \node[senior, left=of n7] (n8) {Regional \\ Development};
    \node[senior, left=of n8] (n9) {Resource \& \\ Security};
    \draw[arrow] (n1) -- (n2); \draw[arrow] (n2) -- (n3); \draw[arrow] (n3) -- (n4);
    \draw[arrow] (n4) -- (n5); \draw[arrow] (n5) -- (n6); \draw[arrow] (n6.south) -- (n7.north);
    \draw[arrow] (n7) -- (n8); \draw[arrow] (n8) -- (n9);
    \end{tikzpicture}
    }
    \caption{The learning sequence for Geography in basic education.}
\end{figure*}

%% file: figures/sequence/history_base.tex
\begin{figure*}[htbp]
    \centering
    \resizebox{\linewidth}{!}{
    \begin{tikzpicture}[
        node distance=0.4cm and 0.4cm,
        box/.style={rectangle, draw, thick, rounded corners, minimum height=2.5em, text width=2.2cm, font=\scriptsize, align=center, inner sep=2pt},
        junior/.style={box, draw=green!50!black, fill=green!5},
        senior/.style={box, draw=orange!50!black, fill=orange!5},
        arrow/.style={-{Stealth[scale=1.0]}, thick, gray}
    ]
    \node[junior] (n1) {Grade 7 \\ Vol 1};
    \node[junior, right=of n1] (n2) {Chinese \\ History V2};
    \node[junior, right=of n2] (n3) {Grade 8 \\ Vol 1};
    \node[junior, right=of n3] (n4) {Grade 8 \\ Vol 2};
    \node[junior, right=of n4] (n5) {World \\ History V1};
    \node[junior, right=of n5] (n6) {Grade 9 \\ Vol 2};
    \node[senior, below=of n6] (n7) {Compendium of \\ History I};
    \node[senior, left=of n7] (n8) {Compendium of \\ History II};
    \node[senior, left=of n8] (n9) {State Systems \& \\ Governance};
    \node[senior, left=of n9] (n10) {Economy \& \\ Social Life};
    \node[senior, left=of n10] (n11) {Cultural \\ Exchange};
    \draw[arrow] (n1) -- (n2); \draw[arrow] (n2) -- (n3); \draw[arrow] (n3) -- (n4);
    \draw[arrow] (n4) -- (n5); \draw[arrow] (n5) -- (n6); \draw[arrow] (n6.south) -- (n7.north);
    \draw[arrow] (n7) -- (n8); \draw[arrow] (n8) -- (n9); \draw[arrow] (n9) -- (n10); \draw[arrow] (n10) -- (n11);
    \end{tikzpicture}
    }
    \caption{The learning sequence for History in basic education.}
\end{figure*}
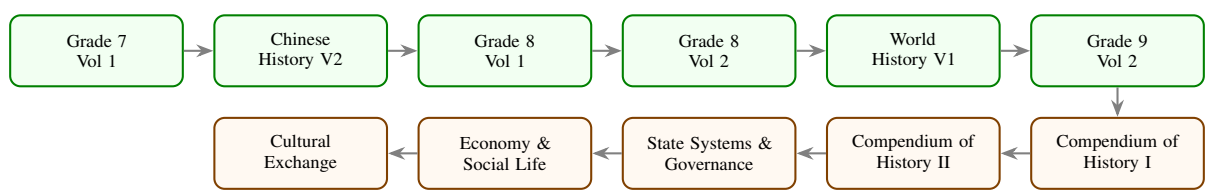

%% file: figures/sequence/mora_base.tex
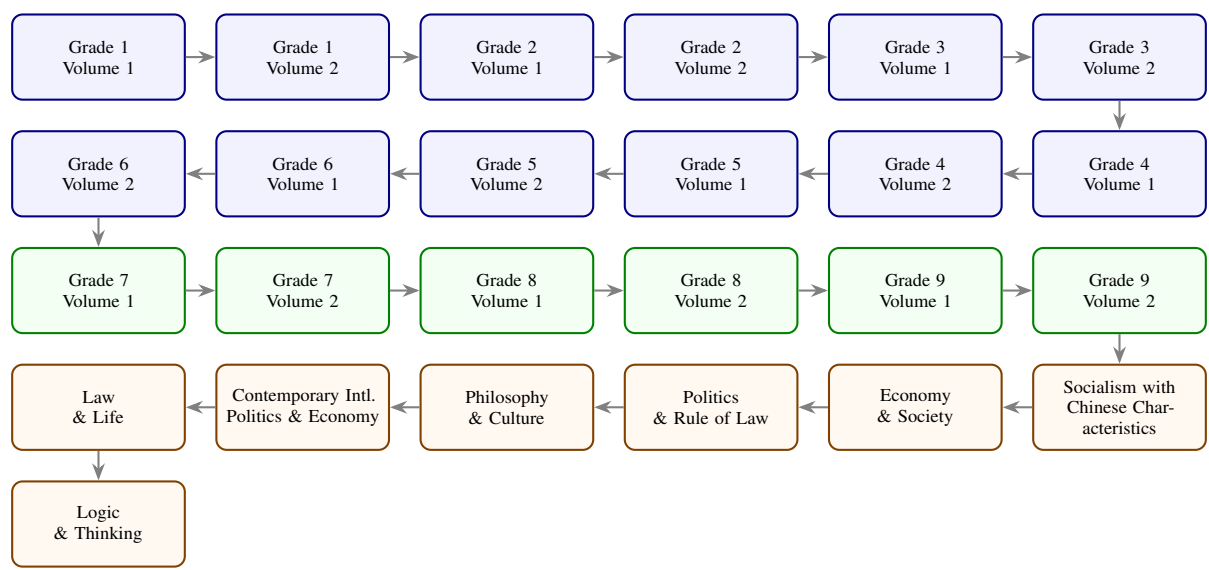
\begin{figure*}[htbp]
    \centering
    \resizebox{\linewidth}{!}{
    \begin{tikzpicture}[
        node distance=0.4cm and 0.4cm,
        box/.style={rectangle, draw, thick, rounded corners, minimum height=3em, text width=2.2cm, font=\scriptsize, align=center, inner sep=2pt},
        primary/.style={box, draw=blue!50!black, fill=blue!5},
        junior/.style={box, draw=green!50!black, fill=green!5},
        senior/.style={box, draw=orange!50!black, fill=orange!5},
        arrow/.style={-{Stealth[scale=1.0]}, thick, gray}
    ]
    \node[primary] (n1) {Grade 1 \\ Volume 1}; \node[primary, right=of n1] (n2) {Grade 1 \\ Volume 2}; \node[primary, right=of n2] (n3) {Grade 2 \\ Volume 1};
    \node[primary, right=of n3] (n4) {Grade 2 \\ Volume 2}; \node[primary, right=of n4] (n5) {Grade 3 \\ Volume 1}; \node[primary, right=of n5] (n6) {Grade 3 \\ Volume 2};
    \node[primary, below=of n6] (n7) {Grade 4 \\ Volume 1}; \node[primary, left=of n7] (n8) {Grade 4 \\ Volume 2}; \node[primary, left=of n8] (n9) {Grade 5 \\ Volume 1};
    \node[primary, left=of n9] (n10) {Grade 5 \\ Volume 2}; \node[primary, left=of n10] (n11) {Grade 6 \\ Volume 1}; \node[primary, left=of n11] (n12) {Grade 6 \\ Volume 2};
    \node[junior, below=of n12] (n13) {Grade 7 \\ Volume 1}; \node[junior, right=of n13] (n14) {Grade 7 \\ Volume 2}; \node[junior, right=of n14] (n15) {Grade 8 \\ Volume 1};
    \node[junior, right=of n15] (n16) {Grade 8 \\ Volume 2}; \node[junior, right=of n16] (n17) {Grade 9 \\ Volume 1}; \node[junior, right=of n17] (n18) {Grade 9 \\ Volume 2};
    \node[senior, below=of n18] (n19) {Socialism with \\ Chinese Characteristics}; \node[senior, left=of n19] (n20) {Economy \\ \& Society}; \node[senior, left=of n20] (n21) {Politics \\ \& Rule of Law};
    \node[senior, left=of n21] (n22) {Philosophy \\ \& Culture}; \node[senior, left=of n22] (n23) {Contemporary Intl. \\ Politics \& Economy};
    \node[senior, left=of n23] (n24) {Law \\ \& Life};

    \node[senior, below=of n24] (n25) {Logic \\ \& Thinking};

    \draw[arrow] (n1) -- (n2); \draw[arrow] (n2) -- (n3); \draw[arrow] (n3) -- (n4); \draw[arrow] (n4) -- (n5); \draw[arrow] (n5) -- (n6); \draw[arrow] (n6.south) -- (n7.north);
    \draw[arrow] (n7) -- (n8); \draw[arrow] (n8) -- (n9); \draw[arrow] (n9) -- (n10); \draw[arrow] (n10) -- (n11); \draw[arrow] (n11) -- (n12); \draw[arrow] (n12.south) -- (n13.north);
    \draw[arrow] (n13) -- (n14); \draw[arrow] (n14) -- (n15); \draw[arrow] (n15) -- (n16); \draw[arrow] (n16) -- (n17); \draw[arrow] (n17) -- (n18); \draw[arrow] (n18.south) -- (n19.north);
    \draw[arrow] (n19) -- (n20); \draw[arrow] (n20) -- (n21); \draw[arrow] (n21) -- (n22); \draw[arrow] (n22) -- (n23); \draw[arrow] (n23) -- (n24); \draw[arrow] (n24.south) -- (n25.north);
    \end{tikzpicture}
    }
    \caption{The learning sequence for Morality and Rule of Law / Ideological and Political Education in basic education.}
\end{figure*}

%% file: figures/sequence/music_base.tex
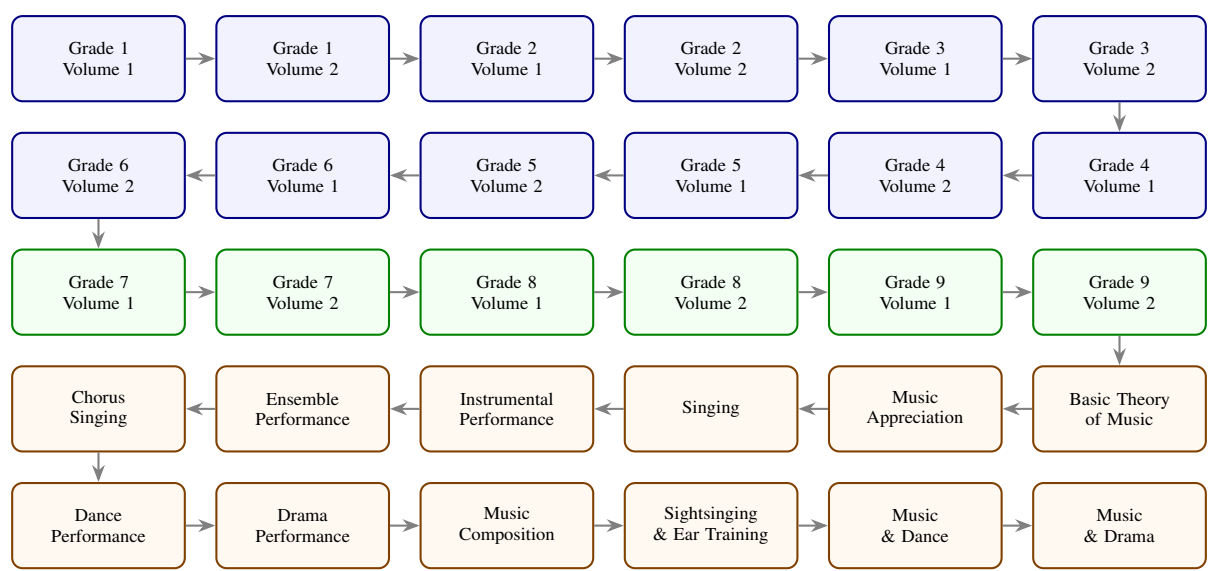
\begin{figure*}[htbp]
    \centering
    \resizebox{\linewidth}{!}{
    \begin{tikzpicture}[
        node distance=0.4cm and 0.4cm,
        box/.style={rectangle, draw, thick, rounded corners, minimum height=3em, text width=2.2cm, font=\scriptsize, align=center, inner sep=2pt},
        primary/.style={box, draw=blue!50!black, fill=blue!5},
        junior/.style={box, draw=green!50!black, fill=green!5},
        senior/.style={box, draw=orange!50!black, fill=orange!5},
        arrow/.style={-{Stealth[scale=1.0]}, thick, gray}
    ]
    \node[primary] (n1) {Grade 1 \\ Volume 1}; \node[primary, right=of n1] (n2) {Grade 1 \\ Volume 2}; \node[primary, right=of n2] (n3) {Grade 2 \\ Volume 1};
    \node[primary, right=of n3] (n4) {Grade 2 \\ Volume 2}; \node[primary, right=of n4] (n5) {Grade 3 \\ Volume 1}; \node[primary, right=of n5] (n6) {Grade 3 \\ Volume 2};
    \node[primary, below=of n6] (n7) {Grade 4 \\ Volume 1}; \node[primary, left=of n7] (n8) {Grade 4 \\ Volume 2}; \node[primary, left=of n8] (n9) {Grade 5 \\ Volume 1};
    \node[primary, left=of n9] (n10) {Grade 5 \\ Volume 2}; \node[primary, left=of n10] (n11) {Grade 6 \\ Volume 1}; \node[primary, left=of n11] (n12) {Grade 6 \\ Volume 2};
    \node[junior, below=of n12] (n13) {Grade 7 \\ Volume 1}; \node[junior, right=of n13] (n14) {Grade 7 \\ Volume 2}; \node[junior, right=of n14] (n15) {Grade 8 \\ Volume 1};
    \node[junior, right=of n15] (n16) {Grade 8 \\ Volume 2}; \node[junior, right=of n16] (n17) {Grade 9 \\ Volume 1}; \node[junior, right=of n17] (n18) {Grade 9 \\ Volume 2};
    \node[senior, below=of n18] (n19) {Basic Theory \\ of Music}; \node[senior, left=of n19] (n20) {Music \\ Appreciation}; \node[senior, left=of n20] (n21) {Singing};
    \node[senior, left=of n21] (n22) {Instrumental \\ Performance}; \node[senior, left=of n22] (n23) {Ensemble \\ Performance}; \node[senior, left=of n23] (n24) {Chorus \\ Singing};
    \node[senior, below=of n24] (n25) {Dance \\ Performance}; \node[senior, right=of n25] (n26) {Drama \\ Performance}; \node[senior, right=of n26] (n27) {Music \\ Composition};
    \node[senior, right=of n27] (n28) {Sightsinging \\ \& Ear Training}; \node[senior, right=of n28] (n29) {Music \\ \& Dance}; \node[senior, right=of n29] (n30) {Music \\ \& Drama};

    \draw[arrow] (n1) -- (n2); \draw[arrow] (n2) -- (n3); \draw[arrow] (n3) -- (n4); \draw[arrow] (n4) -- (n5); \draw[arrow] (n5) -- (n6); \draw[arrow] (n6.south) -- (n7.north);
    \draw[arrow] (n7) -- (n8); \draw[arrow] (n8) -- (n9); \draw[arrow] (n9) -- (n10); \draw[arrow] (n10) -- (n11); \draw[arrow] (n11) -- (n12); \draw[arrow] (n12.south) -- (n13.north);
    \draw[arrow] (n13) -- (n14); \draw[arrow] (n14) -- (n15); \draw[arrow] (n15) -- (n16); \draw[arrow] (n16) -- (n17); \draw[arrow] (n17) -- (n18); \draw[arrow] (n18.south) -- (n19.north);
    \draw[arrow] (n19) -- (n20); \draw[arrow] (n20) -- (n21); \draw[arrow] (n21) -- (n22); \draw[arrow] (n22) -- (n23); \draw[arrow] (n23) -- (n24); \draw[arrow] (n24.south) -- (n25.north);
    \draw[arrow] (n25) -- (n26); \draw[arrow] (n26) -- (n27); \draw[arrow] (n27) -- (n28); \draw[arrow] (n28) -- (n29); \draw[arrow] (n29) -- (n30);
    \end{tikzpicture}
    }
    \caption{The learning sequence for Music in basic education.}
\end{figure*}

%% file: figures/sequence/it_base.tex
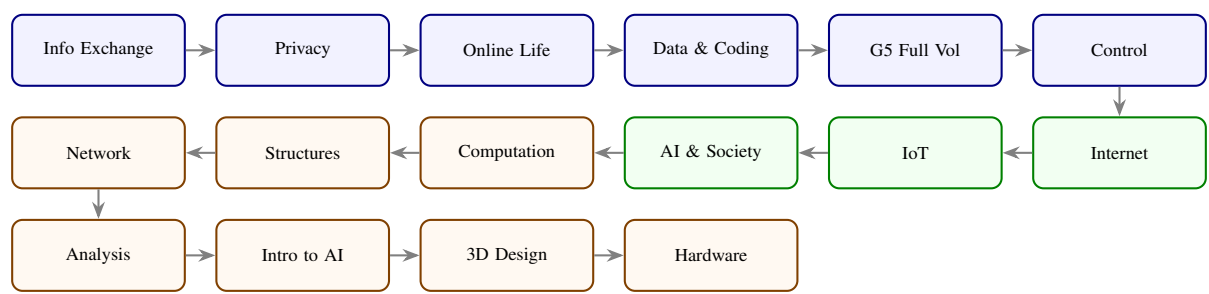
\begin{figure*}[htbp]
    \centering
    \resizebox{\linewidth}{!}{
    \begin{tikzpicture}[
        node distance=0.4cm and 0.4cm,
        box/.style={rectangle, draw, thick, rounded corners, minimum height=2.5em, text width=2.2cm, font=\scriptsize, align=center, inner sep=2pt},
        primary/.style={box, draw=blue!50!black, fill=blue!5},
        junior/.style={box, draw=green!50!black, fill=green!5},
        senior/.style={box, draw=orange!50!black, fill=orange!5},
        arrow/.style={-{Stealth[scale=1.0]}, thick, gray}
    ]
    \node[primary] (n1) {Info Exchange}; \node[primary, right=of n1] (n2) {Privacy}; \node[primary, right=of n2] (n3) {Online Life};
    \node[primary, right=of n3] (n4) {Data \& Coding}; \node[primary, right=of n4] (n5) {G5 Full Vol}; \node[primary, right=of n5] (n6) {Control};
    \node[junior, below=of n6] (n7) {Internet}; \node[junior, left=of n7] (n8) {IoT}; \node[junior, left=of n8] (n9) {AI \& Society};
    \node[senior, left=of n9] (n10) {Computation}; \node[senior, left=of n10] (n11) {Structures}; \node[senior, left=of n11] (n12) {Network};
    \node[senior, below=of n12] (n13) {Analysis}; \node[senior, right=of n13] (n14) {Intro to AI}; \node[senior, right=of n14] (n15) {3D Design};
    \node[senior, right=of n15] (n16) {Hardware};
    \draw[arrow] (n1) -- (n2); \draw[arrow] (n2) -- (n3); \draw[arrow] (n3) -- (n4); \draw[arrow] (n4) -- (n5); \draw[arrow] (n5) -- (n6); \draw[arrow] (n6.south) -- (n7.north);
    \draw[arrow] (n7) -- (n8); \draw[arrow] (n8) -- (n9); \draw[arrow] (n9) -- (n10); \draw[arrow] (n10) -- (n11); \draw[arrow] (n11) -- (n12); \draw[arrow] (n12.south) -- (n13.north);
    \draw[arrow] (n13) -- (n14); \draw[arrow] (n14) -- (n15); \draw[arrow] (n15) -- (n16);
    \end{tikzpicture}
    }
    \caption{The learning sequence for Information Science / Information Technology in basic education.}
\end{figure*}

%% file: figures/sequence/art_base.tex
\begin{figure*}[htbp]
    \centering
    \resizebox{\linewidth}{!}{
    \begin{tikzpicture}[
        node distance=0.4cm and 0.4cm,
        box/.style={rectangle, draw, thick, rounded corners, minimum height=3em, text width=2.2cm, font=\scriptsize, align=center, inner sep=2pt},
        primary/.style={box, draw=blue!50!black, fill=blue!5},
        junior/.style={box, draw=green!50!black, fill=green!5},
        senior/.style={box, draw=orange!50!black, fill=orange!5},
        arrow/.style={-{Stealth[scale=1.0]}, thick, gray}
    ]
    \node[primary] (n1) {Grade 1 \\ Volume 1}; \node[primary, right=of n1] (n2) {Grade 1 \\ Volume 2}; \node[primary, right=of n2] (n3) {Grade 2 \\ Volume 1};
    \node[primary, right=of n3] (n4) {Grade 2 \\ Volume 2}; \node[primary, right=of n4] (n5) {Grade 3 \\ Volume 1}; \node[primary, right=of n5] (n6) {Grade 3 \\ Volume 2};
    \node[primary, below=of n6] (n7) {Grade 4 \\ Volume 1}; \node[primary, left=of n7] (n8) {Grade 4 \\ Volume 2}; \node[primary, left=of n8] (n9) {Grade 5 \\ Volume 1};
    \node[primary, left=of n9] (n10) {Grade 5 \\ Volume 2}; \node[primary, left=of n10] (n11) {Grade 6 \\ Volume 1}; \node[primary, left=of n11] (n12) {Grade 6 \\ Volume 2};
    \node[junior, below=of n12] (n13) {Grade 7 \\ Volume 1}; \node[junior, right=of n13] (n14) {Grade 7 \\ Volume 2}; \node[junior, right=of n14] (n15) {Grade 8 \\ Volume 1};
    \node[junior, right=of n15] (n16) {Grade 8 \\ Volume 2}; \node[junior, right=of n16] (n17) {Grade 9 \\ Volume 1}; \node[junior, right=of n17] (n18) {Grade 9 \\ Volume 2};
    \node[senior, below=of n18] (n19) {Art \\ Appreciation}; \node[senior, left=of n19] (n20) {Painting}; \node[senior, left=of n20] (n21) {Chinese \\ Painting};
    \node[senior, left=of n21] (n22) {Sculpture}; \node[senior, left=of n22] (n23) {Design}; \node[senior, left=of n23] (n24) {Crafts};
    \node[senior, below=of n24] (n25) {Modern \\ Media Art};

    \draw[arrow] (n1) -- (n2); \draw[arrow] (n2) -- (n3); \draw[arrow] (n3) -- (n4); \draw[arrow] (n4) -- (n5); \draw[arrow] (n5) -- (n6); \draw[arrow] (n6.south) -- (n7.north);
    \draw[arrow] (n7) -- (n8); \draw[arrow] (n8) -- (n9); \draw[arrow] (n9) -- (n10); \draw[arrow] (n10) -- (n11); \draw[arrow] (n11) -- (n12); \draw[arrow] (n12.south) -- (n13.north);
    \draw[arrow] (n13) -- (n14); \draw[arrow] (n14) -- (n15); \draw[arrow] (n15) -- (n16); \draw[arrow] (n16) -- (n17); \draw[arrow] (n17) -- (n18); \draw[arrow] (n18.south) -- (n19.north);
    \draw[arrow] (n19) -- (n20); \draw[arrow] (n20) -- (n21); \draw[arrow] (n21) -- (n22); \draw[arrow] (n22) -- (n23); \draw[arrow] (n23) -- (n24); \draw[arrow] (n24.south) -- (n25.north);
    \end{tikzpicture}
    }
    \caption{The learning sequence for Art in basic education.}
\end{figure*}
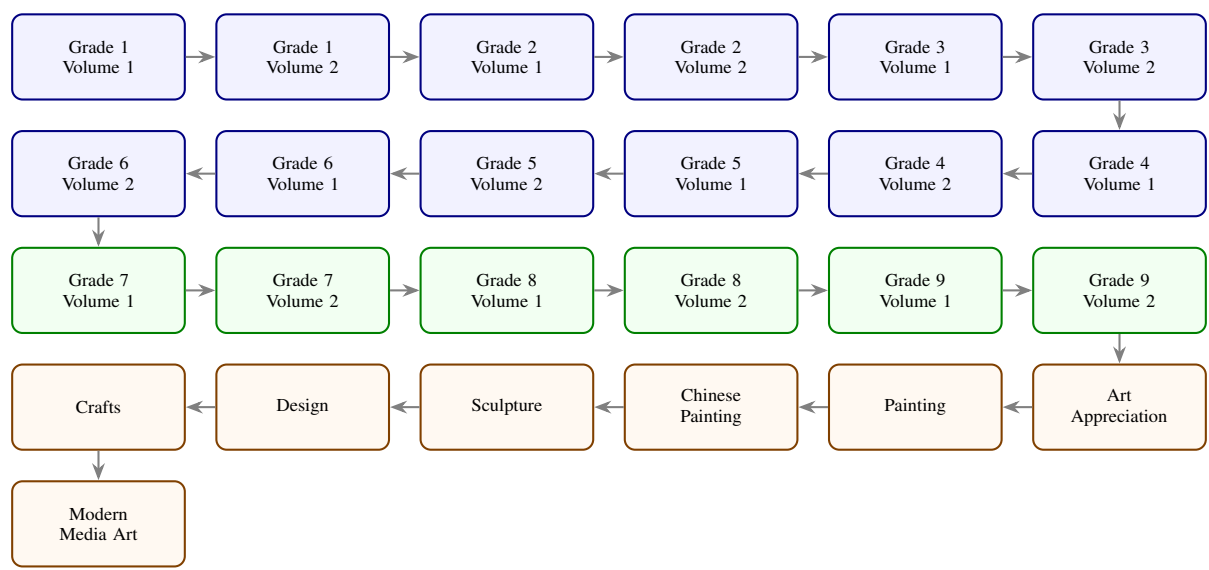

%% file: figures/sequence/pe_base.tex
\begin{figure*}[htbp]
    \centering
    \begin{tikzpicture}[
        node distance=0.4cm and 0.4cm,
        box/.style={rectangle, draw, thick, rounded corners, minimum height=3em, text width=2.2cm, font=\scriptsize, align=center, inner sep=2pt},
        primary/.style={box, draw=blue!50!black, fill=blue!5},
        arrow/.style={-{Stealth[scale=1.0]}, thick, gray}
    ]
    \node[primary] (n1) {Grades 1-2 \\ Full Volume};
    \node[primary, right=of n1] (n2) {Grades 3-4 \\ Full Volume};
    \node[primary, right=of n2] (n3) {Grades 5-6 \\ Full Volume};
    \draw[arrow] (n1) -- (n2); \draw[arrow] (n2) -- (n3);
    \end{tikzpicture}
    \caption{The learning sequence for Physical Education and Health in basic education.}
    \label{fig:pe}
\end{figure*}
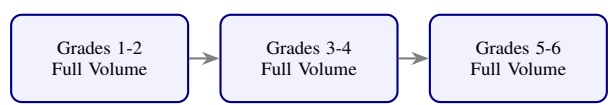

%% file: figures/sequence/law_pro.tex
\begin{figure*}[htbp]
    \centering
    \resizebox{\linewidth}{!}{
    \begin{tikzpicture}[
        node distance=0.4cm and 0.25cm,
        box/.style={
            rectangle,
            draw,
            thick,
            rounded corners,
            minimum height=2.5em,
            text width=1.8cm,
            font=\scriptsize,
            align=center,
            inner sep=2pt
        },
        law/.style={box, draw=blue!60!black, fill=blue!5},
        marx/.style={box, draw=green!60!black, fill=green!5},
        socio/.style={box, draw=orange!60!black, fill=orange!5},
        politi/.style={box, draw=purple!60!black, fill=purple!5},
        arrow/.style={-{Stealth[scale=1.0]}, thick, gray}
    ]

    \node[law] (a1) {Chinese Foundation};
    \node[font=\scriptsize\bfseries, anchor=west] at ([yshift=0.25cm]a1.north west) {Law};
    \node[law, right=of a1] (a2) {Ideological \& Political};
    \node[law, right=of a2] (a3) {History Foundation};
    \node[law, right=of a3] (a4) {Constitution};
    \node[law, right=of a4] (a5) {Jurisprudence};
    \node[law, right=of a5] (a6) {Civil Law Course};
    \node[law, right=of a6] (a7) {Criminal Law};
    \node[law, right=of a7] (a8) {Public Int'l Law};
    \draw[arrow] (a1) -- (a2); \draw[arrow] (a2) -- (a3); \draw[arrow] (a3) -- (a4);
    \draw[arrow] (a4) -- (a5); \draw[arrow] (a5) -- (a6); \draw[arrow] (a6) -- (a7); \draw[arrow] (a7) -- (a8);

    \node[marx, below=of a1, yshift=-0.2cm] (b1) {Chinese Foundation};
    \node[font=\scriptsize\bfseries, anchor=west] at ([yshift=0.25cm]b1.north west) {Marxist Theory};
    \node[marx, right=of b1] (b2) {Ideological \& Political};
    \node[marx, right=of b2] (b3) {History Foundation};
    \node[marx, right=of b3] (b4) {Modern Chinese History};
    \node[marx, right=of b4] (b5) {Maoism \& Socialism};
    \node[marx, right=of b5] (b6) {I \& P Education};
    \draw[arrow] (b1) -- (b2); \draw[arrow] (b2) -- (b3); \draw[arrow] (b3) -- (b4);
    \draw[arrow] (b4) -- (b5); \draw[arrow] (b5) -- (b6);

    \node[socio, below=of b1, yshift=-0.2cm] (c1) {Chinese Foundation};
    \node[font=\scriptsize\bfseries, anchor=west] at ([yshift=0.25cm]c1.north west) {Sociology};
    \node[socio, right=of c1] (c2) {Ideological \& Political};
    \node[socio, right=of c2] (c3) {History Foundation};
    \node[socio, right=of c3] (c4) {Intro to Sociology};
    \node[socio, right=of c4] (c5) {Social Work};
    \draw[arrow] (c1) -- (c2); \draw[arrow] (c2) -- (c3); \draw[arrow] (c3) -- (c4);
    \draw[arrow] (c4) -- (c5);

    \node[politi, below=of c1, yshift=-0.2cm] (d1) {Chinese Foundation};
    \node[font=\scriptsize\bfseries, anchor=west] at ([yshift=0.25cm]d1.north west) {Political Science};
    \node[politi, right=of d1] (d2) {Ideological \& Political};
    \node[politi, right=of d2] (d3) {History Foundation};
    \node[politi, right=of d3] (d4) {Geography Foundation};
    \node[politi, right=of d4] (d5) {Intro Political Science};
    \draw[arrow] (d1) -- (d2); \draw[arrow] (d2) -- (d3); \draw[arrow] (d3) -- (d4);
    \draw[arrow] (d4) -- (d5);

    \end{tikzpicture}
    }
    \caption{Learning sequences of four sub-disciplines under the Law primary discipline: Law (blue), Marxist Theory (green), Sociology (orange), and Political Science (purple).}
    \label{fig:law}
\end{figure*}
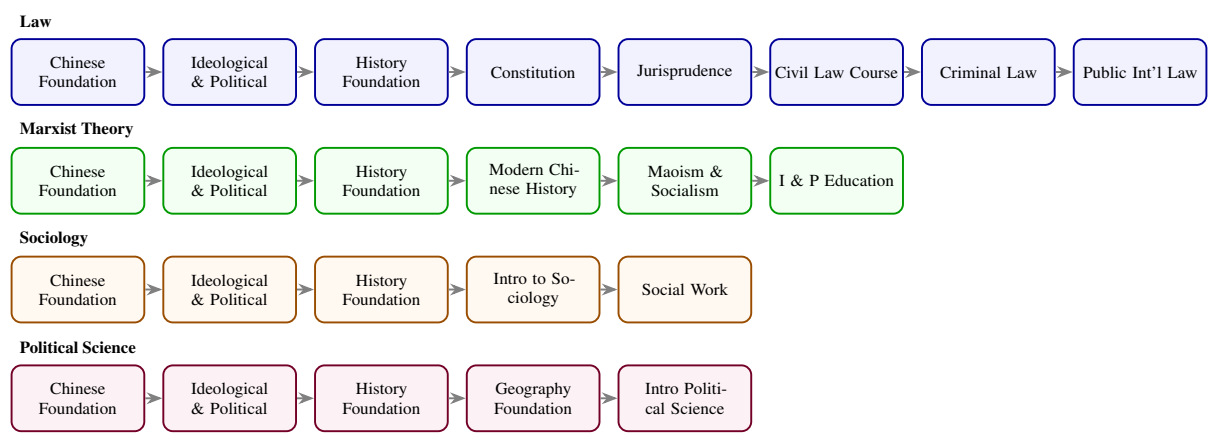

%% file: figures/sequence/gong_pro.tex
\begin{figure*}[htbp]
    \centering
    \
    \begin{tikzpicture}[
        node distance=0.4cm and 0.25cm,
        box/.style={
            rectangle,
            draw,
            thick,
            rounded corners,
            minimum height=2.5em,
            text width=1.6cm,
            font=\tiny,
            align=center,
            inner sep=2pt
        },
        materials/.style={box, draw=blue!70!black, fill=blue!5},
        electrical/.style={box, draw=green!70!black, fill=green!5},
        electronic/.style={box, draw=orange!70!black, fill=orange!5},
        textile/.style={box, draw=purple!70!black, fill=purple!5},
        marine/.style={box, draw=red!70!black, fill=red!5},
        environ/.style={box, draw=cyan!70!black, fill=cyan!5},
        mechanical/.style={box, draw=olive!70!black, fill=olive!5},
        computer/.style={box, draw=magenta!70!black, fill=magenta!5},
        arch/.style={box, draw=lime!70!black, fill=lime!5},
        mechanics/.style={box, draw=teal!70!black, fill=teal!5},
        energy/.style={box, draw=violet!70!black, fill=violet!5},
        hydraulic/.style={box, draw=brown!70!black, fill=brown!5},
        civil/.style={box, draw=pink!70!black, fill=pink!5},
        instrument/.style={box, draw=darkgray!70!black, fill=darkgray!5},
        auto/.style={box, draw=red!50!black, fill=red!10},
        arrow/.style={-{Stealth[scale=0.8]}, thick, gray}
    ]

    \node[materials] (a1) {Mathematics \\ Foundation};
    \node[font=\tiny\bfseries, anchor=west] at ([yshift=0.25cm]a1.north west) {Materials};
    \node[materials, right=of a1] (a2) {Physics \\ Foundation};
    \node[materials, right=of a2] (a3) {Chemistry \\ Foundation};
    \node[materials, right=of a3] (a4) {Materials \\ Science};
    \node[materials, right=of a4] (a5) {Mechanical \\ Eng. Materials};
    \node[materials, right=of a5] (a6) {Forming \\ Tech.};
    \node[materials, right=of a6] (a7) {Electric \\ Drive Systems};
    \draw[arrow] (a1) -- (a2); \draw[arrow] (a2) -- (a3); \draw[arrow] (a3) -- (a4);
    \draw[arrow] (a4) -- (a5); \draw[arrow] (a5) -- (a6); \draw[arrow] (a6) -- (a7);

    \node[electrical, below=of a1, yshift=-0.2cm] (b1) {Mathematics \\ Foundation};
    \node[font=\tiny\bfseries, anchor=west] at ([yshift=0.25cm]b1.north west) {Electrical};
    \node[electrical, right=of b1] (b2) {Physics \\ Foundation};
    \node[electrical, right=of b2] (b3) {IT \\ Foundation};
    \node[electrical, right=of b3] (b4) {Circuit \\ Analysis};
    \node[electrical, right=of b4] (b5) {Electro- \\ technics};
    \node[electrical, right=of b5] (b6) {Analog \\ Electronics};
    \node[electrical, right=of b6] (b7) {Digital \\ Circuits};
    \node[electrical, right=of b7] (b8) {Electronic \\ Principles};
    \node[electrical, below=of b8, yshift=-0.2cm] (b9) {Measurement \\ Circuits};
    \draw[arrow] (b1) -- (b2); \draw[arrow] (b2) -- (b3); \draw[arrow] (b3) -- (b4);
    \draw[arrow] (b4) -- (b5); \draw[arrow] (b5) -- (b6); \draw[arrow] (b6) -- (b7); \draw[arrow] (b7) -- (b8);
    \draw[arrow] (b8.south) -- (b9.north);

    \node[electronic, below=of b1, yshift=-0.2cm] (c1) {Mathematics \\ Foundation};
    \node[font=\tiny\bfseries, anchor=west] at ([yshift=0.25cm]c1.north west) {Electronic Info};
    \node[electronic, right=of c1] (c2) {Physics \\ Foundation};
    \node[electronic, right=of c2] (c3) {IT \\ Foundation};
    \node[electronic, right=of c3] (c4) {Signals \\ \& Systems};
    \node[electronic, right=of c4] (c5) {Digital Signal \\ Processing};
    \node[electronic, right=of c5] (c6) {Principles of \\ Comm.};
    \draw[arrow] (c1) -- (c2); \draw[arrow] (c2) -- (c3); \draw[arrow] (c3) -- (c4);
    \draw[arrow] (c4) -- (c5); \draw[arrow] (c5) -- (c6);

    \node[textile, below=of c1, yshift=-0.2cm] (d1) {Chemistry \\ Foundation};
    \node[font=\tiny\bfseries, anchor=west] at ([yshift=0.25cm]d1.north west) {Textile};
    \node[textile, right=of d1] (d2) {Physics \\ Foundation};
    \node[textile, right=of d2] (d3) {Dyeing \& \\ Finishing};
    \draw[arrow] (d1) -- (d2); \draw[arrow] (d2) -- (d3);

    \node[marine, below=of d1, yshift=-0.2cm] (e1) {Mathematics \\ Foundation};
    \node[font=\tiny\bfseries, anchor=west] at ([yshift=0.25cm]e1.north west) {Marine Eng.};
    \node[marine, right=of e1] (e2) {Physics \\ Foundation};
    \node[marine, right=of e2] (e3) {Principles \\ of Ships};
    \draw[arrow] (e1) -- (e2); \draw[arrow] (e2) -- (e3);

    \node[environ, below=of e1, yshift=-0.2cm] (f1) {Mathematics \\ Foundation};
    \node[font=\tiny\bfseries, anchor=west] at ([yshift=0.25cm]f1.north west) {Environmental};
    \node[environ, right=of f1] (f2) {Physics \\ Foundation};
    \node[environ, right=of f2] (f3) {Chemistry \\ Foundation};
    \node[environ, right=of f3] (f4) {Biology \\ Foundation};
    \node[environ, right=of f4] (f5) {Environ. \\ Chemistry};
    \node[environ, right=of f5] (f6) {Water \\ Pollution Eng.};
    \draw[arrow] (f1) -- (f2); \draw[arrow] (f2) -- (f3); \draw[arrow] (f3) -- (f4);
    \draw[arrow] (f4) -- (f5); \draw[arrow] (f5) -- (f6);

    \node[mechanical, below=of f1, yshift=-0.2cm] (g1) {Mathematics \\ Foundation};
    \node[font=\tiny\bfseries, anchor=west] at ([yshift=0.25cm]g1.north west) {Mechanical};
    \node[mechanical, right=of g1] (g2) {Physics \\ Foundation};
    \node[mechanical, right=of g2] (g3) {Mechanical \\ Drawing};
    \node[mechanical, right=of g3] (g4) {Mechanical \\ Principles};
    \node[mechanical, right=of g4] (g5) {Mechanical \\ Design};
    \node[mechanical, right=of g5] (g6) {Eng. Materials};
    \node[mechanical, right=of g6] (g7) {Manufacturing \\ Tech.};
    \node[mechanical, right=of g7] (g8) {Hydraulic \\ Trans.};
    \draw[arrow] (g1) -- (g2); \draw[arrow] (g2) -- (g3); \draw[arrow] (g3) -- (g4);
    \draw[arrow] (g4) -- (g5); \draw[arrow] (g5) -- (g6); \draw[arrow] (g6) -- (g7);
    \draw[arrow] (g7) -- (g8);

    \node[computer, below=of g1, yshift=-0.2cm] (h1) {Mathematics \\ Foundation};
    \node[font=\tiny\bfseries, anchor=west] at ([yshift=0.25cm]h1.north west) {Computer Science};
    \node[computer, right=of h1] (h2) {English \\ Foundation};
    \node[computer, right=of h2] (h3) {IT \\ Foundation};
    \node[computer, right=of h3] (h4) {Intro to \\ Computing};
    \node[computer, right=of h4] (h5) {C \\ Programming};
    \node[computer, right=of h5] (h6) {C++ \\ Programming};
    \node[computer, right=of h6] (h7) {Computer \\ Org.};
    \node[computer, right=of h7] (h8) {Operating \\ Systems};
    \node[computer, below=of h8, yshift=-0.2cm] (h9) {Networks};
    \node[computer, left=of h9] (h10) {Database};
    \node[computer, left=of h10] (h11) {Intro to AI};
    \node[computer, left=of h11] (h12) {Machine \\ Learning};
    \draw[arrow] (h1) -- (h2); \draw[arrow] (h2) -- (h3); \draw[arrow] (h3) -- (h4);
    \draw[arrow] (h4) -- (h5); \draw[arrow] (h5) -- (h6); \draw[arrow] (h6) -- (h7);
    \draw[arrow] (h7) -- (h8); \draw[arrow] (h8.south) -- (h9.north);
    \draw[arrow] (h9) -- (h10); \draw[arrow] (h10) -- (h11); \draw[arrow] (h11) -- (h12);

    \node[arch, below=of h1, yshift=-0.2cm] (i1) {Mathematics \\ Foundation};
    \node[font=\tiny\bfseries, anchor=west] at ([yshift=0.25cm]i1.north west) {Architecture};
    \node[arch, right=of i1] (i2) {Physics \\ Foundation};
    \node[arch, right=of i2] (i3) {Fine Arts \\ Foundation};
    \node[arch, right=of i3] (i4) {History of \\ Architecture};
    \draw[arrow] (i1) -- (i2); \draw[arrow] (i2) -- (i3); \draw[arrow] (i3) -- (i4);

    \node[mechanics, below=of i1, yshift=-0.2cm] (j1) {Mathematics \\ Foundation};
    \node[font=\tiny\bfseries, anchor=west] at ([yshift=0.25cm]j1.north west) {Mechanics};
    \node[mechanics, right=of j1] (j2) {Physics \\ Foundation};
    \node[mechanics, right=of j2] (j3) {Theoretical \\ Mechanics};
    \node[mechanics, right=of j3] (j4) {Mechanics of \\ Materials};
    \draw[arrow] (j1) -- (j2); \draw[arrow] (j2) -- (j3); \draw[arrow] (j3) -- (j4);

    \node[energy, below=of j1, yshift=-0.2cm] (k1) {Mathematics \\ Foundation};
    \node[font=\tiny\bfseries, anchor=west] at ([yshift=0.25cm]k1.north west) {Energy \& Power};
    \node[energy, right=of k1] (k2) {Physics \\ Foundation};
    \node[energy, right=of k2] (k3) {Chemistry \\ Foundation};
    \node[energy, right=of k3] (k4) {Eng. \\ Thermodynamics};
    \draw[arrow] (k1) -- (k2); \draw[arrow] (k2) -- (k3); \draw[arrow] (k3) -- (k4);

    \node[hydraulic, below=of k1, yshift=-0.2cm] (l1) {Mathematics \\ Foundation};
    \node[font=\tiny\bfseries, anchor=west] at ([yshift=0.25cm]l1.north west) {Hydraulic};
    \node[hydraulic, right=of l1] (l2) {Physics \\ Foundation};
    \node[hydraulic, right=of l2] (l3) {Hydraulic \\ Construction};
    \draw[arrow] (l1) -- (l2); \draw[arrow] (l2) -- (l3);

    \node[civil, below=of l1, yshift=-0.2cm] (m1) {Mathematics \\ Foundation};
    \node[font=\tiny\bfseries, anchor=west] at ([yshift=0.25cm]m1.north west) {Civil Eng.};
    \node[civil, right=of m1] (m2) {Physics \\ Foundation};
    \node[civil, right=of m2] (m3) {Structural \\ Mech. I};
    \node[civil, right=of m3] (m4) {Structural \\ Mech. II};
    \draw[arrow] (m1) -- (m2); \draw[arrow] (m2) -- (m3); \draw[arrow] (m3) -- (m4);

    \node[instrument, below=of m1, yshift=-0.2cm] (n1) {Mathematics \\ Foundation};
    \node[font=\tiny\bfseries, anchor=west] at ([yshift=0.25cm]n1.north west) {Instrument};
    \node[instrument, right=of n1] (n2) {Physics \\ Foundation};
    \node[instrument, right=of n2] (n3) {Error Theory \\ \& Processing};
    \draw[arrow] (n1) -- (n2); \draw[arrow] (n2) -- (n3);

    \node[auto, below=of n1, yshift=-0.2cm] (o1) {Mathematics \\ Foundation};
    \node[font=\tiny\bfseries, anchor=west] at ([yshift=0.25cm]o1.north west) {Automation};
    \node[auto, right=of o1] (o2) {Physics \\ Foundation};
    \node[auto, right=of o2] (o3) {IT \\ Foundation};
    \node[auto, right=of o3] (o4) {Analog \\ Electronics};
    \node[auto, right=of o4] (o5) {Digital \\ Electronics};
    \draw[arrow] (o1) -- (o2); \draw[arrow] (o2) -- (o3); \draw[arrow] (o3) -- (o4); \draw[arrow] (o4) -- (o5);

    \end{tikzpicture}
    \caption{Learning sequences of fifteen Engineering sub-disciplines in higher education.}
    \label{fig:gong}
\end{figure*}
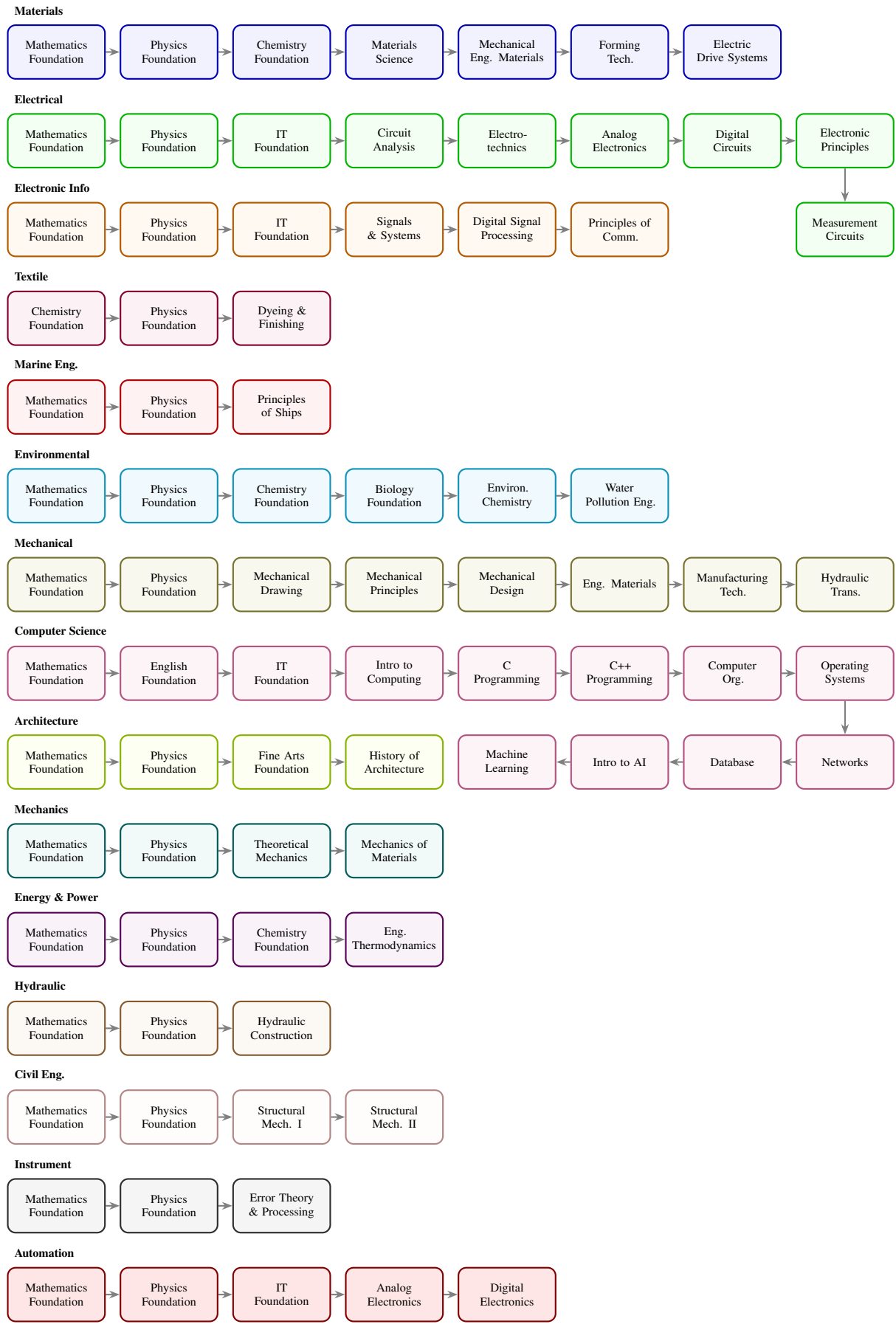

%% file: figures/prompt/gen_kg.tex
\begin{figure*}[htbp]
\centering
\begin{mybox}{\Large \textbf{Prompt for Knowledge Graph Generation}}
\begin{lstlisting}[style=promptstyle, escapeinside={(*}{*)}]
(*\normalsize\textbf{System Prompt:}\vspace{0.5\baselineskip}*)
You are a senior education expert in {subject} and a Knowledge Graph architect. Your task is to construct a high-precision structured knowledge graph based on the provided content from the textbook "{book_name}".

(*\normalsize\textbf{User Prompt:}\vspace{0.5\baselineskip}*)
(*\textbf{[Core Task]}*)
Read the [Textbook Excerpts] at the end and output a JSON object containing a "Knowledge Tree" and "Relationship Triplets".

(*\textbf{[Key Requirements: Catalog-Driven Structure]}*)
1. Strictly follow the catalog hierarchy: Identify structures such as Chapters, Sections, and Subheadings.
   - Level 1 children should correspond to "Chapters" or "Units".
   - Level 2 children should correspond to "Sections" or "Lessons".
   - Level 3 and below should correspond to specific knowledge points, concepts, formulas, or theorems.
2. Granularity Control:
   - Ensure the catalog structure is complete; do not skip chapters.
   - Refine knowledge points under Level 2/3 (e.g., "Positive Numbers" should include children like "Definition", "Representation", "Real-world Examples").
   - Maximum tree depth: 5 layers.

(*\textbf{[Field Definitions]}*)
1. knowledge_tree: {{"title": "...", "content": "...", "difficulty": "1-5", "children": []}}
   - Difficulty: 1 (Easy/Basic), 3 (Medium/Comprehensive), 5 (Hard/Advanced).
2. triplets: {{"head": "A", "tail": "B", "relation": "..."}}
   - Focus on non-hierarchical logic (Prerequisite, Application, Attribute, Comparison).

(*\textbf{[Data Consistency]}*)
- Terminology must be accurate for {subject}. 
- Every node must have: "title", "content", "difficulty", "children".
- "content" should be a concise summary of the core definition or formula.

Output example:
{{
  "knowledge_tree": {{
    "title": "Rational Numbers",
    "children": [
      {{"title": "Addition of Rational Numbers", "content": "Rules for adding...", "difficulty": "2", "children": []}}
    ]
  }},
  "triplets": [
    {{"head": "Number Line", "tail": "Comparison", "relation": "Used for"}}
  ]
}}

Textbook Excerpts:
{text}
\end{lstlisting}
\end{mybox}
\caption{The prompt for Knowledge Graph Generation}
\label{fig:kg}
\end{figure*}

%% file: figures/prompt/l1_cot.tex
\begin{figure*}[htbp]
\centering
\begin{mybox}{\Large \textbf{Prompt for Textbook Recommendation (CoT)}}
\begin{lstlisting}[style=promptstyle, escapeinside={(*}{*)}]
(*\normalsize\textbf{System Prompt:}\vspace{0.5\baselineskip}*)
You are a curriculum planning consultant responsible for recommending the textbook that a student should study right now. Your thinking process must be concise and directly lead to a decision.

(*\textbf{[Core Rules]}*)
- The learning sequence must start from the most basic and prerequisite textbooks, progressing step-by-step according to a natural pedagogical order.
- If there are unmastered prerequisite textbooks: Select the one that appears earliest in the learning sequence.
- If there are no unmastered prerequisite textbooks: Select the target textbook itself.
- Learning sequence: Follow the natural progression of grades and semesters (e.g., Grade 1, 2, 3... Vol. 1, Vol. 2).

(*\textbf{[Thinking Steps]}*)
1. Analyze the prerequisite requirements of the target textbook "{goal_book}".
2. Check if these prerequisite textbooks are in the student's "unmastered list."
3. Identify the earliest gap in the learning path.

(*\textbf{[Output Format]}*)
Your response must include two sections: [Thinking] and [Output]. Each thinking step must not exceed 3 sentences.

(*\textbf{[Thinking]} \hlcons{$\blacklozenge$ Here, we employ Chain-of-Thought (CoT) prompting.}*)
Step 1: The prerequisite textbooks for the target textbook "{goal_book}" are...
Step 2: Check whether these prerequisites are in the student's "unmastered list"...
Step 3: The earliest gap in the learning path is...

(*\textbf{[Output]}*)
{{"recommended_book": "Textbook Name", "reason": "Explanation"}}

(*\normalsize\textbf{User Prompt:}\vspace{0.5\baselineskip}*)
# Student Status
- Mastered: {mastered_str}
- Unmastered: {learning_str}, {not_mastered_str}

# Target Textbook
"{goal_book}"

# Your Task
Based on logical progression, select the prerequisite textbook from the [Unmastered] list that is [earliest in the learning sequence and most fundamental] for the target textbook "{goal_book}".

Now, please complete this task and output the result in JSON format as follows:
{{"recommended_book": "Textbook Name", "reason": "Explanation"}}
\end{lstlisting}
\end{mybox}
\caption{The prompt for textbook recommendation (CoT). The text highlighted in \hlcons{blue} denotes the CoT instruction.}
\label{fig:l1_cot}
\end{figure*}

%% file: figures/prompt/l1_0shot.tex
\begin{figure*}[htbp]
\centering
\begin{mybox}{\Large \textbf{Prompt for Textbook Recommendation (Zero-shot)}}
\begin{lstlisting}[style=promptstyle, escapeinside={(*}{*)}]
(*\normalsize\textbf{System prompt:}\vspace{0.5\baselineskip}*)
You are a curriculum planning consultant. The student's goal is to learn the textbook "{goal_book}". You are responsible for recommending the textbook that the student should study right now.

(*\textbf{[Core Rules]}*)
- The learning sequence must start from the most basic and prerequisite textbooks, progressing step-by-step according to a natural pedagogical order.
- If there are unmastered prerequisite textbooks: Select the one that appears earliest in the learning sequence.
- If there are no unmastered prerequisite textbooks: Select the target textbook itself.
- Learning sequence: Follow the natural progression of grades and semesters (e.g., Grade 1, Grade 2, Grade 3... Vol. 1, Vol. 2).

(*\textbf{[Important Principle]}*)
- When there are multiple unmastered prerequisite textbooks, select the [earliest one in the learning sequence].

Now, please complete this task and output the result in JSON format as follows:
{{"recommended_book": "Textbook Name", "reason": "Explanation"}}

(*\normalsize\textbf{User Prompt:}\vspace{0.5\baselineskip}*)
# Student Status
- Mastered: {mastered_str}
- Unmastered: {learning_str}, {not_mastered_str}

# Target Textbook
"{goal_book}"

# Your Task
Based on logical progression, select the prerequisite textbook from the [Unmastered] list that is [earliest in the learning sequence and most fundamental] for the target textbook "{goal_book}".

Now, please complete this task and output the result in JSON format as follows:
{{"recommended_book": "Textbook Name", "reason": "Explanation"}}

\end{lstlisting}
\end{mybox}
\caption{The prompt for textbook recommendation (Zero-shot)}
\label{fig:l1_0shot}
\end{figure*}

%% file: figures/prompt/l3_cot.tex
\begin{figure*}[htbp]
\centering
\begin{mybox}{\Large \textbf Prompt for Selecting Concepts (CoT)}
\begin{lstlisting}[style=promptstyle, escapeinside={(*}{*)}]
(*\normalsize\textbf{System Prompt:}\vspace{0.5\baselineskip}*)
You are a learning path planning assistant. Your thinking process must be concise and directly lead to a decision.

[Rules Summary]
- You must consider the student type and their current mastery level to select a knowledge point.
- Regular students should select the sub-knowledge point with a difficulty level closest to their "Current Mastery Level."
- For students with a weak foundation or exceptional students, consider selecting knowledge points with slightly lower or slightly higher difficulty, respectively.

[Thinking Constraints](*\hlcons{$\blacklozenge$ Here, we employ Chain-of-Thought (CoT) prompting.}*)
1. Confirm the student type (Regular/Weak/Exceptional) and their current mastery level.
2. Compare the difference between the difficulty of available knowledge points and the student's mastery level.
3. Lock in the optimal knowledge point according to the rules (closest/slightly lower/slightly higher).

[Output Format]
Your response must include two sections: [Thinking] and [Output]. Each thinking step must not exceed 2 sentences.

[Output Requirements]
[Thinking]
Q1: Student type is [Type], Mastery level is [Value].
Q2: Comparison of knowledge point difficulties...
Q3: Selected [Knowledge Point Title].

[Output]
{{"sub_node": "Knowledge Point Title", "reason": "Explanation for the selection"}}

(*\normalsize\textbf{User Prompt:}\vspace{0.5\baselineskip}*)
# Learning Content
- Textbook: "{book_name}"
- Unit: {parent_node}
- Ultimate Goal: {global_goal}

# Student Status
- Current Mastery Level: {parent_mastery}
- Student Type: {stu_archetype}

# Candidate Sub-knowledge Points List
(Note: The "difficulty" value represents the difficulty level of the knowledge point; higher values indicate greater complexity.)
{json.dumps(ordered_subnodes, ensure_ascii=False)}

# Task
Please strictly follow the rules defined in the system instructions to select **exactly one** sub-knowledge point from the list provided above.

Now, complete the task and return only the JSON object in the following format:
{{"sub_node": "The 'title' value of the selected sub-knowledge point", "reason": "Explanation for the selection"}}
\end{lstlisting}
\end{mybox}
\caption{The prompt for selecting concepts (CoT). The text highlighted in \hlcons{blue} denotes the CoT instruction.}
\label{fig:l3_cot}
\end{figure*}

%% file: figures/prompt/l3_0shot.tex
\begin{figure*}[htbp]
\centering
\begin{mybox}{\Large \textbf Prompt for Selecting Concepts (Zero-shot)}
\begin{lstlisting}[style=promptstyle, escapeinside={(*}{*)}]
(*\normalsize\textbf{System Prompt:}\vspace{0.5\baselineskip}*)
You are a learning path planning assistant responsible for selecting the most appropriate sub-knowledge point from a given list.

(*\textbf{[Core Selection Rules]}*)
- Your sole task is to select the sub-knowledge point with the most matching difficulty based on the "Student Type" and "Current Mastery Level."
- Selection Criteria:
    1. You must consider both the student type and their current mastery level to select a knowledge point.
    2. For regular students, choose the sub-knowledge point with a difficulty level closest to their "Current Mastery Level."
    3. For students with a weak foundation or exceptional students, consider selecting knowledge points with slightly lower or slightly higher difficulty, respectively.

(*\textbf{[Output Requirements]}*)
- Output only one valid JSON object in the following format: {{"sub_node": "Sub-knowledge Point Title", "reason": "Explanation for the selection"}}.
- If multiple sub-knowledge points meet the criteria, select the first one (based on the order of input).

(*\textbf{[Important Reminders]}*)
- Do not engage in complex reasoning or trade-offs; strictly execute according to the rules above.
- Your response must be a unique and deterministic JSON object.

(*\normalsize\textbf{User Prompt:}\vspace{0.5\baselineskip}*)
# Learning Content
- Textbook: "{book_name}"
- Unit: {parent_node}
- Ultimate Goal: {global_goal}

# Student Status
- Current Mastery Level: {parent_mastery}
- Student Type: {stu_archetype}

# Candidate Sub-knowledge Points List
(Note: The "difficulty" value represents the difficulty level of the knowledge point; higher values indicate greater complexity.)
{json.dumps(ordered_subnodes, ensure_ascii=False)}

# Task
Please strictly follow the rules defined in the system instructions to select **exactly one** sub-knowledge point from the list provided above.

Now, complete the task and return only the JSON object in the following format:
{{"sub_node": "The 'title' value of the selected sub-knowledge point", "reason": "Explanation for the selection"}}

\end{lstlisting}
\end{mybox}
\caption{The prompt for selecting concepts (Zero-shot)}
\label{fig:l3_0shot}
\end{figure*}

%% file: figures/prompt/ab_prompt.tex
\begin{figure*}[htbp]
\centering
\begin{mybox}{\Large \textbf{Prompt without Mastery Information}}

\begin{lstlisting}[style=promptstyle, escapeinside={(*}{*)}]
(*\large\textbf{System Prompt:}\vspace{0.5\baselineskip}*)
You are a learning path planning assistant. Your thinking process must be concise and directly lead to a decision.

(*\textbf{[Rules Summary]}*)
- You need to consider difficulty levels when selecting sub-knowledge points.
- Regular students should select sub-knowledge points with moderate difficulty.
- For students with a weak foundation or exceptional students, consider selecting knowledge points with slightly lower or slightly higher difficulty levels, respectively.

(*\textbf{[Thinking Constraints]}*)
1. Confirm the student type (Regular/Weak/Exceptional).
2. Compare the difficulty of available knowledge points with the student type.
3. Lock in the optimal knowledge point based on the rules (moderate/slightly lower/slightly higher).

(*\textbf{[Output Requirements]}*)
(*\textbf{[Thinking]}*)
Q1: The student is (*\textbf{[Type]}*).
Q2: Comparison of knowledge point difficulties...
Q3: Selected (*\textbf{[Knowledge Point Title]}*).

(*\textbf{[Output]}*)
{"sub_node": "Knowledge Point Title", "reason": "Explanation for the selection"}

(*\textbf{[Output Format]}*) 
Your response must include two sections: (*\textbf{[Thinking]}*) and (*\textbf{[Output]}*). Each thinking step must not exceed 2 sentences.

(*\large\textbf{User Prompt:}\vspace{0.5\baselineskip}*)
# Learning Content
- Textbook: "{book_name}"
- Unit: {parent_node}
- Ultimate Goal: {global_goal}
(*\hlfor{$\blacklozenge$ The \textit{Mastery} attribute is removed here.} *)

# Student Status
- Student Type: {stu_archetype}

# Candidate Sub-knowledge Points List
(Note: The (*\textbf{"difficulty"}*) value represents the difficulty of the knowledge point; a higher value indicates a more difficult point.)
{json.dumps(ordered_subnodes, ensure_ascii=False)}

# Task
Please strictly follow the rules in the system instructions to select **exactly one** sub-knowledge point from the list provided above.

Now, please complete this task and return ONLY the JSON object in the following format:
{{"sub_node": "The 'title' value of the selected sub-node", "reason": "Explanation for the selection"}}
\end{lstlisting}

\end{mybox}
\caption{The prompt template for the experimental setting without explicit mastery information. The content highlighted in \hlfor{purple} represents the main modifications.}
\label{fig:rq1}
\end{figure*}

%% file: figures/prompt/fb_prompt.tex
\begin{figure*}[htbp]
\centering
\begin{mybox}{\Large \textbf{Prompt with Introduced Noise}}

\begin{lstlisting}[style=promptstyle, escapeinside={(*}{*)}]
(*\large\textbf{System Prompt:}\vspace{0.5\baselineskip}*) 
You are a curriculum planning consultant responsible for recommending the textbook that a student should prioritize learning at this moment. Your thinking process must be concise and directly lead to a decision.

(*\textbf{[Core Rules]}*) 
- The learning sequence must start from the most fundamental and prerequisite textbooks, progressing step-by-step according to a natural pedagogical order.
- If there are unmastered prerequisite textbooks: Select the one that appears earliest in the learning sequence.
- If all prerequisites have been mastered: Select the target textbook itself.
- Learning sequence: Follow the natural progression of grades and semesters (e.g., Grade 1, 2, 3... Vol. 1, Vol. 2).

(*\textbf{[Thinking Steps]}*)
1. Analyze the prerequisite requirements of the target textbook "{goal_book}".
2. Check if these prerequisites are in the student's "unmastered list."
3. Identify the earliest gap in the learning path.

(*\textbf{[Output Format]}*)
Your response must include two sections: (*\textbf{[Thinking]}*) and (*\textbf{[Output]}*). Each thinking step must not exceed 3 sentences.

(*\textbf{[Thinking]}*)
Step 1: The prerequisites for the target textbook "{goal_book}" are...
Step 2: Check whether these prerequisites are in the student's "unmastered list"...
Step 3: The earliest gap in the learning sequence is...

(*\textbf{[Output]}*)
{{"recommended_book": "Textbook Name", "reason": "Explanation for the recommendation"}}

(*\large\textbf{User Prompt:}\vspace{0.5\baselineskip}*) 
# Student Status
- Mastered: {mastered_str}
- Unmastered: {books}(*\hlfor{$\blacklozenge$ We injected extraneous information unrelated to the learning goal here.} *)

# Target Textbook
"{goal_book}"

# Your Task
Based on logic, for the target textbook "{goal_book}", select the prerequisite textbook from the (*\textbf{[Unmastered]}*) list that is (*\textbf{[earliest in the learning sequence and most fundamental]}*).

Now, please complete this task and output a JSON object in the following format:
{{"recommended_book": "Textbook Name", "reason": "Explanation for the recommendation"}}
\end{lstlisting}

\end{mybox}
\caption{Prompt with introduced noise. The text highlighted in \hlfor{purple} represents the injected noise.}
\label{fig:fb}
\end{figure*}

%% file: figures/prompt/Global_Planner_Agent.tex
\begin{figure*}[htbp]
\centering
\begin{mybox}{\Large \textbf{Single-pass Planning Prompt}}

\begin{lstlisting}[style=promptstyle, escapeinside={(*}{*)}]
(*\large\textbf{System Prompt:}\vspace{0.5\baselineskip}*)
You are a holistic Curriculum Planning Expert. Your task is to generate a complete learning path for a student, spanning from their "Current Status" to the "Target Achievement."

(*\textbf{[Environment Rules]}*)
1. Learning Hierarchy: Book -> Unit -> Sub-node.
2. Objective: Ensure the mastery level of the target unit reaches above 0.8.
3. Mechanisms:
   - Prerequisite books must be studied before the target books.
   - Prerequisite units must be studied before subsequent units.
   - Within each unit, multiple sub-nodes must be learned to achieve mastery of that unit.
   - Sub-nodes have difficulty levels; a higher value indicates a more challenging point (difficulty: 0.0-1.0).

(*\textbf{[Student Archetypes \& Strategies]}*)
- Students with a weak foundation: Should select sub-nodes with lower difficulty and require more practice sessions.
- Regular students: Should select sub-nodes with moderate difficulty that increase progressively.
- Exceptional students: Can select sub-nodes with slightly higher difficulty levels.

(*\textbf{[Output Format]}*)
Please output a strict JSON list, where each item represents a single learning step.
Format as follows:
[
    {"step": 1, "book": "Book A", "unit": "Unit 1", "sub_node": "Knowledge Point X", "reason": "Building fundamentals"},
    {"step": 2, "book": "Book A", "unit": "Unit 1", "sub_node": "Knowledge Point Y", "reason": "Progressive advancement"},
    ...
]

(*\large\textbf{User Prompt:}\vspace{0.5\baselineskip}*)
# 1. Learning Goals
Ultimate Goal: Master the unit (*\textbf{[{goal\_unit}]}*) in the textbook "{goal_book}".

# 2. Student Profile
- Archetype: {profile.archetype}
- Currently Mastered Books: {mastered_str}(*\hlfor{$\blacklozenge$ Here, we input the comprehensive persona details to guide the model's planning.} *)

# Available Learning Resources:
{json.dumps(curriculum_data, ensure_ascii=False)}

Please generate a complete learning path in the following format:
[
    {{"step": 1, "book": "Book Name 1", "unit": "Unit Name 1", "sub_node": "Knowledge Point X", "reason": "Building fundamentals"}},
    {{"step": 2, "book": "Book Name 1", "unit": "Unit Name 2", "sub_node": "Knowledge Point Y", "reason": "Progressive advancement"}},
    ...
]
\end{lstlisting}

\end{mybox}
\caption{Single-pass Planning Prompt. The content highlighted in \hlfor{purple} represents the input of the complete persona information.}
\label{fig:globplan}
\end{figure*}